%% file: main.tex
\documentclass[]{style}

\input{resources/packages}
\input{metadata}

\colorlet{OursRowColor}{WeChatGreen!12}
\newcommand{\oursrow}{\rowcolor{OursRowColor}}

\definecolor{codebg}{RGB}{248,248,248}
\definecolor{codecomment}{RGB}{100,100,100}
\lstdefinelanguage{json}{
  basicstyle=\ttfamily\small,
  numbers=none,
  breaklines=true,
  frame=none,
  backgroundcolor=\color{codebg},
  showstringspaces=false,
  literate= *{0}{{{\color{blue}0}}}{1} {1}{{{\color{blue}1}}}{1}
    {2}{{{\color{blue}2}}}{1} {3}{{{\color{blue}3}}}{1}
    {4}{{{\color{blue}4}}}{1} {5}{{{\color{blue}5}}}{1}
    {6}{{{\color{blue}6}}}{1} {7}{{{\color{blue}7}}}{1}
    {8}{{{\color{blue}8}}}{1} {9}{{{\color{blue}9}}}{1}
    {:}{{{\color{gray}:}}}{1} {,}{{{\color{gray},}}}{1}
}
\newtcblisting{jsonbox}{
  listing only,
  listing options={language=json,basicstyle=\ttfamily\scriptsize,
    breaklines=true,breakatwhitespace=false,columns=fullflexible,
    keepspaces=true,showstringspaces=false},
  colback=codebg,colframe=black!15,boxrule=0.5pt,arc=2pt,
  left=5pt,right=5pt,top=5pt,bottom=5pt
}

\AtBeginDocument{%
}
\newcommand{\appref}[1]{\hyperref[#1]{Appendix~\ref*{#1}}}

\let\reportsection\section
\renewcommand{\section}{\Needspace*{5\baselineskip}\reportsection}

\newcommand{\boldtitle}[1]{{\bfseries #1}}
\newcommand\blfootnote[1]{%
  \begingroup
  \renewcommand\thefootnote{}%
  \begin{NoHyper}%
    \footnotetext{#1}%
  \end{NoHyper}%
  \endgroup
}
\title{\boldtitle{\ModelName: From Coverage to Capability for Robust}\\
  End-to-End Document Parsing}
\author{\ReportAuthorNames}
\affiliation{\ReportAffiliation}
\abstract{\ReportAbstract}
\DeclareRobustCommand{\frontlinkicon}[1]{%
  \raisebox{-0.3em}{\includegraphics[width=0.025\linewidth]{#1}}}
\checkdata[\frontlinkicon{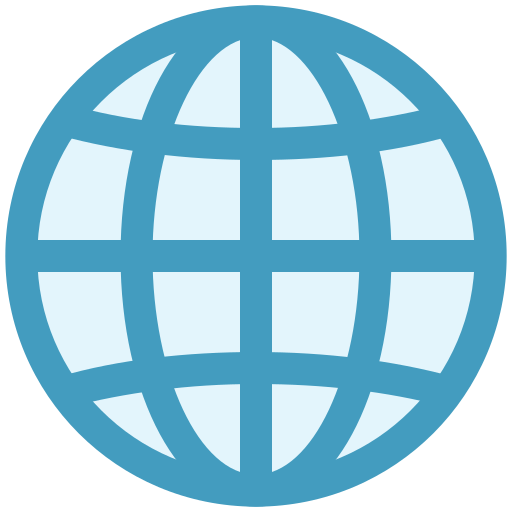}~~Project Page]{%
  \href{\ReportProjectPageURL}{\texttt{\ReportProjectPageURL}}}
\checkdata[\frontlinkicon{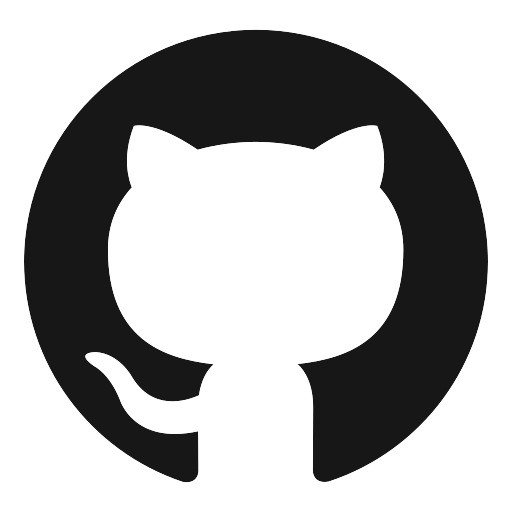}~~GitHub]{%
  \href{\ReportGitHubURL}{\texttt{\ReportGitHubURL}}}
\checkdata[\frontlinkicon{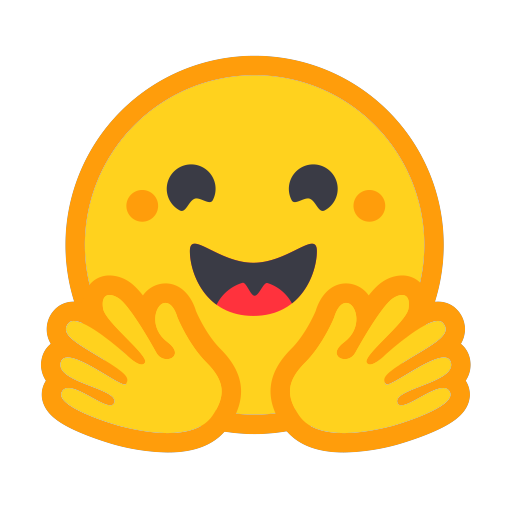}~~WeVisDoc-4B]{%
  \href{\ReportHuggingFaceURLFourB}{\texttt{\ReportHuggingFaceURLFourB}}}
\checkdata[\frontlinkicon{assets/huggingface-icon.png}~~WeVisDoc-2B]{%
  \href{\ReportHuggingFaceURLTwoB}{\texttt{\ReportHuggingFaceURLTwoB}}}

\begin{document}

\maketitle
\blfootnote{\ReportContributionNote}
\input{sections/00_front_page}

\input{sections/01_introduction}
\input{sections/02_related_work}
\input{sections/03_method}
\input{sections/03_experimental_instantiation}
\input{sections/05_experiments}
\input{sections/08_conclusion}

\clearpage
\bibliographystyle{plainnat}
\setlength{\bibhang}{0pt}
\setlength\bibindent{0pt}
\bibliography{references}

\clearpage
\appendix
\input{sections/G_doc_parsing_prompt}
\clearpage
\input{sections/H_qualitative_case_studies}

\end{document}

%% file: resources/packages.tex
\usepackage{xargs}

\usepackage{multirow}

\usepackage{cleveref}

\usepackage{amsmath}
\usepackage{dsfont}
\usepackage{array}
\usepackage{longtable}
\usepackage{xurl}
\usepackage{needspace}
\usepackage{tikz}
\usetikzlibrary{arrows.meta,calc,positioning,shapes.geometric}
\usepackage{pgfplots}
\pgfplotsset{compat=1.18}
\usepgfplotslibrary{groupplots}

\usepackage{mathrsfs}
\usepackage{adjustbox}
\usepackage{multirow}
\usepackage{multicol}
\usepackage{tcolorbox}
\usepackage{changepage}
\usepackage{enumitem}
\usepackage{graphicx}
\usepackage{amssymb}
\usepackage{xcolor}
\usepackage{float}
\usepackage{multirow}
\usepackage{threeparttable}
\usepackage{graphicx}
\usepackage{wrapfig}
\usepackage[table]{xcolor}  
\usepackage{colortbl}       
\usepackage{tabularx} 
\usepackage{makecell} 
\usepackage[dvipsnames]{xcolor}

\newcolumntype{g}{>{\columncolor{gray!10}}c} 
\newcolumntype{A}{>{\raggedright\arraybackslash}p{4.5cm}} 
\newcolumntype{Y}{>{\raggedright\arraybackslash}X} 

\definecolor{catgray}{gray}{0.9}
\definecolor{skyblue}{rgb}{0.53,0.81,0.92} 

\colorlet{skyblue!30}{skyblue!30!white} 

\colorlet{themeboxborder}{WeChatDeepGreen}

\newtcolorbox{evolbox}[2][]{%
  enhanced,
  colframe=themeboxborder,
  colback=white,
  coltitle=white,
  rounded corners,
  boxrule=1pt,
  titlerule=0pt,
  toptitle=1mm,
  bottomtitle=1mm,
  fonttitle=\bfseries,
  width=#2\textwidth, 
  #1
}

\usepackage{url}

\PassOptionsToPackage{table,xcdraw}{xcolor}
\usepackage{titletoc}
\usepackage{placeins}
\usepackage{pifont}

\colorlet{RowTheme}{WeChatMint}
\definecolor{RowRed}{HTML}{F9EAEA}
\definecolor{Top1}{HTML}{50DB4B} 
\definecolor{Top2}{HTML}{A5FFA2} 
\definecolor{Top3}{HTML}{D9FFD9} 
\definecolor{Sub1}{HTML}{EAB8B8}
\definecolor{Sub2}{HTML}{E4E4E4}

\definecolor{gearred}{HTML}{D85140}
\definecolor{reprablue}{HTML}{5384ED}
\definecolor{steorange}{HTML}{EF8444}
\definecolor{softgreen}{HTML}{658E40}

\renewcommand{\emph}[1]{\textit{#1}}

\definecolor{codepink}{RGB}{220,20,120}
\definecolor{codegreen}{RGB}{0,150,0}
\definecolor{codegray}{RGB}{140,140,140}
\definecolor{codeorange}{RGB}{230,120,60}

\lstdefinestyle{pytorchstyle}{
    language=Python,
    basicstyle=\ttfamily\small,
    keywordstyle=\color{codepink}\bfseries,
    commentstyle=\color{codegray}\itshape,
    stringstyle=\color{codeorange},
    numberstyle=\tiny\color{codegray},
    numbers=none,
    showstringspaces=false,
    breaklines=true,
    frame=none,
    columns=fullflexible,
    keepspaces=true,
    xleftmargin=1.5em
}

%% file: metadata.tex
\newcommand{\ModelName}{WeVisDoc}

\newcommand{\ModelNameTwoB}{\ModelName-\allowbreak 2B}
\newcommand{\ModelNameFourB}{\ModelName-\allowbreak 4B}

\newcommand{\ReportTitle}{\ModelName: From Coverage to Capability for Robust End-to-End Document Parsing}

\newcommand{\ReportAuthorNames}{%
  Hao Yu\textsuperscript{*}, Kang Liu\textsuperscript{*},
  Linnan Zhao\textsuperscript{*}, Jiabo Zhan\textsuperscript{*},
  Chong Sun\textsuperscript{\ensuremath{\dagger}},
  Chen Li\textsuperscript{\ensuremath{\ddagger}}, Jing LYU%
}
\newcommand{\ReportAffiliation}{WeChat Vision, Tencent Inc.}
\newcommand{\ReportContributionNote}{%
  \textsuperscript{*}Equal contribution.\enspace
  \textsuperscript{\ensuremath{\dagger}}Project leader.\enspace
  \textsuperscript{\ensuremath{\ddagger}}Corresponding author.%
}

\newcommand{\ReportProjectPageURL}{https://tencent.github.io/WeVisDoc}
\newcommand{\ReportGitHubURL}{https://github.com/Tencent/WeVisDoc}
\newcommand{\ReportHuggingFaceURLFourB}{https://huggingface.co/Tencent/WeVisDoc-4B}
\newcommand{\ReportHuggingFaceURLTwoB}{https://huggingface.co/Tencent/WeVisDoc-2B}

\newcommand{\ReportAbstract}{%
  Document parsing converts document images into structured content and requires reliable performance across diverse layouts and acquisition conditions.
  Yet training corpora are biased toward common document types and clean digital pages, while expanding coverage alone does not specify how to address a parser's remaining weaknesses.
  We present \ModelName{}, a two-stage data-centric framework for robust end-to-end document parsing.
  Stage~I broadens semantic, structural, and appearance coverage through heterogeneous data and structure-preserving degradation synthesis.
  Stage~II uses a held-out probe to measure the Stage~I parser's residual errors within fixed visual--structural clusters.
  These diagnostics guide targeted data construction and reallocation of the target-token budget.
  \ModelNameFourB{} achieves an Overall score of 95.38 on OmniDocBench v1.6 and a mean Overall score of 75.54 across the three PureDocBench tracks, ranking first among the compared end-to-end parsers in all four settings.
  Compared with Stage~I, Stage~II improves Overall scores for the 2B and 4B models on both benchmarks, with larger gains on the degraded PureDocBench tracks, including a 4.03-point gain for the 4B model on the Real Degraded track.%
}

\hypersetup{
  pdftitle={\ReportTitle}, pdfauthor={Hao Yu, Kang Liu, Linnan Zhao, Jiabo Zhan, Chong Sun, Chen Li, Jing LYU}, pdfsubject={Data-centric specialization for robust end-to-end document parsing}, pdfkeywords={document parsing, OCR, vision-language model, degradation synthesis, data mixture, residual-aware rebalancing}
}

%% file: sections/00_front_page.tex
\vspace{-6mm}
\noindent\begin{minipage}{\linewidth}
  \centering
  \input{sections/00_performance_chart}
  \captionsetup{hypcap=false,skip=6pt}
  
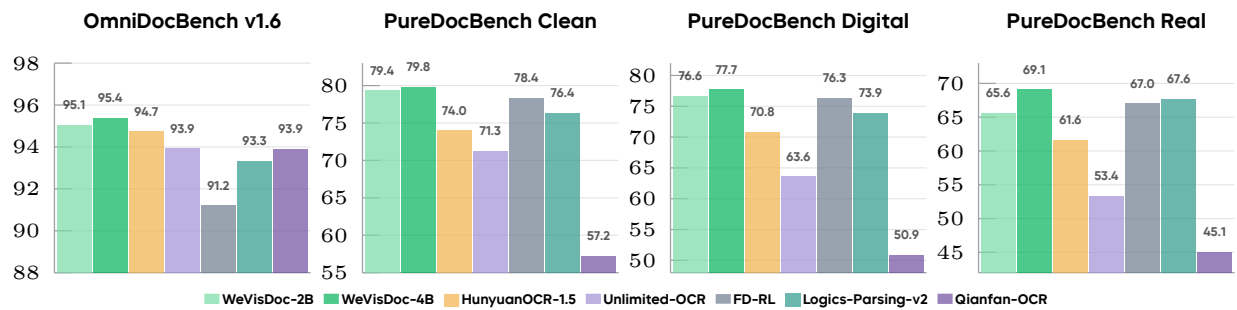
\captionof{figure}{\small\textbf{\ModelNameFourB{} leads the compared
    end-to-end parsers across all four reported settings.} Bars show scores on
    OmniDocBench v1.6 and the three PureDocBench robustness tracks;
    \ModelNameTwoB{} delivers top-tier end-to-end performance at half the parameter count.}
  \label{fig:performance-overview}
\end{minipage}
\clearpage

%% file: sections/00_performance_chart.tex
\definecolor{PerfOursTwo}{HTML}{7ADBA6}
\definecolor{PerfOursFour}{HTML}{07B45E}
\definecolor{PerfHunyuan}{HTML}{F2B34F}
\definecolor{PerfUnlimited}{HTML}{A393D6}
\definecolor{PerfFDRL}{HTML}{718092}
\definecolor{PerfLogics}{HTML}{319B8D}
\definecolor{PerfQianfan}{HTML}{7352A3}

\newcommand{\PerfLegendItem}[2]{%
  \begingroup
    \setbox0=\hbox{#2}%
    \raisebox{0.15pt}{{\color{#1!72}\vrule width 2mm height\ht0 depth\dp0}}%
    \hspace{0.35mm}#2%
  \endgroup
}

\newcommand{\PerfBar}[4]{%
  \addplot+[
    ybar,
    bar width=8.4pt,
    mark=none,
    draw=none,
    fill=#1,
    fill opacity=0.72,
    nodes near coords={#4},
    bar shift=0pt
  ] coordinates {(#2,#3)};
}

\vspace{0.8mm}
\resizebox{\linewidth}{!}{%
\begin{tikzpicture}
  \begin{groupplot}[
    group style={
      group size=4 by 1,
      horizontal sep=4.5mm
    },
    width=38mm,
    height=34mm,
    xmin=0.45,
    xmax=7.55,
    xtick=\empty,
    tick label style={font=\fontsize{3.7}{4.1}\selectfont\sffamily},
    title style={
      font=\fontsize{5.1}{5.7}\selectfont\sffamily\bfseries,
      at={(axis description cs:0.5,1.015)},
      anchor=south,
      yshift=-1.2mm
    },
    axis x line*=bottom,
    axis y line*=left,
    axis line style={WeChatGray!72,line width=.32pt},
    tick style={WeChatGray!72,line width=.3pt},
    ymajorgrids=true,
    grid style={WeChatLine!68,line width=.28pt},
    clip=true,
    every node near coord/.append style={
      font=\fontsize{3.1}{3.5}\selectfont\sffamily,
      text=WeChatInk,
      anchor=south,
      yshift=.35pt
    }
  ]
    \nextgroupplot[
      title={\strut OmniDocBench v1.6},
      ymin=88,ymax=98,
      ytick={88,90,92,94,96,98}
    ]
      \PerfBar{PerfOursTwo}{1}{95.06}{95.1}
      \PerfBar{PerfOursFour}{2}{95.38}{95.4}
      \PerfBar{PerfHunyuan}{3}{94.74}{94.7}
      \PerfBar{PerfUnlimited}{4}{93.92}{93.9}
      \PerfBar{PerfFDRL}{5}{91.21}{91.2}
      \PerfBar{PerfLogics}{6}{93.33}{93.3}
      \PerfBar{PerfQianfan}{7}{93.90}{93.9}

    \nextgroupplot[
      title={\strut PureDocBench Clean},
      ymin=55,ymax=83,
      ytick={55,60,65,70,75,80}
    ]
      \PerfBar{PerfOursTwo}{1}{79.36}{79.4}
      \PerfBar{PerfOursFour}{2}{79.81}{79.8}
      \PerfBar{PerfHunyuan}{3}{73.98}{74.0}
      \PerfBar{PerfUnlimited}{4}{71.28}{71.3}
      \PerfBar{PerfFDRL}{5}{78.38}{78.4}
      \PerfBar{PerfLogics}{6}{76.35}{76.4}
      \PerfBar{PerfQianfan}{7}{57.22}{57.2}

    \nextgroupplot[
      title={\strut PureDocBench Digital},
      ymin=48,ymax=82,
      ytick={50,55,60,65,70,75,80}
    ]
      \PerfBar{PerfOursTwo}{1}{76.62}{76.6}
      \PerfBar{PerfOursFour}{2}{77.74}{77.7}
      \PerfBar{PerfHunyuan}{3}{70.81}{70.8}
      \PerfBar{PerfUnlimited}{4}{63.62}{63.6}
      \PerfBar{PerfFDRL}{5}{76.33}{76.3}
      \PerfBar{PerfLogics}{6}{73.85}{73.9}
      \PerfBar{PerfQianfan}{7}{50.85}{50.9}

    \nextgroupplot[
      title={\strut PureDocBench Real},
      ymin=42,ymax=73,
      ytick={45,50,55,60,65,70}
    ]
      \PerfBar{PerfOursTwo}{1}{65.60}{65.6}
      \PerfBar{PerfOursFour}{2}{69.08}{69.1}
      \PerfBar{PerfHunyuan}{3}{61.59}{61.6}
      \PerfBar{PerfUnlimited}{4}{53.39}{53.4}
      \PerfBar{PerfFDRL}{5}{67.04}{67.0}
      \PerfBar{PerfLogics}{6}{67.64}{67.6}
      \PerfBar{PerfQianfan}{7}{45.06}{45.1}
  \end{groupplot}
\end{tikzpicture}
}\\[-2.7mm]
{\sffamily\fontsize{5.2}{5.7}\selectfont
\makebox[\linewidth][c]{%
  \PerfLegendItem{PerfOursTwo}{\ModelNameTwoB{}}\hspace{1.3mm}%
  \PerfLegendItem{PerfOursFour}{\ModelNameFourB{}}\hspace{1.3mm}%
  \PerfLegendItem{PerfHunyuan}{HunyuanOCR-1.5}\hspace{1.3mm}%
  \PerfLegendItem{PerfUnlimited}{Unlimited-OCR}\hspace{1.3mm}%
  \PerfLegendItem{PerfFDRL}{FD-RL}\hspace{1.3mm}%
  \PerfLegendItem{PerfLogics}{Logics-Parsing-v2}\hspace{1.3mm}%
  \PerfLegendItem{PerfQianfan}{Qianfan-OCR}%
}}\par

%% file: sections/01_introduction.tex
\section{Introduction}
\label{sec:introduction}

Document parsing maps a document image to structured text while preserving textual content, element types, structural organization, and reading order. Modern parsers either decode an entire page directly~\cite{kim2022ocr,blecher2024nougat} or combine global layout understanding with regional recognition~\cite{feng2025dolphin,niu2026mineru2}. Despite rapid architectural progress, reliable parsing remains challenging because documents vary simultaneously in language, layout, content density, element composition, and acquisition condition. These variations produce distinct failure modes that extend beyond character recognition. Benchmarks such as OmniDocBench~\cite{ouyang2025omnidocbench} and PureDocBench~\cite{li2026far} evaluate complementary aspects of this variation and underscore the importance of training-data composition alongside parser architecture.

Recent work has therefore devoted substantial effort to improving document training data. Large-scale curation, annotation repair, and controllable synthesis broaden linguistic and structural diversity~\cite{wei2024vary,wang2026mineru2,li2026towards,huang2026infinity}. Degradation pipelines expand appearance variation for training~\cite{groleau2023augraphy}, while source-aligned acquisition tracks measure robustness under such variation~\cite{li2026far}. Model-aware strategies in document parsing and language-model pretraining use output divergence, proxy mixture search, or representation-space diagnostics to prioritize training data or optimize data mixtures~\cite{cui2026paddleocr,diao2026nemotron,zhang2026paddleocr}. A central challenge for refinement is to distinguish weaknesses that reflect missing data support from those associated with insufficient exposure to patterns already represented in the training distribution.

Expanding empirical support and increasing exposure to existing support are distinct interventions. Collecting, retrieving, or synthesizing records expands empirical support, whereas replaying or reweighting existing records changes their exposure. For autoregressive parsers, we characterize the supervised optimization budget in terms of loss-bearing target tokens rather than record count because records vary substantially in target length. Residual error alone is also insufficient evidence of a coverage deficit: it may arise from unreliable targets, low visual observability, decoding failure, or model limitations that additional nearby examples cannot resolve. Effective allocation therefore requires an explicit assessment of whether the relevant training support is absent.

We organize document-data scaling into two stages: coverage construction followed by capability-aware refinement. The first stage establishes broad training support; the second diagnoses residual weaknesses, validates targeted additions, and reallocates supervised exposure over the revised pool. We introduce \ModelName{}, a two-stage data-centric framework that implements this procedure for document parsing, with a separate target-token budget for each stage.

Stage~I constructs broad, validated coverage from approximately 40 million records spanning supervision granularities, layouts, languages, data sources, and acquisition conditions. It combines heterogeneous supervision, executable page synthesis, and source-conditioned appearance generation using unified parsing targets and annotation-quality checks. Source-aware balancing limits domination by large corpora, while appearance variants expand visual coverage without being treated as additional semantic support. Stage~II then refines the parser according to its residual weaknesses. A diagnostic probe, disjoint from training, hard-example mining, and final evaluation, measures residual parsing errors. Together with audits of existing support, these measurements guide the choice between adding validated data and increasing exposure to available records. Validated additions, Stage~I replay, and separately audited hard examples form a refinement pool of approximately 5 million records, with all exposure counted against a fixed Stage~II target-token budget.

We instantiate \ModelName{} from Qwen3-VL-Instruct~\cite{bai2025qwen3} at 2B and 4B scales, keeping the backbone, output serialization, and autoregressive objective fixed within each scale. Figure~\ref{fig:performance-overview} compares the resulting models with five representative end-to-end systems on OmniDocBench v1.6 and the three PureDocBench tracks. On OmniDocBench v1.6, the final 2B and 4B models achieve Overall scores of 95.06 and 95.38, respectively; their corresponding $\mathrm{Avg}_3$ scores on PureDocBench are 73.86 and 75.54. The 4B model ranks first among the compared end-to-end parsers in all four settings, while the 2B model remains competitive with half as many parameters. The complete Stage~II protocol improves both benchmarks at both scales, with larger gains on degraded pages than on clean pages. These stage-wise results are consistent with the benefit of the complete refinement protocol, particularly under degraded acquisition conditions; because Stage~II combines several interventions, they do not identify the effect of residual-aware allocation in isolation.

%% file: sections/02_related_work.tex
\section{Related Work}
\label{sec:related-work}

\subsection{Parsing Architectures}

Generative document parsers convert document images into structured text, providing a common output interface for text, tables, and formulas. Early image-to-sequence methods, including Donut~\cite{kim2022ocr}, Pix2Struct~\cite{lee2023pix2struct}, and Nougat~\cite{blecher2024nougat}, established this direction through document understanding, screenshot pretraining, and scientific page conversion. More recent systems improve how page information is represented and encoded: SmolDocling~\cite{nassar2025smoldocling} uses DocTags to express content and structure, while DeepSeek-OCR~\cite{wei2025deepseek} and DeepSeek-OCR 2~\cite{wei2026deepseek} explore visual context compression and semantic reordering of visual tokens, respectively.

Another line of work uses page layout to organize regional recognition. Dolphin~\cite{feng2025dolphin} and MonkeyOCR~\cite{li2025monkeyocr} decompose parsing into structural analysis and content recognition, while MinerU2.5~\cite{niu2026mineru2} and PaddleOCR-VL-\allowbreak 1.5~\cite{cui2026paddleocr} combine page-level layout analysis with high-resolution regional processing. PaDoc~\cite{yu2026padoc} connects parallel regional decoding with shared full-page visual context. These approaches explore different ways to coordinate global structure and local recognition. \ModelName{} adopts end-to-end parsing and studies how training-data coverage and allocation improve a parser within the same backbone architecture at each model scale.

\subsection{Data Coverage and Curation for Document Specialists}

The coverage and quality of supervision shape which document patterns a specialist can learn. Vary~\cite{wei2024vary} emphasizes dense document perception, including non-English content; mPLUG-DocOwl 1.5~\cite{hu2024mplug} combines structural supervision with text localization across diverse visual domains; and dots.ocr~\cite{li2025dots} uses multilingual data to jointly learn layout, content, and relations. Together, these works highlight complementary dimensions of document supervision. Benchmarks such as OmniDocBench~\cite{ouyang2025omnidocbench} make this diversity visible through evaluation across heterogeneous pages and element types.

Recent data-centric systems make corpus construction an explicit part of parser development. MinerU2.5-Pro~\cite{wang2026mineru2} combines diversity-aware sampling with annotation verification and repair. DocHumming~\cite{li2026towards} and Infinity-Parser2~\cite{huang2026infinity} broaden supervision through layout composition and controllable rendering, respectively, while Infinity-Parser~\cite{wang2026infinityparser} pairs a curated parsing corpus with layout-aware reinforcement learning. These efforts motivate broad, reliable coverage, but effective training also requires deciding which patterns need additional data or greater exposure. \ModelName{} connects these decisions through two stages: broad coverage construction followed by refinement guided by the trained parser's remaining weaknesses.

\subsection{Robustness to Document Degradation}

Beyond content and layout diversity, document parsers must remain reliable under degradations introduced by scanning, photography, and reproduction. Document unwarping methods such as DewarpNet~\cite{das2019dewarpnet} and UVDoc~\cite{verhoeven2023uvdoc} correct geometric distortions to recover a flat view before recognition. Data augmentation offers a complementary route: Augraphy~\cite{groleau2023augraphy} simulates artifacts from printing, scanning, and related processes to expose models to degraded document images during training.

Recent parsing studies evaluate how document degradation affects parsing quality. PaddleOCR-VL-1.5~\cite{cui2026paddleocr} and DocHumming~\cite{li2026towards} cover geometric distortions and degradation in photographed or screen-mediated documents. PureDocBench~\cite{li2026far} evaluates aligned clean and degraded views using a common source-derived target. For supervised parser training, degraded views provide valid supervision only when their content remains readable and consistent with the target. \ModelName{} incorporates degradation-based augmentation into Stage~I coverage, validating degraded image--target pairs and controlling their training exposure to improve robustness while preserving semantic and structural supervision.

\subsection{Model-Aware Data Allocation}

Data selection and weighting seek to use finite training budgets more effectively. Curriculum learning~\cite{bengio2009curriculum} organizes exposure by difficulty, while DoReMi~\cite{xie2023doremi} and Nemotron-CLIMB~\cite{diao2026nemotron} use proxy models to optimize domain or cluster mixtures. Rho-1~\cite{lin2024rho} instead selects tokens according to excess loss relative to a reference model. At the sample level, Core-Set~\cite{sener2017active}, JTT~\cite{liu2021just}, and LESS~\cite{xia2024less} prioritize representation coverage, initial-model errors, and gradient similarity to target examples, respectively. These methods provide complementary ways to decide where training effort should be spent.

Document-specific methods increasingly connect such decisions to diagnosed parsing weaknesses. Uncertainty-Aware Cluster Sampling~\cite{cui2026paddleocr} uses stochastic-output divergence to allocate samples across visual clusters. MinerU2.5-Pro~\cite{wang2026mineru2} and PaddleOCR-VL-1.6~\cite{zhang2026paddleocr} combine data grouping and model feedback with targeted data expansion or annotation repair, while HunyuanOCR-1.5~\cite{li2026hunyuanocr} turns long-tail weaknesses into data-construction requirements. Training feedback can also shape optimization directly: Infinity-Parser~\cite{wang2026infinityparser} and FD-RL~\cite{zhong2026fdrl} use reinforcement learning with document-specific rewards. Together, these approaches motivate connecting model diagnosis to both data construction and subsequent training.

\ModelName{} combines broad coverage with model-aware refinement in two-stage supervised training. Stage~I trains the parser on diverse document content, structures, and degradations. Stage~II uses the resulting model's remaining weaknesses to guide further data construction and training. We measure page and component errors against ground-truth targets on a diagnostic probe disjoint from training, hard-example mining, and final evaluation, and aggregate them over representation-space groups. Combined with checks on existing data coverage, these signals guide targeted additions and exposure allocation under a fixed budget of supervised output tokens. Pool expansion and exposure reallocation are therefore separate interventions in supervised model training. Within each model scale, training updates model parameters while preserving the backbone architecture, output serialization, and autoregressive objective.

%% file: sections/03_method.tex
\section{Method}
\label{sec:method}

Our method first builds broad data coverage, then adds data and adjusts training to address remaining parsing errors. Section~\ref{sec:formulation} defines the task and training objective; Sections~\ref{sec:stage1-support} and~\ref{sec:stage2-allocation} describe the two stages, and Section~\ref{sec:model-setup} gives the experimental configuration.

\subsection{Task Formulation and Method Overview}
\label{sec:formulation}

\paragraph{Unified document parsing.}
Document parsing converts document images into readable text while preserving their structure and reading order. For record $i$, the input image $x_i$ may contain a complete page, a text crop, a mixed-content region, an isolated formula, or a table. Its target $y_i$ represents the visible content in reading order: text uses Markdown, formulas use LaTeX, and tables use HTML. Full pages and mixed-content regions combine these representations in one sequence; crops use the corresponding format for the content they contain.

A parser with parameters $\theta$ models the target sequence $y_i=(y_{i,1},\ldots,y_{i,T_i})$ autoregressively:
\begin{equation}
  p_\theta(y_i\mid x_i)
  =\prod_{t=1}^{T_i}p_\theta(y_{i,t}\mid x_i,y_{i,<t}).
  \label{eq:parsing-factorization}
\end{equation}
where $T_i$ is the target length, $t$ indexes its tokens, and $y_{i,<t}$ denotes the preceding tokens. The same parser and output formats apply to pages, regions, and components. During evaluation, the target is represented as a decoded, normalized string.

\paragraph{Supervision and training exposure.}
Let $m_{i,t}\in\{0,1\}$ indicate whether target position $t$ contributes to training. The effective target length $L_i$ and summed negative log-likelihood $\ell_i(\theta)$ are
\begin{equation}
\begin{aligned}
  L_i&=\sum_{t=1}^{T_i}m_{i,t}>0,\\
  \ell_i(\theta)&=-\sum_{t=1}^{T_i}m_{i,t}
  \log p_\theta(y_{i,t}\mid x_i,y_{i,<t}).
\end{aligned}
  \label{eq:record-supervision}
\end{equation}
Only target tokens included in the loss count toward $L_i$; input tokens, padding, and discarded target positions do not. For a record-sampling distribution $q$, where $q(i)$ is the probability of drawing record $i$, the training objective, normalized by supervised target length, is
\begin{equation}
  \mathcal L(\theta;q)=
  \frac{\mathbb E_{i\sim q}[\ell_i(\theta)]}
       {\mathbb E_{i\sim q}[L_i]}.
  \label{eq:objective}
\end{equation}
This defines loss per supervised token rather than an equal average of per-record losses. Long pages contribute more supervised tokens per sampled record than short components, so pool size and record-sampling probability do not directly determine training exposure. Section~\ref{sec:source-allocation} describes how desired token shares are converted into record-sampling probabilities.

\paragraph{From coverage to capability.}
Starting from pretrained parameters $\theta_0$, Stage~I produces checkpoint $\theta_1$ by sampling from distribution $q_1$ under a supervised-token budget $B_1$. It establishes broad parsing capability across content, layouts, languages, and acquisition conditions, with appearance augmentation adding visual variation while keeping the content readable (Section~\ref{sec:stage1-support}).

Stage~II diagnoses the remaining errors of $\theta_1$, separates annotation errors from verified model errors, and adds supervision for insufficiently covered patterns. It then trains under a revised distribution $q_2$ and token budget $B_2$ to obtain $\theta_2$. The revised mixture assigns more training tokens to examples with verified model errors while retaining Stage~I examples to reduce forgetting. Sections~\ref{sec:hard-refinement}--\ref{sec:rebalancing} describe verification, diagnosis, targeted additions, and rebalancing. Both stages use Equation~\ref{eq:objective}; Stage~II freezes the visual encoder and updates the language model (Section~\ref{sec:model-setup}).

\subsection{Stage~I: Broad-Coverage Data Construction}
\label{sec:stage1-support}
\label{sec:data-engine}

\begin{figure}[htbp]
\centering
\includegraphics[width=\linewidth]{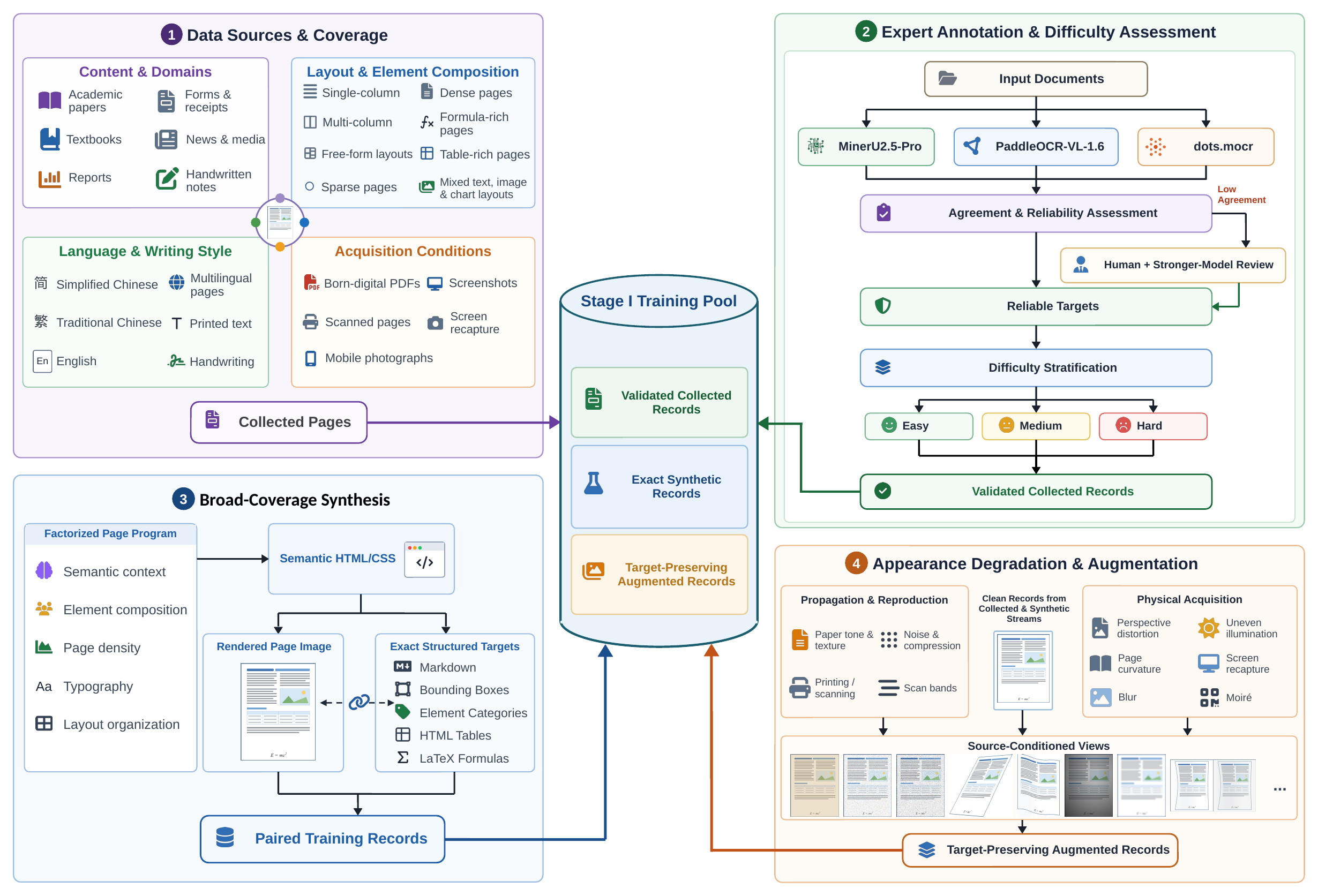}
\caption{Stage~I data construction. Collected sources are normalized through multi-model labeling, while general synthesis generates exact semantic and structural targets. Both data sources enter a shared clean pool. Appearance augmentation is then applied according to the source to form the Stage~I mixture.}
\label{fig:stage1-workflow}
\end{figure}

Stage~I builds a pool of approximately 40 million records by combining data sources, converting annotations, filtering low-quality data, and generating new examples. \emph{Supervision conversion} turns source annotations into training pairs for complete pages, regions, or components (Section~\ref{sec:supervision-granularities}). Coverage spans document domains, layouts, output structures, languages, and acquisition conditions, with these attributes tracked separately. Figure~\ref{fig:stage1-workflow} summarizes the construction process.

\subsubsection{Data Sources and Coverage}
\label{sec:data-coverage}
\label{sec:semantic-coverage}

\paragraph{Diverse data sources.}
The corpus combines open-source datasets, in-house collections, targeted web and PDF crawling, and generated documents. Public datasets provide broad OCR coverage and reusable annotations for articles, books, forms, receipts, tables, and formulas. In-house collections add production documents and acquisition conditions underrepresented in benchmarks. Targeted crawling adds examples from underrepresented domains and patterns, including educational material, newspapers, handwriting, and mixed-language pages.

Real documents preserve the relationships among content, layout, fonts, and visual effects introduced during capture or scanning. Synthetic data systematically vary structures that are rare in collected sources. Source information keeps collected and synthetic data separately traceable for analysis. Appearance variants remain linked to their source page, separating visual augmentation from new document content.

Source and generation metadata support split construction, deduplication (removing or grouping repeated content), source-aware sampling, and error analysis.

\paragraph{Page, region, and component supervision.}
\label{sec:supervision-granularities}
The corpus includes complete pages, pure-text crops, mixed-content regions, isolated tables, and isolated formulas. Full pages teach global reading order and relations among text, tables, formulas, and captions. Text crops preserve recognition resolution for small or densely packed characters; mixed-content regions retain local interactions, such as formulas with explanations or figures with captions.

Tables and formulas receive dedicated supervision for their structural requirements. HTML tables encode cell content and structure, including row and column spans. LaTeX formulas encode symbols, operator hierarchy, and grouping. Component records provide more training on these structures.

Supervision conversion uses each source's most informative reliable annotation. Verified page elements and reading order yield a Markdown target and, where valid, corresponding text, mixed-region, table, or formula crops. Sources containing only table or formula annotations use the corresponding component representation without invented page context. Crops receive only visible-content targets, normalized consistently with their page-level counterparts. Source and conversion identifiers link derived records, keeping page, region, and component targets consistent and limiting repeated exposure from one page.

Before training-specific sampling, roughly 40\% of records provide full-page supervision, 25\% text or mixed regions, 25\% isolated tables, and 10\% isolated formulas.

\paragraph{Diverse document layouts.}
Layout coverage includes single- and multi-column pages, changing column counts, sidebars, floating elements, footnotes, and forms with field-based reading order. Handwritten notes, posters, and presentation material add layouts without a regular grid. These examples require local grouping and reading-order inference beyond a fixed top-to-bottom template.

Density and scale vary within each layout family, from sparse title pages to dense academic or legal documents. Tables may occupy a full page or be embedded in prose. Layout labels and component statistics support coverage analysis and batching, while all inputs retain the shared output representation.

\paragraph{Image sources and languages.}
Acquisition coverage spans born-digital pages, scans, photographs, screenshots, and screen recaptures, introducing variations in geometry, illumination, resolution, noise, and compression. Languages include Simplified Chinese, English, Traditional Chinese, and mixed-language pages.

Quality checks remove corrupt or unreadable images and route uncertain targets to annotation review. Image hashes detect exact duplicates; normalized visual embeddings identify near duplicates under source-aware thresholds. Deduplication precedes supervision conversion, and derived views retain their duplicate-group identity so repeated content does not overstate data coverage or dominate training.

These attributes guide source-aware sampling and later residual analysis.

\paragraph{Source allocation and token accounting.}
\label{sec:source-allocation}
To retain source diversity without letting large sources dominate, we balance sources using their available supervised-token counts. Let $s$ index data sources and $\mathcal I_s$ contain the eligible record indices from source $s$ after filtering and duplicate control. Using $L_i$ from Section~\ref{sec:formulation}, we define
\begin{equation}
  M_s=\sum_{i\in\mathcal I_s}L_i,
  \qquad
  \omega_s=\frac{M_s^{\rho}}{\sum_{s'}M_{s'}^{\rho}},
  \quad 0\leq\rho\leq1.
  \label{eq:source-mixture}
\end{equation}
where $M_s$ counts the supervised tokens in eligible records once, and $\omega_s$ is the desired source token share. The index $s'$ runs over nonempty sources. Setting $\rho=1$ follows available token counts; $\rho=0$ gives equal shares; intermediate values reduce the influence of large sources.

Token shares require length correction to become record-sampling probabilities. A stream is a subset sampled for a particular role, such as clean pages or hard examples. For $G$ nonempty streams indexed by $g$, with $h$ a summation index, let $q^{(g)}(i)$ be the within-stream record probability, $\bar L_g=\sum_iq^{(g)}(i)L_i$ its mean target length, and $v_g\geq0$ its desired token share, with $\sum_gv_g=1$. The resulting record-sampling distribution is
\begin{equation}
  q(i)=\sum_{g=1}^{G}\alpha_g q^{(g)}(i),
  \qquad
  \alpha_g=\frac{v_g/\bar L_g}{\sum_{h=1}^{G}v_h/\bar L_h}.
  \label{eq:token-mixture-realization}
\end{equation}
The stream-selection probability $\alpha_g$ yields expected token share $\alpha_g\bar L_g/\sum_{h=1}^{G}\alpha_h\bar L_h=v_g$. Thus a stream of shorter targets requires more record draws to supply the same token share.

For source balancing, stream $g$ corresponds to source $s$, with $v_g=\omega_s$; the resulting sampler is $q_{\mathrm{clean}}$. Records have one sampling role per mixture. Duplicate-group caps limit repeated exposure, and actual token shares are monitored against planned shares.

\subsubsection{Unified Expert Annotation and Difficulty Assessment}
\label{sec:expert-annotation}
\label{sec:data-quality}
\label{sec:data-difficulty}

The Stage~I pool combines converted source labels with model-generated annotation. Both are mapped to common element categories and representations. Categories identify content such as text, headings, tables, and formulas; page and region instead describe how much of a document the input contains. Text symbols, whitespace, and equivalent markup are normalized consistently, with shared HTML-table, LaTeX-formula, and reading-order Markdown conventions.

When source labels are incomplete, inconsistent, or unavailable, we obtain candidate targets from three expert systems. The expert parsers are MinerU2.5-Pro~\cite{wang2026mineru2}, PaddleOCR-VL-1.6~\cite{zhang2026paddleocr}, and dots.mocr~\cite{zheng2026multimodal}. The multi-model labeling branch in Figure~\ref{fig:stage1-workflow} shows how normalized agreement determines whether a target enters Stage~I directly, requires review, or becomes a difficult candidate for Stage~II.

\paragraph{Multi-model joint annotation and difficulty control.}
Expert outputs first undergo common normalization and representation-specific syntax checks. Let $m,n$ index expert parsers and $\widetilde y_i^{(m)},\widetilde y_i^{(n)}$ be their normalized predictions for image $x_i$: candidate annotations for $y_i$, not verified references. Their normalized edit distance (NED) is
\begin{equation}
  d_i^{(m,n)}=
  \frac{\operatorname{ED}(\widetilde y_i^{(m)},\widetilde y_i^{(n)})}
  {\max\{1,|\widetilde y_i^{(m)}|,|\widetilde y_i^{(n)}|\}}.
  \label{eq:expert-ned}
\end{equation}
Here $\operatorname{ED}$ is character-level Levenshtein distance and $|\cdot|$ counts characters. High-confidence annotation requires predictions from all three experts, with every pair's NED below the annotation-agreement threshold $\delta_{\mathrm{ann}}$ ($\mathrm{ann}$ denotes annotation).

Agreement is combined with HTML/LaTeX parseability and structural checks. Converted source ground truth remains preferred when valid; otherwise a valid representative of agreeing expert outputs is selected.

If two experts agree and the third differs, the agreed output requires image-based review before inclusion in training. If all three differ, the record is withheld from ordinary Stage~I supervision and reviewed by a stronger multimodal model and human reviewers for possible Stage~II use. Missing predictions or unresolved disagreements likewise require review: agreement alone does not establish correctness.

Confidence concerns target reliability; difficulty concerns visible content, characterized by structural complexity, component density, and expert disagreement. Reliable records are stratified into Easy, Medium, and Hard levels according to these attributes; the strata guide Stage~I sampling and Stage~II hard-example mining. Corrupt images, invalid markup, incorrect reading order, severe repetition, and duplicates are filtered or reviewed. Hashes, visual embeddings, recognized content, and layout support duplicate control. Unresolved image--target disagreements enter the audit in Section~\ref{sec:hard-refinement}.

\subsubsection{Broad-Coverage Data Synthesis}
\label{sec:content-synthesis}

Broad-coverage synthesis creates executable page programs---structured specifications that produce both a page image and its target---for combinations rarely found in collected data. Content and structure are generated before appearance augmentation in Section~\ref{sec:degradation}. Related work also uses compositional synthesis for controllable full-page supervision~\cite{li2026towards}.

\paragraph{Compositional document programs.}
A page program specifies content dependencies, language, layout, reading order, elements, and visual style. Its source and random state permit regeneration. Joint semantic and rendering constraints require consistent table structure, formulas compatible with their explanations, and visible content.

Source and structural metadata identify underrepresented combinations after duplicate control. The annotation review in Section~\ref{sec:expert-annotation} identifies uncertain targets, while generator-family limits prevent a few templates from dominating.

Programs establish semantic relations before layout: claims connect to evidence, line items to totals, and concepts to equations. These relations determine headings, tables, caption attachment, and cross-region reading order. Content and style vary within the resulting constraints.

\paragraph{Dual compilation from semantic HTML.}
Semantic HTML is the executable intermediate representation. Each block becomes a Document Object Model (DOM) node with an element type and unique identifier; CSS controls its visual layout. One compilation path renders the image and node geometry, while another traverses the same tree to produce element labels, reading-order Markdown, HTML tables, and source LaTeX. Shared identifiers align image regions and structured targets, yielding exact supervision from the generating source.

\begin{figure}[htbp]
\centering
\includegraphics[width=\linewidth]{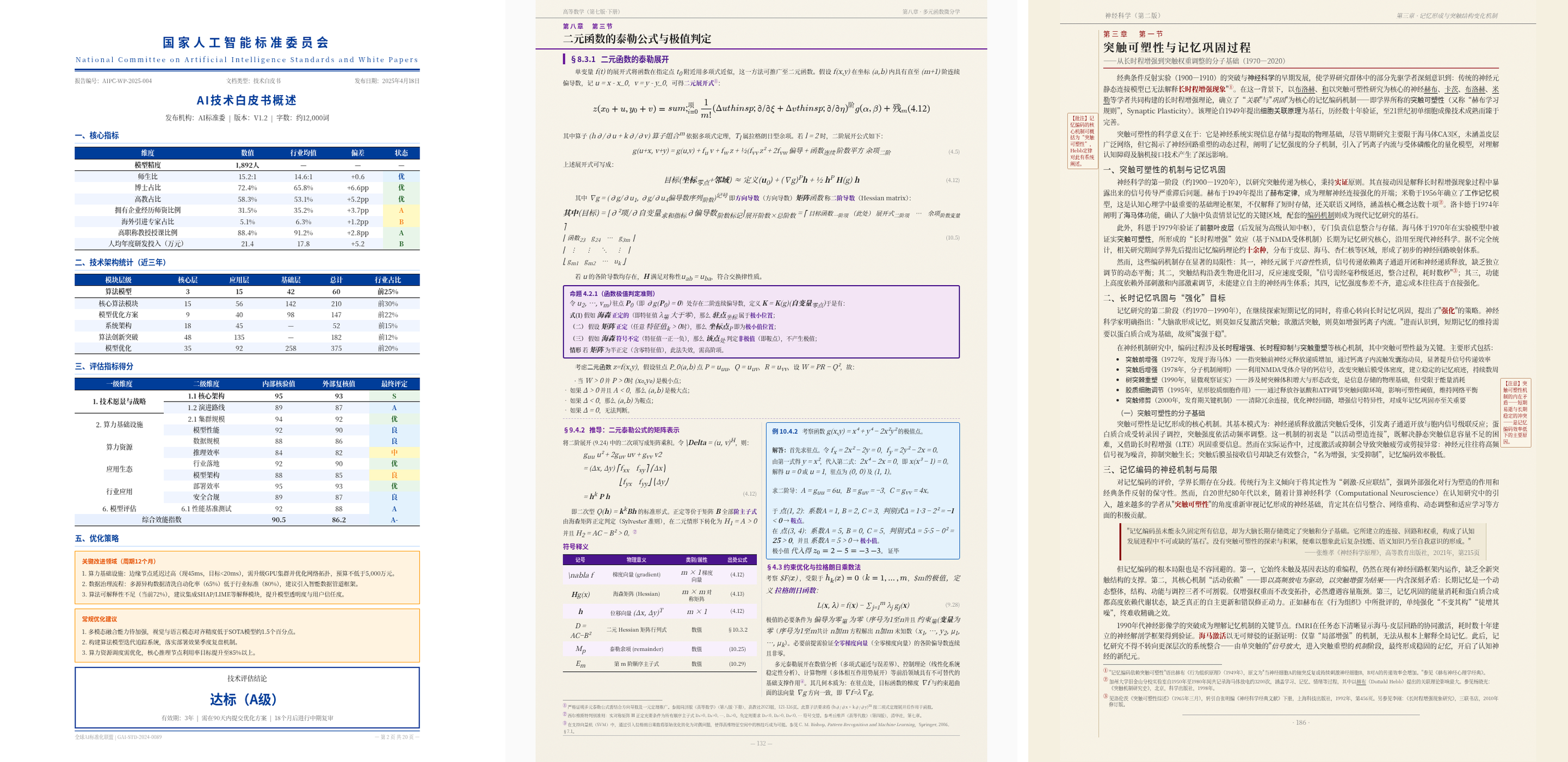}
\caption{Representative pages produced by general data synthesis. The gallery spans a structured technical report, an equation-rich educational page, and a long-form scholarly page, illustrating joint variation in semantic context, element composition, page density, typography, and layout organization.}
\label{fig:general-synthesis-grid}
\end{figure}

Generated pages undergo the same normalization, validation, and duplicate control as collected data. Failed rendering, clipped content, invalid tables or formulas, and unreadable text are rejected. Complexity and expert disagreement characterize difficulty, with program-level caps preserving diversity. Figure~\ref{fig:general-synthesis-grid} shows representative outputs.

\subsubsection{Appearance Degradation and Augmentation}
\label{sec:degradation}

Appearance augmentation follows source-specific acquisition paths: born-digital pages may be resized or recaptured, paper documents printed or scanned, and historical material affected by aging. Applied to collected pages and the clean synthesis in Section~\ref{sec:content-synthesis}, it expands acquisition coverage while retaining valid semantic targets. We group compatible degradation operators by acquisition path into operator families, each specifying the operation order and parameter ranges. This organization draws on document-image augmentation and source-aware robustness comparisons~\cite{groleau2023augraphy,li2026far}.

An acquisition scenario combines compatible operator families, their parameters, and their physical order. For example, a photocopied photograph passes through paper and ink formation, device reproduction, and camera capture. Source media constrain valid paths; content density, symbol size, formula strokes, and table-rule width constrain severity.

A transformed view retains its source target only while all supervised content remains visible and readable. Views that lose content are rejected or assigned visible-content crop targets following Section~\ref{sec:supervision-granularities}.

\paragraph{Paper, copying, and transmission effects.}
Paper texture, aging, stains, and ink fading or bleeding simulate changes to the paper and print. Printing, photocopying, and scanning introduce halftone patterns, blur, stripes, and contrast loss, while digital transmission adds resizing, re-encoding, and compression artifacts. Effects are scaled to resolution and stroke width, preserving relevant differences among text, formulas, and table rules.

\paragraph{Physical acquisition.}
Surface maps and constrained meshes vary perspective, page curvature, and creases, following document-unwarping representations~\cite{das2019dewarpnet,verhoeven2023uvdoc}. Illumination follows the sampled geometry to produce spatially consistent shadows and exposure variation. Screen recapture models the display-to-camera sampling chain, producing moir\'e and colored edges; camera effects also include motion or depth-dependent blur.

\begin{figure}[htbp]
\centering
\begin{minipage}[t]{.49\linewidth}
\centering
\includegraphics[width=\linewidth,height=.20\textheight,keepaspectratio]{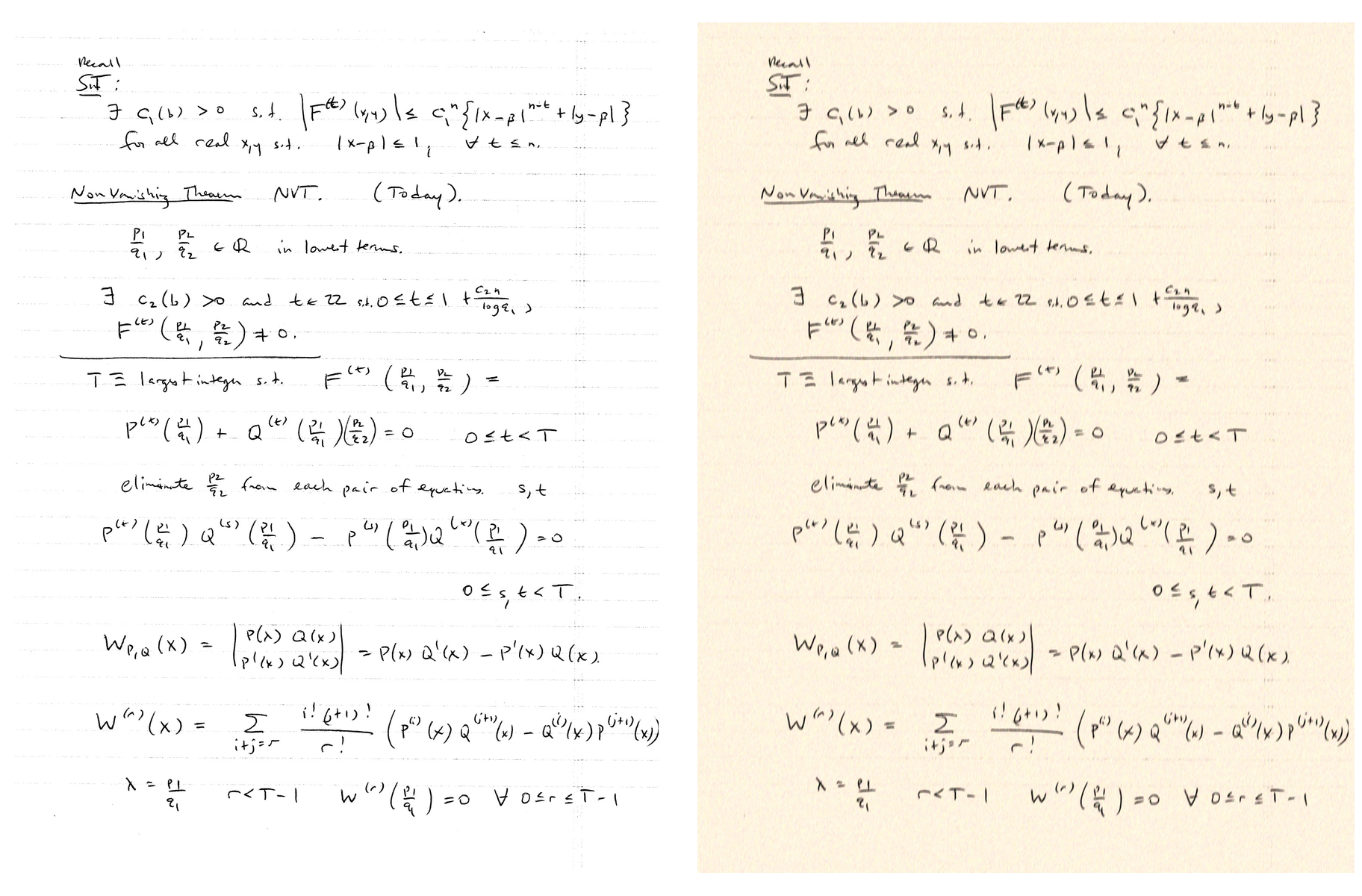}\par
\vspace{.5mm}\scriptsize (a) Paper tone, texture, and signal noise
\end{minipage}\hfill
\begin{minipage}[t]{.49\linewidth}
\centering
\includegraphics[width=\linewidth,height=.20\textheight,keepaspectratio]{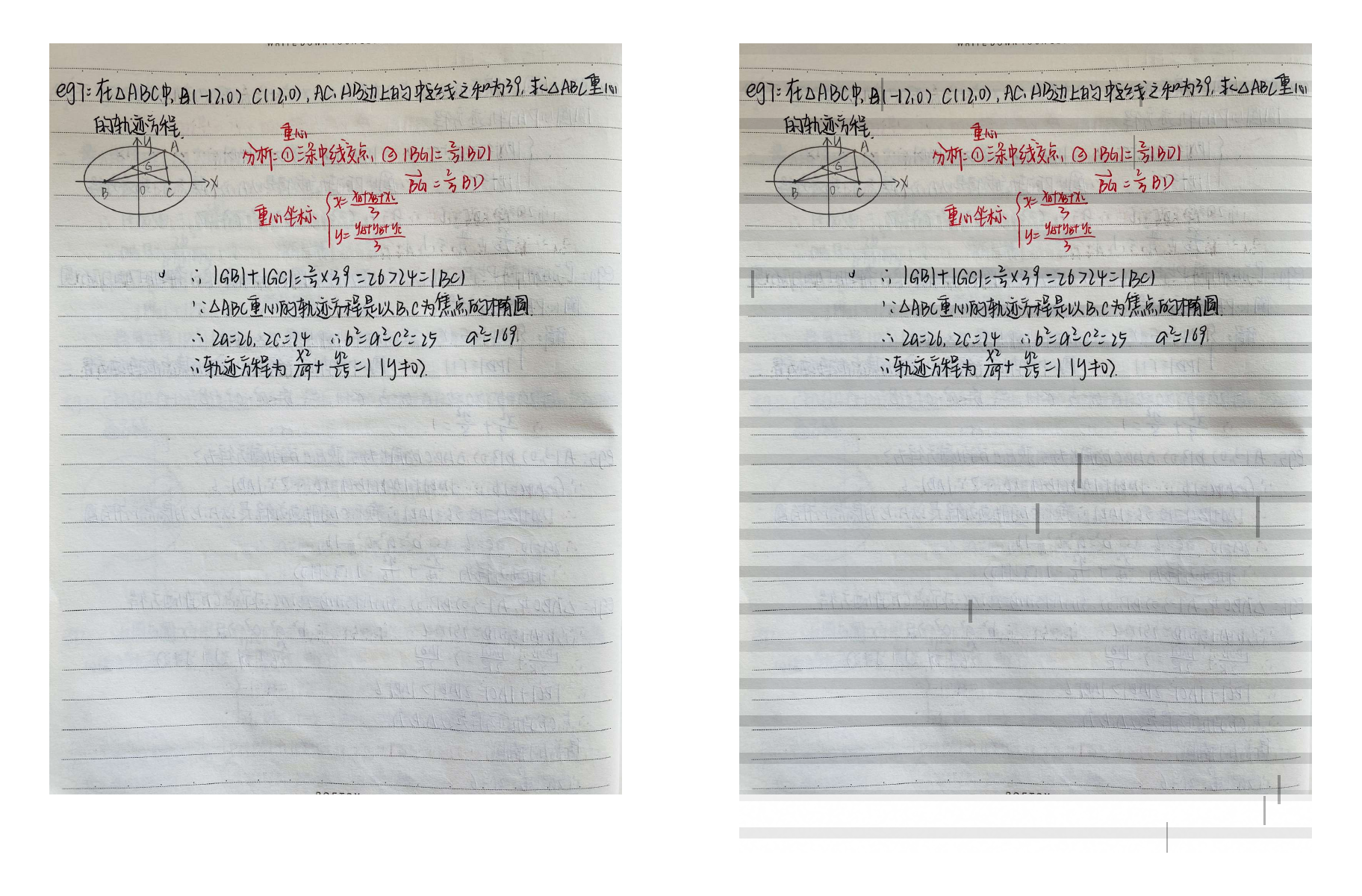}\par
\vspace{.5mm}\scriptsize (b) Scan bands and partial signal dropout
\end{minipage}
\vspace{2mm}

\begin{minipage}[t]{.49\linewidth}
\centering
\includegraphics[width=\linewidth,height=.20\textheight,keepaspectratio]{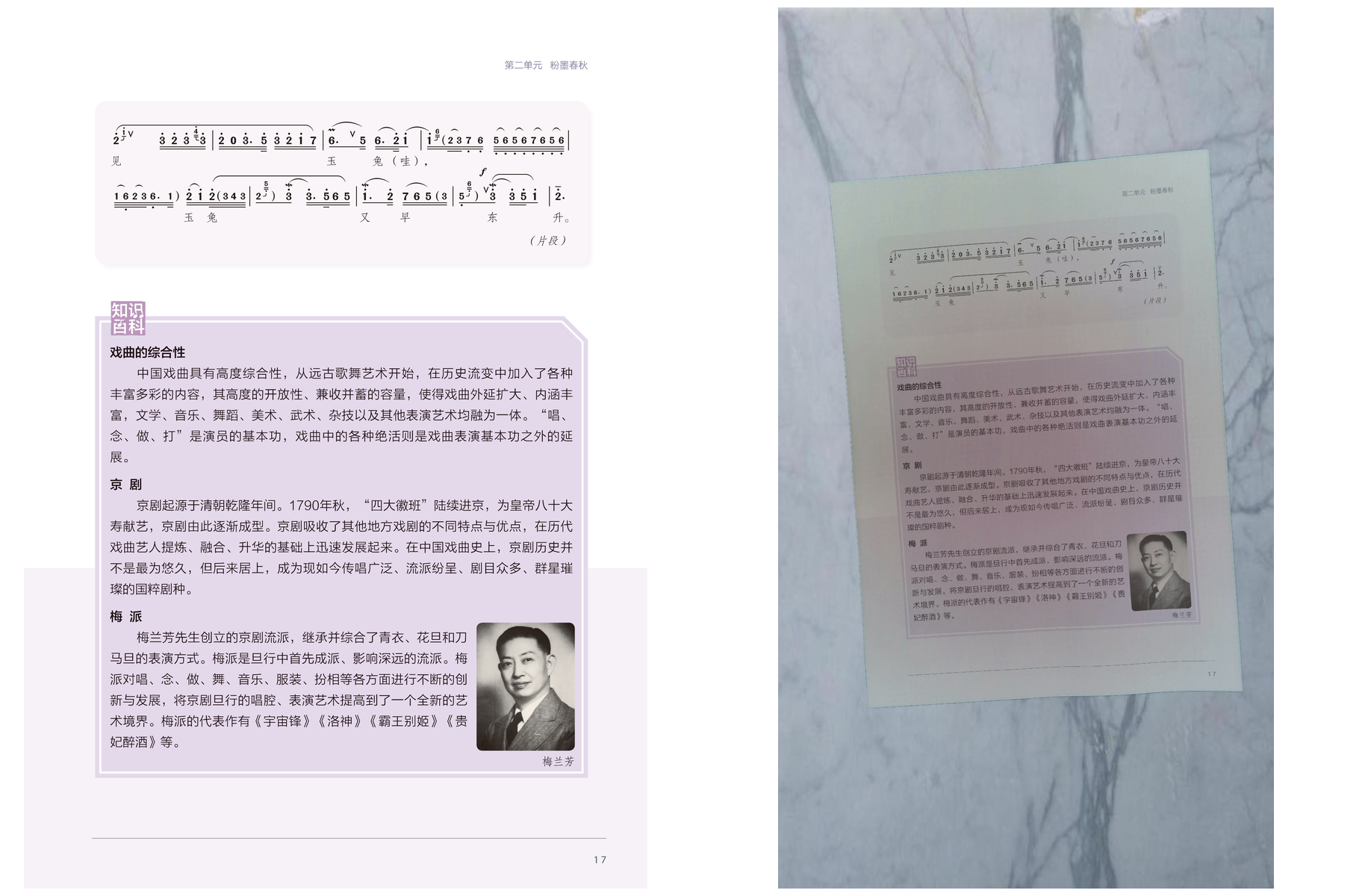}\par
\vspace{.5mm}\scriptsize (c) Perspective and handheld capture
\end{minipage}\hfill
\begin{minipage}[t]{.49\linewidth}
\centering
\includegraphics[width=\linewidth,height=.20\textheight,keepaspectratio]{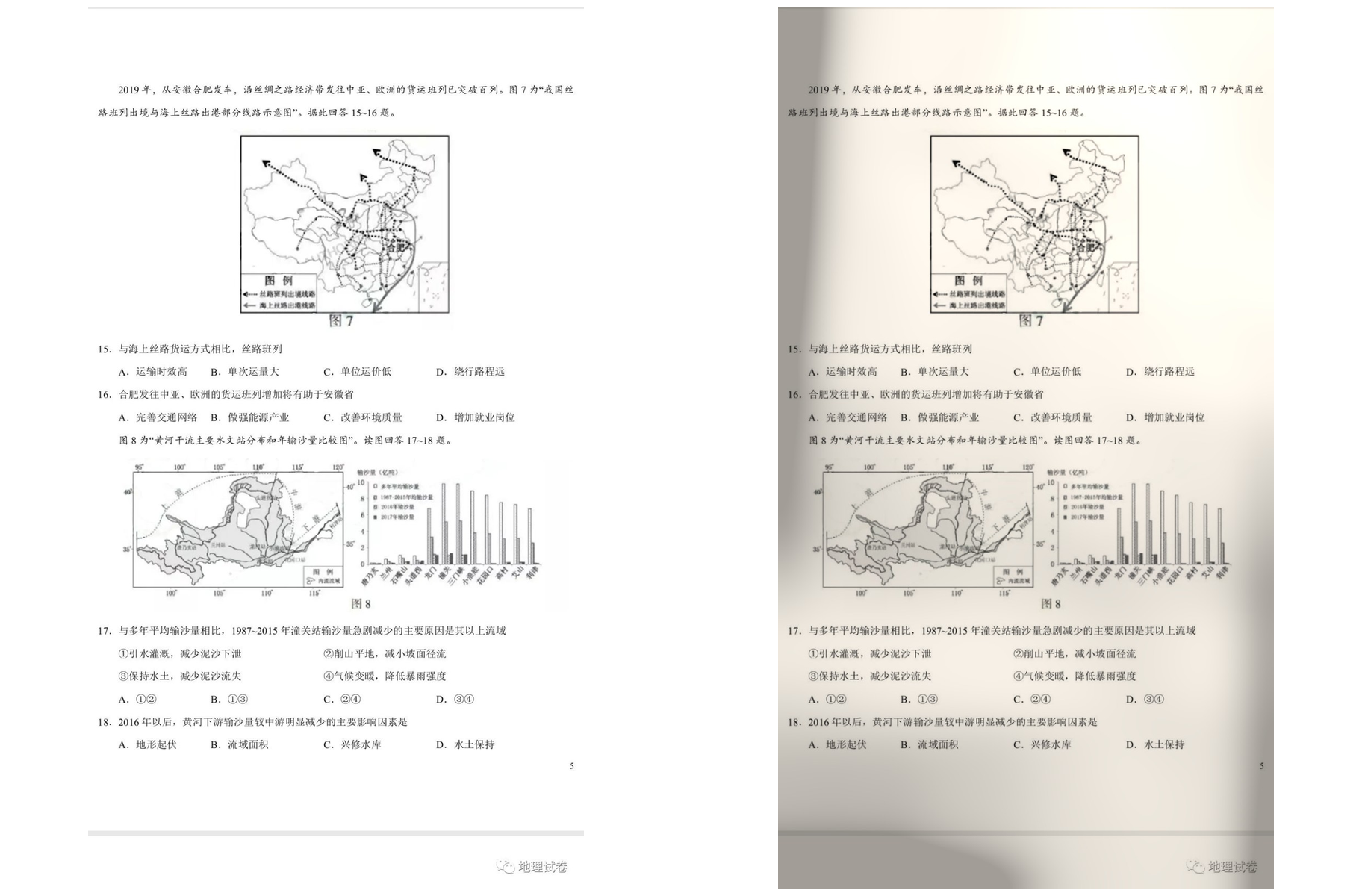}\par
\vspace{.5mm}\scriptsize (d) Curvature and uneven illumination
\end{minipage}
\caption{Representative appearance transformations. Each cell shows the source page on the left and the transformed view on the right: (a--b) paper, copying, and transmission effects, and (c--d) physical acquisition effects.}
\label{fig:appearance-degradation-gallery}
\end{figure}

\paragraph{Difficulty and observability control.}
Content retention and local legibility determine admission; expert disagreement indicates difficulty but does not establish target validity. Variants retain their clean page's duplicate-group identity, so semantic-support statistics count the source once while sampling controls variant exposure. Figure~\ref{fig:appearance-degradation-gallery} illustrates the transformations.

Stage~I combines the clean-data sampler $q_{\mathrm{clean}}$ from Section~\ref{sec:source-allocation} with validated appearance variants. The degraded stream samples a clean record and a transformation that preserves its target. The desired degraded-token share is selected by internal validation; Equation~\ref{eq:token-mixture-realization} converts the clean and degraded shares into record-selection probabilities, defining $q_1$. Duplicate-group caps limit repeated views, and actual token exposure is monitored separately from record draws.

The resulting parser $\theta_1$ supports the Stage~II diagnosis of whether remaining weaknesses require replay, new collection, or targeted synthesis.

\subsection{Stage~II: Capability Diagnosis and Targeted Refinement}
\label{sec:stage2-allocation}
\label{sec:capability-handoff}

Stage~II constructs an approximately 5M-record refinement pool around the weaknesses of $\theta_1$. \emph{Residual error} is parsing error remaining after Stage~I; \emph{replay} is continued training on high-quality Stage~I examples. We call a record high-error when its page- or component-level error exceeds the review threshold; because this may reflect annotation errors or unreadable content, refinement verifies target reliability before treating it as a hard example.

\begin{figure}[htbp]
\centering
\includegraphics[width=0.98\linewidth]{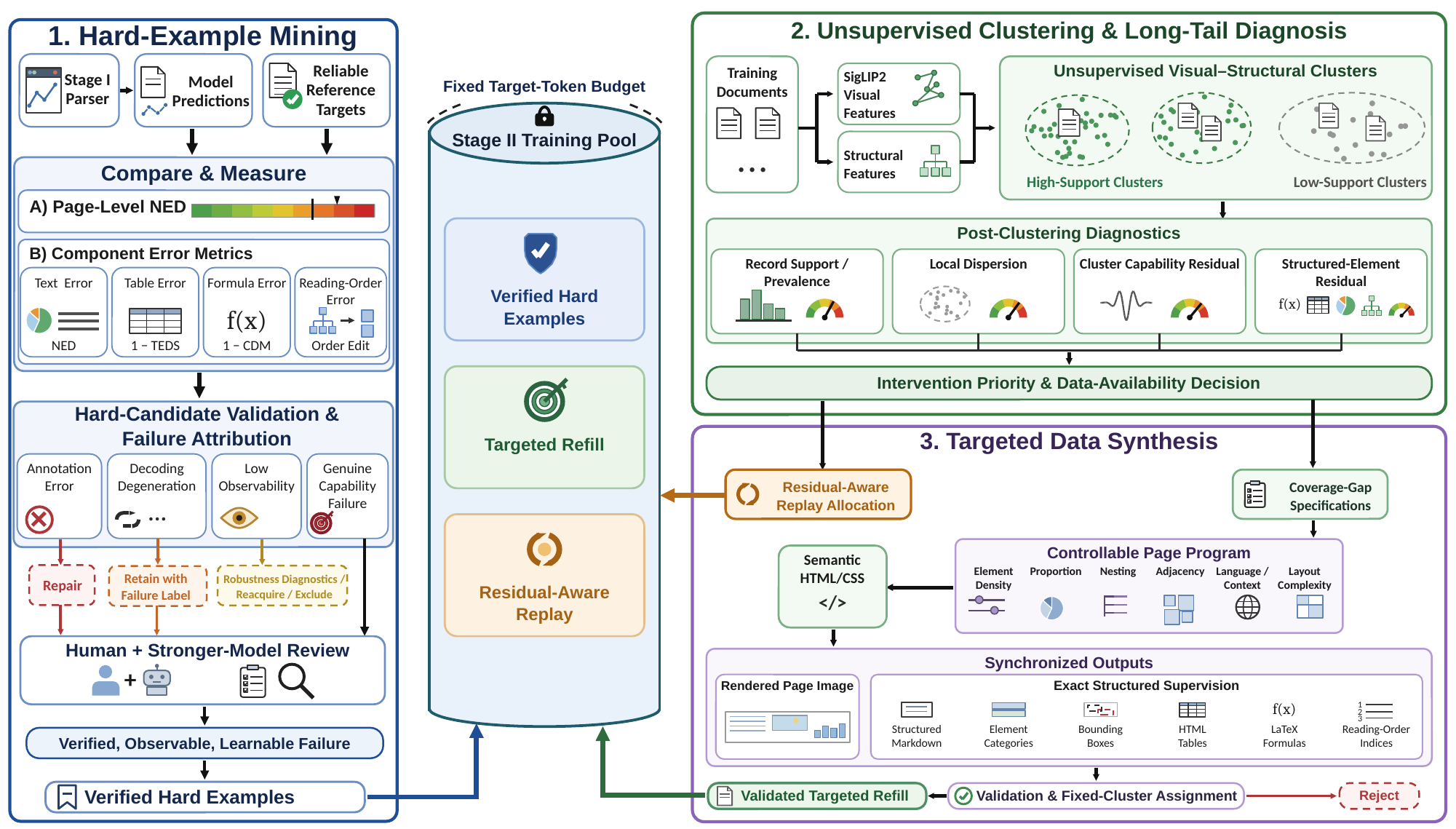}
\caption{Stage~II data construction. A disjoint allocation probe determines cluster-level residual priorities, while training-pool mining supplies a separately audited hard-example stream. Verified retrieval and targeted page-program synthesis then add examples for underrepresented patterns before training tokens are allocated according to residual errors under a fixed budget.}
\label{fig:stage2-workflow}
\end{figure}

Diagnosis combines document-level errors with component metrics that reveal failures occupying little page area. The photographed contract in Figure~\ref{fig:case-pdb-real}(b), for example, contains formulas missing from the Stage~I output despite readable surrounding text. Record-level mining selects verified hard examples; group-level analysis identifies recurring errors with insufficient data coverage. Figure~\ref{fig:stage2-workflow} links diagnosis (Section~\ref{sec:clustering}) to targeted construction (Section~\ref{sec:targeted-synthesis}) and allocation.

Each model scale is diagnosed with its frozen Stage~I checkpoint. Hard examples come from training-pool mining or separately audited annotation candidates. A held-out allocation probe, $\mathcal D_{\mathrm{probe}}$, estimates group-level weaknesses and is excluded from training, mining, and final benchmark evaluation. Pages and derived crops stay within the same content split. Targets must pass image--target correspondence, format, and structural checks; Sections~\ref{sec:hard-refinement} and~\ref{sec:clustering} describe verification and group-level aggregation.

\subsubsection{Hard-Example Mining and Verification}
\label{sec:hard-refinement}
\label{sec:residual}

\paragraph{Document-level mining.}
Let $\mathcal D_{\mathrm{mine}}$ be the analysis sample from the Stage~I training pool, disjoint in content from $\mathcal D_{\mathrm{probe}}$. Annotation candidates withheld in Section~\ref{sec:expert-annotation} follow a separate verification queue; they are not Stage~I training records. Both routes exclude probe and benchmark content.

For a full-page image $x_i$, let $\widehat y_i$ be the Stage~I prediction and $y_i$ the reliable full-page target. Both are reading-order Markdown strings containing text, HTML tables, and LaTeX formulas. Their complete-page error is $d_{i,\mathrm{page}}=\operatorname{NED}(\widehat y_i,y_i)$, using Equation~\ref{eq:expert-ned}. This compares complete serialized content and reading order, not an average of element scores. The score selects review candidates rather than certifying hard examples, and scores are recomputed for each Stage~I checkpoint.

The representation in Section~\ref{sec:clustering} groups related failures for review. Training-pool scores select hard examples; held-out probe errors determine allocation across document groups.

\paragraph{Component-level mining.}
Whole-page NED includes tables and formulas, but a small failed region can contribute little to the page score. We therefore examine text, tables, and formulas separately; figures and charts are excluded from these component metrics. For record $i$, let $c$ denote a component type and let $y_{i,c}$ and $\widehat y_{i,c}$ denote its reference and predicted content. The corresponding error is
\begin{equation}
  e_{i,c}=\begin{cases}
    \operatorname{NED}(\widehat y_{i,c},y_{i,c}), & c=\mathrm{text},\\
    1-\operatorname{TEDS}_{[0,1]}(\widehat y_{i,c},y_{i,c}), & c=\mathrm{table},\\
    1-\operatorname{CDM}_{[0,1]}(\widehat y_{i,c},y_{i,c}), & c=\mathrm{formula}.
  \end{cases}
  \label{eq:component-difficulty}
\end{equation}
The subscript $[0,1]$ indicates a similarity score expressed on the unit scale. Tree-Edit-Distance-based Similarity (TEDS)~\cite{zhong2020image} measures table content and tree structure, while Character Detection Matching (CDM)~\cite{wang2025image} compares rendered formulas through character detection and matching. Subtracting either similarity from one makes all three metrics increase with error. Component correspondence and aggregation follow a consistent type-specific evaluation protocol. Reading-order consistency is evaluated with the serialized page target, while Equation~\ref{eq:component-difficulty} characterizes localized errors for text, tables, and formulas.

Separate page and component thresholds retrieve review candidates; they do not certify target correctness. For allocation, $e_i(\theta_1)\in[0,1]$ summarizes record $i$'s available page and component errors under checkpoint $\theta_1$. The scoring protocol is fixed across groups and applied to the held-out probe. Component channels remain available separately to reveal weaknesses hidden by the combined score.

\paragraph{High-error audit.}
High-error candidates are checked for three other possible causes of error. Incorrect targets may omit content, misorder regions, or contain invalid markup. OCR-EDR's rendering-aware diagnosis and repair uses DocEDR to compare the image, editable target, and rendered output~\cite{zhao2026ocredr}; repairs must remain structurally valid and visually consistent. Repetitive or outputs that fail to stop receive a decoding-failure label. Content that cannot be read because of cropping, blur, occlusion, or low resolution is excluded from hard-example training but may still help assess robustness.

Human reviewers and a stronger multimodal model verify difficult image--target pairs and structured outputs. The model supplies a second interpretation, not automatic ground truth; unresolved disagreements require review or exclusion. Accepted records retain scores, checkpoint, error type, verification route, and target revision.

Let $\mathcal H$ be the verified hard set. A record is included only if the mining checkpoint still makes substantial errors against the final reliable target. If repair removes the apparent error, the record may enter ordinary supervision but not hard-example oversampling. Duplicate and exposure caps limit repeated templates and decoding failures.

\subsubsection{Visual--Structural Clustering and Coverage Diagnosis}
\label{sec:targeted-acquisition}
\label{sec:clustering}

Clustering groups Stage~I records by visual structure and component composition to reveal recurring failures. These groups are described after fitting using content, language, layout, and acquisition metadata.

\paragraph{Visual--structural representation.}
Each clustering record is a complete page or supervised region, with features describing its actual input content. Let $h_i$ be the unit-length embedding of $x_i$ from a frozen SigLIP2 encoder~\cite{tschannen2025siglip}. Its clustering feature is
\begin{equation}
  z_i=\left[\operatorname{PCA}(h_i);\eta_{\mathrm{str}}\widetilde b_i\right].
  \label{eq:cluster-feature}
\end{equation}
Here PCA denotes principal component analysis, which reduces the visual embedding dimension; brackets and the semicolon denote concatenation. The normalized metadata vector $\widetilde b_i$ describes component composition, content density, language, layout, acquisition condition, and target length. The subscript $\mathrm{str}$ denotes structural metadata, whose weight $\eta_{\mathrm{str}}\geq0$ controls its contribution relative to the image embedding.

The transform is fitted on a Stage~I dataset after deduplication; records retain input-granularity metadata and source-page links. A cluster center is the mean feature of its fitting records. Features are grouped by squared distance to centers, with groups split further when their features vary widely. After fitting, clusters are indexed by $k$ and their centers and transform are frozen. Training, probe, and new records are assigned to the nearest center without refitting.

\paragraph{Coverage and residual diagnosis.}
Before targeted additions, $p_k^{\mathrm{rec}}$ denotes the fraction of deduplicated fitting records assigned to cluster $k$; here, a record is a sampling unit that may be a page or a derived crop. Coverage priority combines the negative logarithm of the cluster's record fraction with its within-cluster feature dispersion. Small or diffuse clusters are flagged for coverage review, while residual error is measured separately on reliable probe records.

Errors from the held-out allocation probe provide complementary evidence. Let $\mathcal D_{\mathrm{probe},k}$ contain the probe records assigned to cluster $k$ whose image--target pairs pass the reliability checks in Section~\ref{sec:hard-refinement}. Their record errors $e_i(\theta_1)$ are defined in that section. Let $w_i>0$ be the fixed evaluation weight of record $i$. Equal weights summarize the observed probe; sampling-correction weights instead summarize a specified reference population. Let $\bar R$ denote the weighted mean error over all reliable probe records. For a nonempty group, its raw and stabilized residuals are
\begin{equation}
\begin{aligned}
  R_k&=\frac{\sum_{i\in\mathcal D_{\mathrm{probe},k}}w_i e_i(\theta_1)}
                  {\sum_{i\in\mathcal D_{\mathrm{probe},k}}w_i},\\
  R_k^{\mathrm{sh}}&=\gamma_k R_k+(1-\gamma_k)\bar R.
\end{aligned}
  \label{eq:capability-residual}
\end{equation}
The superscript $\mathrm{sh}$ denotes shrinkage toward the pooled mean. The support coefficient is $\gamma_k=n_k^{\mathrm{eff}}/(n_k^{\mathrm{eff}}+\lambda_{\mathrm{sh}})$, where $\lambda_{\mathrm{sh}}>0$ controls pooling strength and $n_k^{\mathrm{eff}}$ is the effective ($\mathrm{eff}$) probe count: the square of the sum of weights divided by the sum of squared weights within the group. With equal weights it is simply the record count. Small groups rely more heavily on the pooled estimate; empty groups use $\bar R$ as a prior, not as an observed group-specific error. At least one reliable probe record is required.

Let $g_k$ be the stabilized residual after subtracting its across-cluster mean, dividing by its across-cluster standard deviation with a positive numerical floor, and clipping to a fixed symmetric interval. A positive value indicates above-average estimated error. For a group without local probe evidence, this value reflects the pooled prior rather than a measured parsing weakness. Applying the same aggregation and stabilization separately to each component channel reveals concentrated text, table, or formula errors. Let $D_k^{\mathrm{elem}}$ be the largest positive standardized component residual, or zero when none is positive; $\mathrm{elem}$ denotes element-level error. Only channels with reliable probe support are used; prior-based estimates are distinguished from locally observed failures.

Priorities for data review combine standardized coverage priority, overall residual $g_k$, and component residual $D_k^{\mathrm{elem}}$ through a nonnegative weighted sum. The ranking guides review rather than automatic inclusion in training. New data require an audited failure pattern and insufficient existing supervision; groups with enough data but persistent errors may instead need more training. Rare groups without local probe evidence remain coverage-audit candidates.

\begin{figure}[htbp]
\centering
\begin{minipage}[t]{.49\linewidth}
\centering
\includegraphics[width=\linewidth,height=.39\textheight,keepaspectratio]{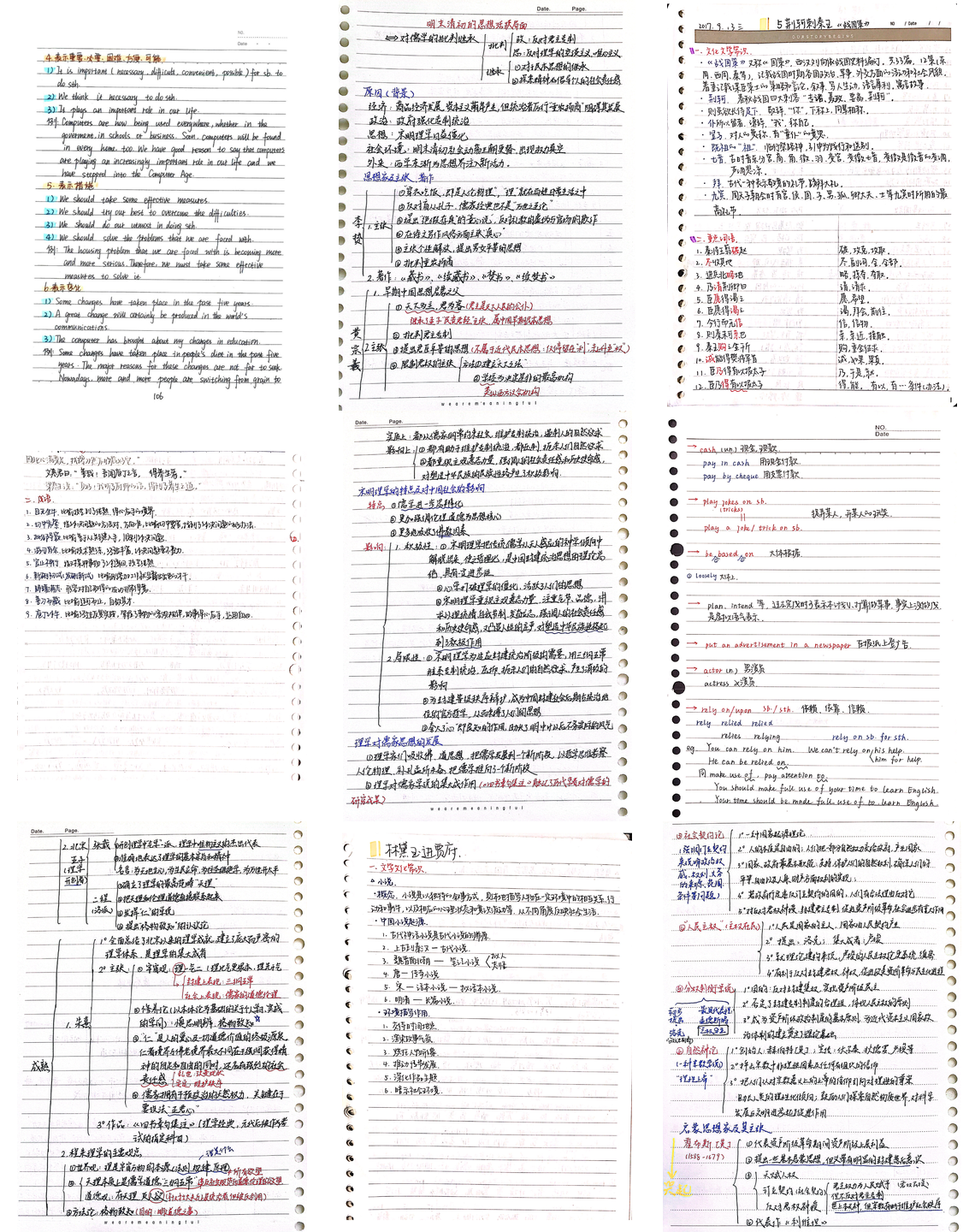}\par
\vspace{1mm}\small (a) Handwriting and notes
\end{minipage}\hfill
\begin{minipage}[t]{.49\linewidth}
\centering
\includegraphics[width=\linewidth,height=.39\textheight,keepaspectratio]{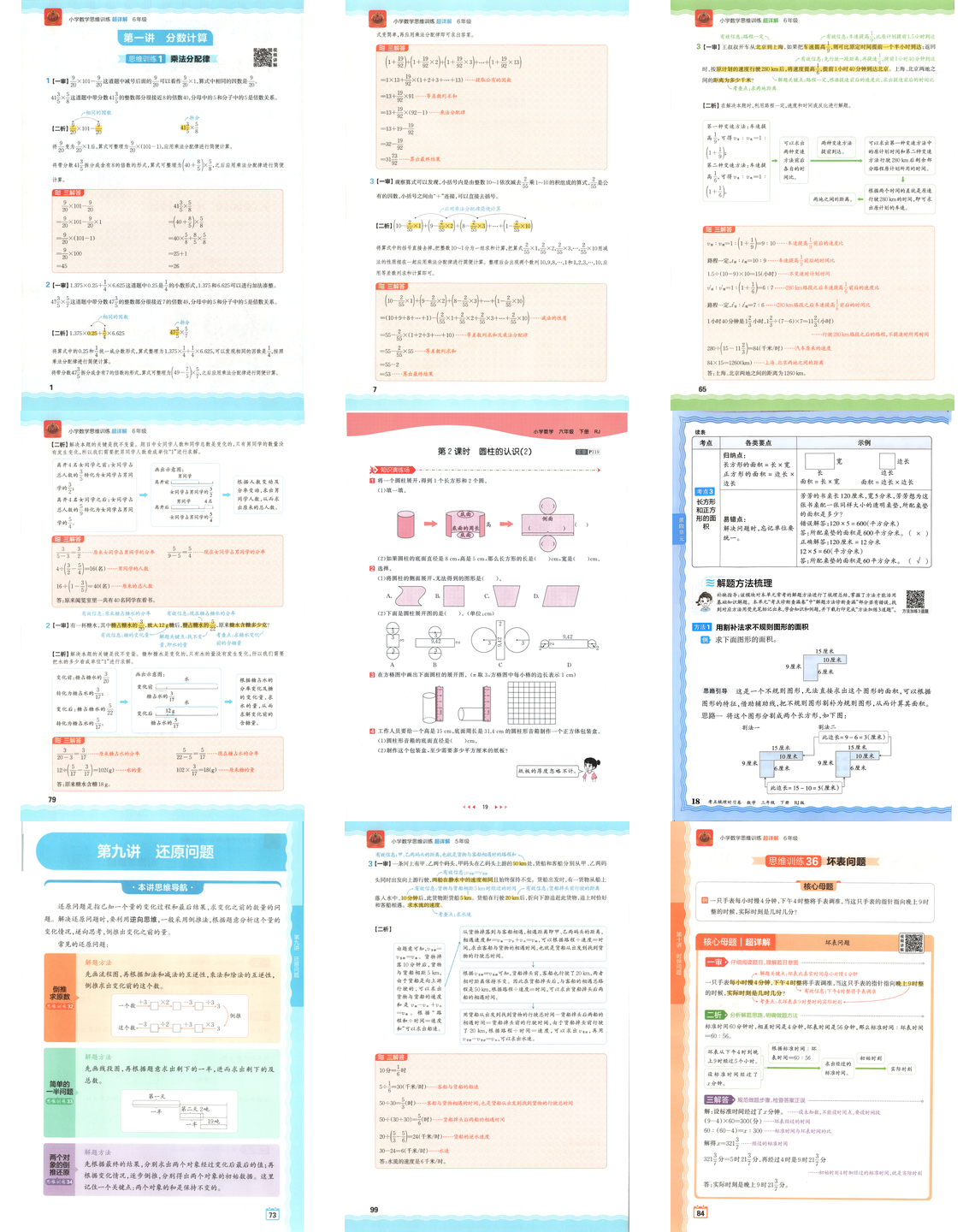}\par
\vspace{1mm}\small (b) Colorful textbooks
\end{minipage}
\caption{Representative groups obtained by global unsupervised clustering. Handwritten pages share ruled-paper texture, stroke organization, and free-form annotation patterns, whereas colorful textbooks are grouped by section frames, worked examples, exercise blocks, diagrams, and illustrated explanations.}
\label{fig:unsupervised-neighborhoods}
\end{figure}

Representative failures turn group priorities into data requests, such as Chinese mathematical text, nested tables, or cross-region reading relations. Requests are met through real-data retrieval or targeted synthesis (Section~\ref{sec:targeted-synthesis}), with additions passing supervision, reliability, and duplicate checks before reassignment. Figure~\ref{fig:unsupervised-neighborhoods} shows two example clusters.

\subsubsection{Residual-Driven Targeted Data Synthesis}
\label{sec:targeted-synthesis}

Targeted synthesis generates examples for verified requests from Section~\ref{sec:clustering} using the page-program family in Section~\ref{sec:content-synthesis}. It focuses on structures the model still parses poorly rather than broad coverage.

Requests specify failed element relations, layouts, or language combinations. Candidates are rendered, validated, and assigned using the frozen representation and cluster centers from Section~\ref{sec:clustering}; after rendering, each candidate is re-encoded and assigned to the nearest frozen cluster center. Priority guides construction, while content and style variation preserve diversity. When the generator cannot meet a request, new real data or an extended generator is needed, with the same validation checks.

\paragraph{Page programs for difficult element families.}
Formula programs combine inline notation, display equations, derivations, piecewise expressions, and explanations with consistent variable meanings, including Chinese mathematical text. Table programs vary borders, nested headers, merged cells, and mixed cell content. Cross-element programs connect captions, notes, references, and equations. Layout controls vary columns, density, adjacency, and reading order, while content and style diversity prevent repeated templates.

\paragraph{Co-generated supervision and cluster validation.}
The semantic-HTML compiler jointly produces the image, element regions and categories, reading-order Markdown, HTML tables, and source LaTeX. Examples are accepted only when their targets match the visible content, they contain the requested structure, and they address the identified weakness. Fixed-cluster assignment alone does not establish that the target difficulty has been reproduced.

\begin{figure}[htbp]
\centering
\begin{minipage}[t]{.49\linewidth}
\centering
\includegraphics[width=\linewidth,height=.19\textheight,keepaspectratio]{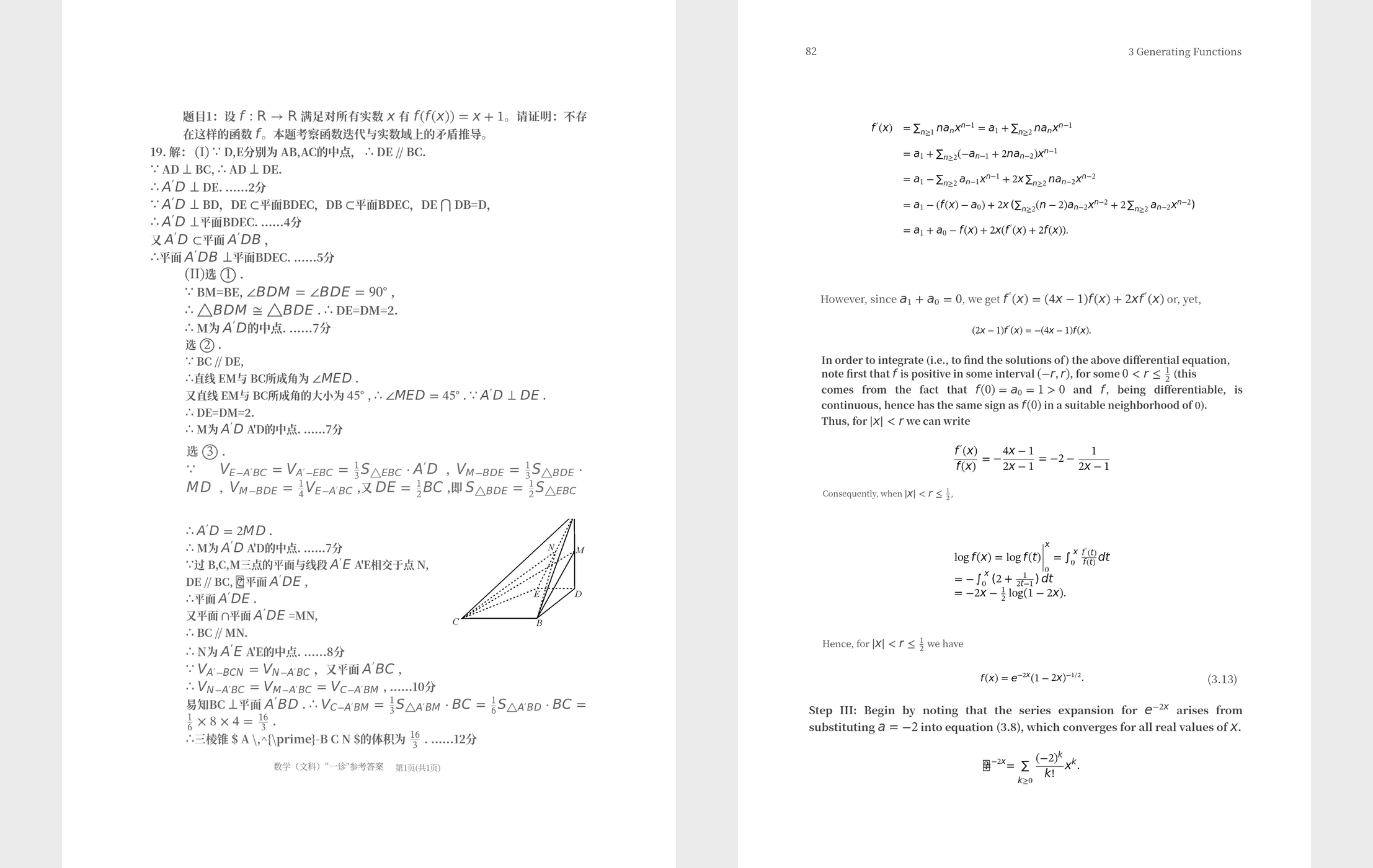}\par
\vspace{.5mm}\scriptsize (a) Formula-intensive pages
\end{minipage}\hfill
\begin{minipage}[t]{.49\linewidth}
\centering
\includegraphics[width=\linewidth,height=.19\textheight,keepaspectratio]{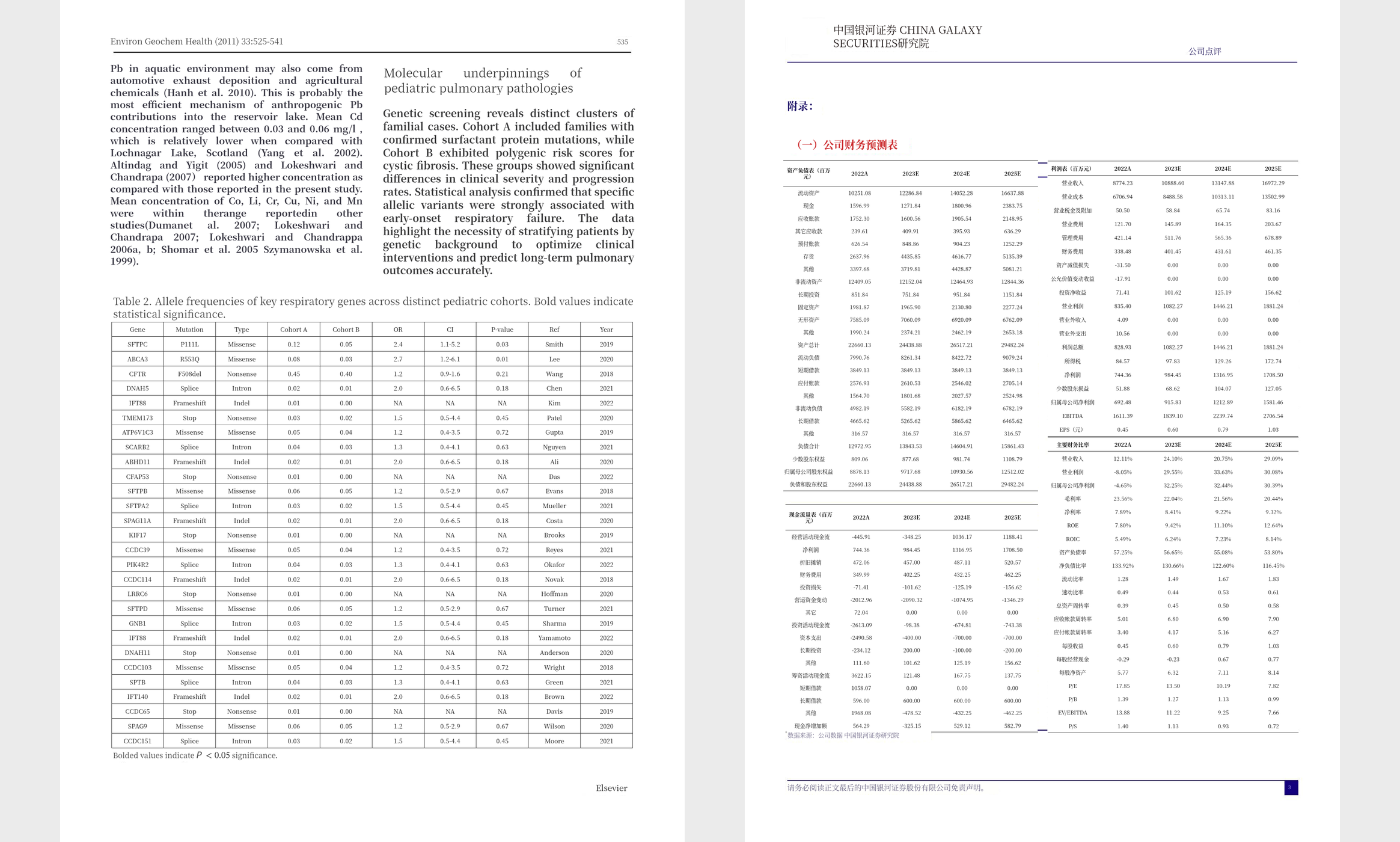}\par
\vspace{.5mm}\scriptsize (b) Table-intensive pages
\end{minipage}
\vspace{2mm}

\begin{minipage}[t]{.49\linewidth}
\centering
\includegraphics[width=\linewidth,height=.19\textheight,keepaspectratio]{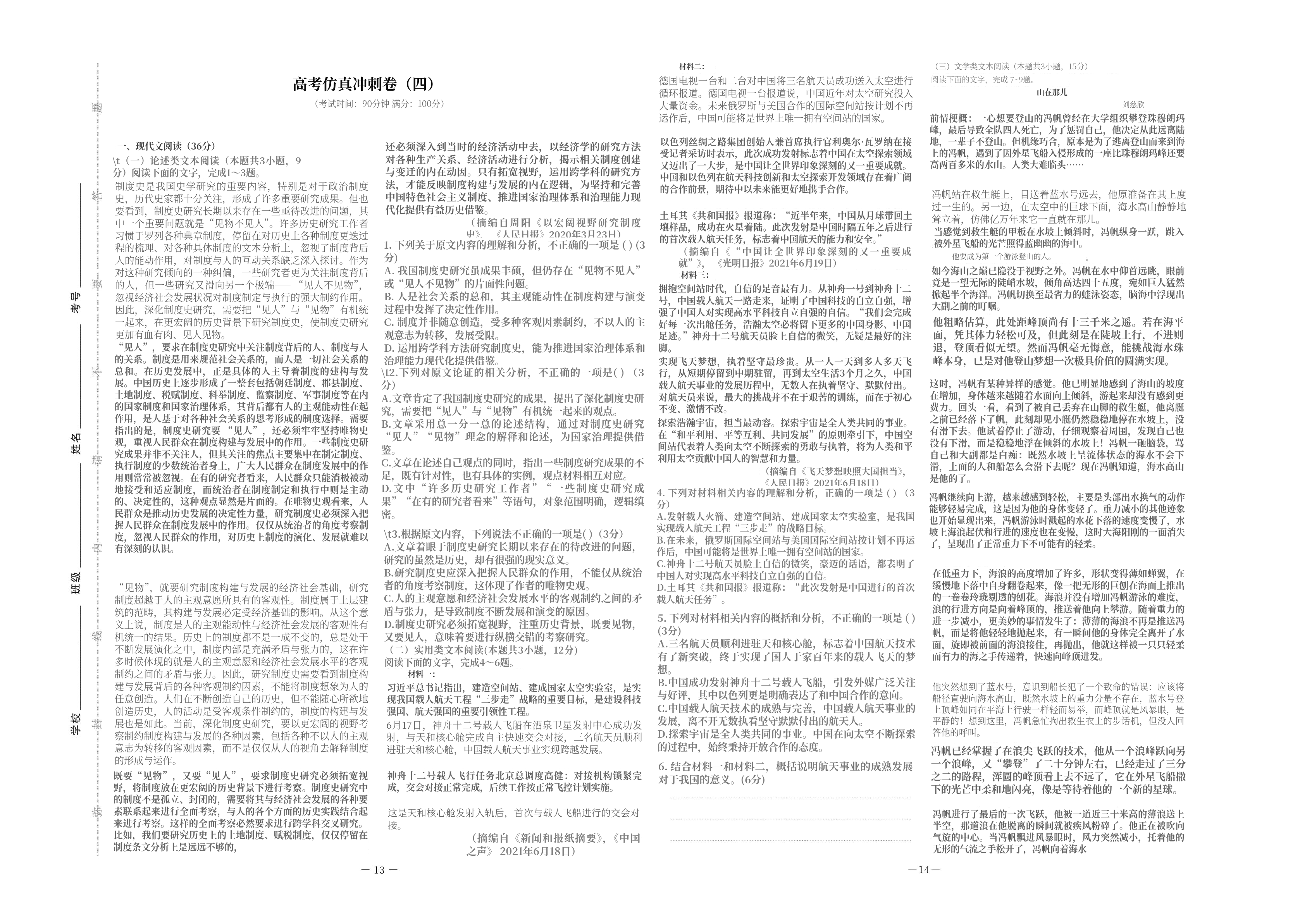}\par
\vspace{.5mm}\scriptsize (c) Dense examination layouts
\end{minipage}\hfill
\begin{minipage}[t]{.49\linewidth}
\centering
\includegraphics[width=\linewidth,height=.19\textheight,keepaspectratio]{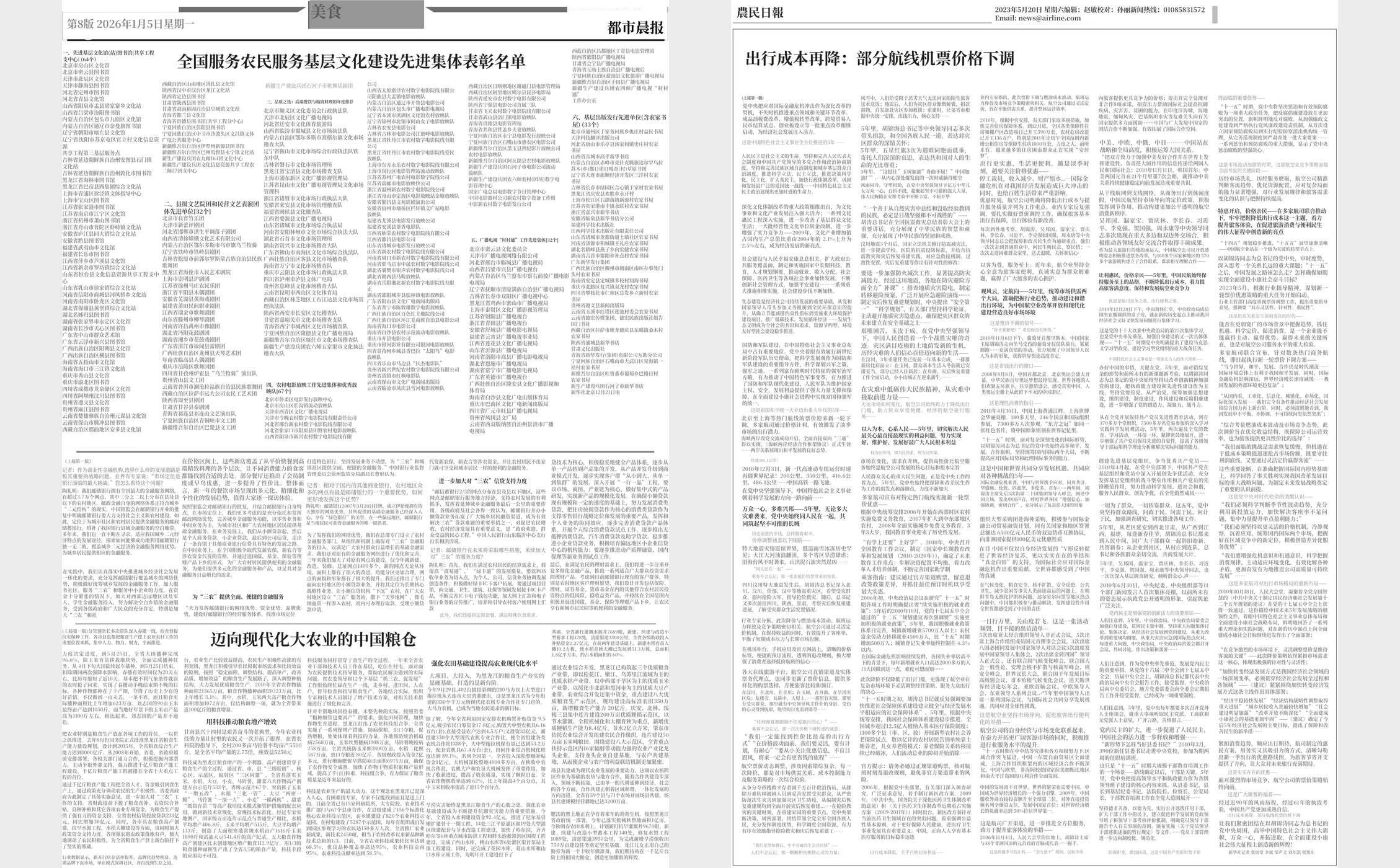}\par
\vspace{.5mm}\scriptsize (d) Dense newspaper layouts
\end{minipage}
\caption{Representative targeted constructions for diagnosed low-coverage capabilities: (a) formula-rich composition, (b) table-rich composition, (c) multi-region examination structure, and (d) dense newspaper structure.}
\label{fig:targeted-synthesis-gallery}
\end{figure}

Synthetic records retain source information and program identifiers, with family-, program-, and duplicate-group caps. Their contribution can therefore be distinguished from new real data, hard examples, and replay. Figure~\ref{fig:targeted-synthesis-gallery} illustrates representative constructions.

\subsubsection{Training Rebalancing after Targeted Data Addition}
\label{sec:rebalancing}

Validated additions retain their assignments to the fixed cluster centers. Let $\mathcal C_{\mathrm{res}}$ be the residual-stream pool: eligible Stage~I replay records, newly collected records, and validated targeted synthesis. Records assigned to the separately controlled hard stream are excluded from this pool for that sampling role.

For each cluster $k$, sum the effective target lengths $L_i$ of its records in $\mathcal C_{\mathrm{res}}$ and normalize by the pool total. The resulting fraction, $p_k^{\mathrm{tok}}$, is its natural token share after data addition ($\mathrm{tok}$ denotes tokens). It differs from the pre-addition record fraction $p_k^{\mathrm{rec}}$ used for coverage diagnosis. Let $\mathcal K_+$ be the set of clusters with a positive number of eligible target tokens. Empty clusters keep their identities and frozen centers but receive no allocation; they are not merged or refitted.

The proposed training token share combines the updated pool composition with the standardized residual $g_k$ defined in Section~\ref{sec:clustering}:
\begin{equation}
  \widetilde v_k=
  \frac{(p_k^{\mathrm{tok}})^{\alpha_{\mathrm{nat}}}
        \exp(\beta_{\mathrm{alloc}}g_k)}
       {\sum_{j\in\mathcal K_+}(p_j^{\mathrm{tok}})^{\alpha_{\mathrm{nat}}}
        \exp(\beta_{\mathrm{alloc}}g_j)},\qquad k\in\mathcal K_+.
  \label{eq:residual-mixture}
\end{equation}
Here $\widetilde v_k$ is the unconstrained token-share proposal. The nonnegative coefficients $\alpha_{\mathrm{nat}}$ and $\beta_{\mathrm{alloc}}$ weight the natural token share ($\mathrm{nat}$) and residual-driven allocation ($\mathrm{alloc}$), respectively. To prevent extreme changes, we choose the closest distribution satisfying relative exposure bounds:
\begin{equation}
\begin{aligned}
  \boldsymbol v^{\star}
  &=\underset{\boldsymbol v}{\arg\min}\,
    \sum_{k\in\mathcal K_+}v_k\log\frac{v_k}{\widetilde v_k},\\
  &\text{subject to }\quad \sum_{k\in\mathcal K_+}v_k=1,\qquad
    \vartheta_{\min}p_k^{\mathrm{tok}}\leq v_k\leq
    \vartheta_{\max}p_k^{\mathrm{tok}}.
\end{aligned}
  \label{eq:bounded-cluster-rebalancing}
\end{equation}
The vector $\boldsymbol v$ contains candidate token shares, and $\boldsymbol v^{\star}$ is the selected distribution. The objective is Kullback--Leibler divergence from the proposal. The lower and upper multipliers satisfy $0\leq\vartheta_{\min}\leq1\leq\vartheta_{\max}$, so the natural token-share vector is feasible. These bounds limit each cluster's training share, while Stage~I replay provides broad supervision to reduce forgetting.

Within the residual-stream pool ($\mathrm{pool}$), let $q_k^{\mathrm{pool}}(i)$ be the conditional record sampler for cluster $k$, after source, program-family, and duplicate-group exposure limits. Equation~\ref{eq:token-mixture-realization} converts $v_k^{\star}$ and the mean target lengths of these samplers into record-selection coefficients $\alpha_k^{\mathrm{pool}}$. Let $q_{\mathrm{hard}}$ sample verified hard records from $\mathcal H$. Its desired token share is $\lambda_{\mathrm{hard}}\in[0,1]$, and the same length correction gives the hard-stream selection probability $\alpha_{\mathrm{hard}}$. The two levels of sampling are
\begin{equation}
\begin{aligned}
  q_{\mathrm{residual}}(i)
    &=\sum_{k\in\mathcal K_+}\alpha_k^{\mathrm{pool}}q_k^{\mathrm{pool}}(i),\\
  q_2(i)&=(1-\alpha_{\mathrm{hard}})q_{\mathrm{residual}}(i)+\alpha_{\mathrm{hard}}q_{\mathrm{hard}}(i).
\end{aligned}
  \label{eq:hard-mixture}
\end{equation}
Here $q_{\mathrm{residual}}$ samples the rebalanced pool and $q_2$ is the Stage~II distribution introduced in Section~\ref{sec:formulation}. When both streams contribute, $\alpha_{\mathrm{hard}}$ equals $\lambda_{\mathrm{hard}}$ only if their mean effective target lengths are equal. Actual token shares are monitored separately from the planned shares.

Reallocation holds the Stage~II token budget $B_2$ fixed, not the cumulative training of both checkpoints. Stage-wise experiments assess the complete Stage~II training procedure rather than isolating rebalancing under equal total training budgets.

\begin{figure}[htbp]
\centering
\includegraphics[width=\linewidth]{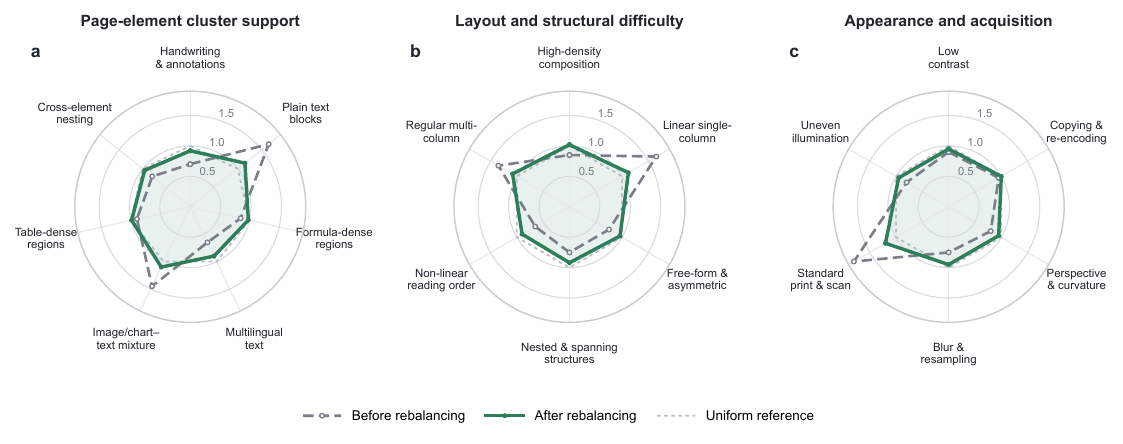}
\caption{Equal-budget cluster rebalancing. (a) Page-element cluster support, (b) layout and structural difficulty, and (c) appearance and acquisition conditions before and after residual-aware allocation. Gray dashed and green solid profiles denote the pre- and post-rebalancing mixtures, respectively; the light-gray unit contour denotes uniform support. Within each panel, expected target-token support is normalized to a mean of $1.0$, and the values in the two profiles have the same sum.}
\label{fig:latent-capability}
\end{figure}

The process combines diagnosis, targeted additions, fixed-center reassignment, and bounded rebalancing; Section~\ref{sec:optimization-protocol} describes how training exposure is recorded. Figure~\ref{fig:latent-capability} shows the allocation shifts across page elements, layouts, and acquisition conditions.

%% file: sections/03_experimental_instantiation.tex
\subsection{Experimental Setup and Controls}
\label{sec:model-setup}
\label{sec:experimental-instantiation}
\label{sec:method-config}

We use matched model and optimization settings for the two-stage method. Comparisons within each model scale assess complete training protocols rather than isolating the effects of individual data changes.

\subsubsection{Training Stages and Data Allocation}
\label{sec:corpus-snapshot}
\label{sec:optimization-protocol}

Stage~I uses the approximately 40M-record broad-coverage pool; Stage~II uses the approximately 5M-record refinement pool with high-quality Stage~I replay. Before training, records undergo quality checks, cross-source deduplication, and benchmark-overlap screening. Pages and derived views share a content-group identity for split assignment. The allocation probe is excluded from training, mining, and final evaluation, while the mining sample belongs to the training split.

\paragraph{Stage~I broad-coverage training.}
Stage~I jointly updates the visual encoder and language model using the page- and component-level supervision in Section~\ref{sec:stage1-support}. Source and appearance shares are specified in target tokens included in the loss and converted to record probabilities with Equation~\ref{eq:token-mixture-realization}. Optimization follows Equation~\ref{eq:objective}, with training shares tracked by input granularity and limits on repeated sampling from each duplicate group. The resulting checkpoint initializes diagnosis and the controlled Stage~II comparisons.

\paragraph{Stage~II capability refinement.}
Stage~II combines verified hard examples, targeted real and synthetic additions, and broad replay. Hard examples originate from training-pool mining or separately reviewed annotation candidates and are assessed against final targets. Additions address audited gaps in Stage~I coverage. Additions and replay form the residual stream; records assigned to the separately capped hard stream are excluded from that sampling role. Equations~\ref{eq:residual-mixture}--\ref{eq:hard-mixture} define rebalancing and stream combination, with Equation~\ref{eq:token-mixture-realization} providing token-to-record conversion. Sampling roles change exposure without changing targets.

Pool sizes count unique eligible records and exclude the probe; repeated draws do not increase these counts. The budgets $B_1$ and $B_2$ instead constrain supervised target tokens consumed in each stage.

\paragraph{Exposure.}
Training logs record source, input-granularity, and duplicate-group identifiers, target shares and sampling probabilities, replay and augmentation policies, and total token budgets. Planned and actual token shares are tracked alongside record-draw counts to distinguish corpus size from training exposure.

\subsubsection{Backbones, Output Representation, and Training Controls}
\label{sec:serialization}

\ModelName{} uses 2B and 4B Qwen3-VL-Instruct backbones~\cite{bai2025qwen3}. Both scales retain the native dynamic-resolution vision transformer (ViT), autoregressive large language model (LLM), tokenizer, and positional encoding, and neither introduces an external document detector, OCR engine, table parser, or specialized decoder.

Training targets and predictions share the representations in Section~\ref{sec:formulation}: reading-order Markdown, HTML tables, and LaTeX formulas. Crops retain their visible-content representation. Training normalization remains separate from benchmark-specific scoring conversion.

Table~\ref{tab:optimization} reports the shared optimization settings used for the 2B and 4B runs and labels only the settings that differ by stage. Stage~I updates the full model, whereas Stage~II freezes the visual encoder and refines the language model from a Stage~I checkpoint of the same model size. The different device and accumulation configurations preserve a global batch of 512 sequences at both scales.

\begin{table}[htbp]
\centering
\caption{Training settings for the reported checkpoints. Stage-specific settings are labeled explicitly; all unlabeled settings are shared by Stage~I and Stage~II.}
\label{tab:optimization}
\adjustbox{max width=\linewidth}{%
\begin{tabular}{l|l}
\toprule
\TableHeader{Item} & \TableHeader{Setting} \\
\midrule
Initialization & \textbf{Stage~I:} Qwen3-VL-Instruct; \textbf{Stage~II:} corresponding Stage~I checkpoint \\
Trainable modules & \textbf{Stage~I:} ViT and LLM Backbone; \textbf{Stage~II:} LLM Backbone only (ViT frozen) \\
Optimizer & Fused AdamW, $\beta_1=0.9$, $\beta_2=0.95$, $\epsilon=10^{-8}$ \\
Learning rate & $5\!\times\!10^{-6}$, cosine decay, 500-step warmup \\
Weight decay / gradient clip & $0.1$ / $1.0$ \\
Global batch size & 512 \\
Sequence length & 8,192 \\
Numerical and systems setup & bfloat16, gradient checkpointing, FlashAttention, DeepSpeed ZeRO-2 \\
\bottomrule
\end{tabular}%
}
\end{table}
Here $\beta_1$ and $\beta_2$ are AdamW's first- and second-moment decay rates, and $\epsilon$ provides numerical stabilization.

\subsubsection{Inference and Evaluation}
\label{sec:inference-protocol}

Training-pool inference supports error analysis and hard-example mining; the separate allocation probe estimates group residuals (Section~\ref{sec:clustering}). Thresholds and mixture settings are calibrated on pre-specified training-pool folds, not final benchmarks. These diagnostic results guide data construction and are not an additional evaluation set.

Public-benchmark inference is performed only after the dataset contents and training configuration are fixed. Across reported benchmarks, models of the same scale use the common document-parsing system prompt reproduced in Appendix~\ref{app:doc-parsing-prompt}, together with a shared image-preprocessing policy and deterministic decoding configuration. Predictions are generated directly in the shared parsing representation without external reranking or content-level correction, and each benchmark is scored with its official protocol under the same evaluator for every controlled variant. For robustness, we repeat inference three times for each checkpoint and report the mean benchmark scores.

%% file: sections/05_experiments.tex
\section{Experiments}
\label{sec:experiments}

We evaluate \ModelName{} at two complementary levels. First, we compare the final 2B and 4B models with existing systems on OmniDocBench v1.6 and PureDocBench. Second, we assess the aggregate effect of capability-aware refinement through Stage~I--Stage~II comparisons at both model scales and examine the corresponding changes in representative predictions from each benchmark split.

\subsection{Evaluation Setup}
\label{sec:benchmarks}

\paragraph{OmniDocBench v1.6: coverage under document diversity.}
OmniDocBench v1.6~\cite{ouyang2025omnidocbench} contains 1,651 document pages spanning diverse document types, layouts, languages, and combinations of text, formulas, tables, and reading-order structures. We use this benchmark to assess document parsing under heterogeneous, naturally occurring content and layout conditions.

\paragraph{PureDocBench: paired degradation robustness.}
PureDocBench~\cite{li2026far} complements annotation-based evaluation with source-derived ground truth. Its audit found 2,580 confirmed annotation errors among 21,353 evaluator-scored blocks in OmniDocBench v1.5; this finding does not estimate the error rate of v1.6, but it motivates evaluation against independently generated, exactly traceable targets. PureDocBench comprises 1,475 HTML/CSS source pages from 10 domains and 66 subcategories. Each source is rendered into aligned \emph{Clean}, \emph{Digital Degraded}, and \emph{Real Degraded} views that share the same ground truth; the latter two tracks cover ten synthetic degradation scenarios and four physical or screen-mediated acquisition pipelines, respectively.

\paragraph{Metrics.}
Both benchmarks report TextEdit$\downarrow$, FormulaCDM$\uparrow$, TableTEDS$\uparrow$, and ROEdit$\downarrow$, together with an Overall score. Equation~\ref{eq:overall-score} defines Overall as the arithmetic mean of text accuracy, FormulaCDM, and TableTEDS:
\begin{equation}
  \operatorname{Overall} = \frac{100\bigl(1-\operatorname{TextEdit}\bigr) + \operatorname{FormulaCDM} + \operatorname{TableTEDS}}{3}.
  \label{eq:overall-score}
\end{equation}
ROEdit is reported separately and does not enter Overall. Equation~\ref{eq:puredoc-avg3} defines the PureDocBench mean Overall across the three tracks:
\begin{equation}
  \mathrm{Avg}_3 = \frac{\operatorname{Overall}_{\mathrm{clean}} + \operatorname{Overall}_{\mathrm{digital}} + \operatorname{Overall}_{\mathrm{real}}}{3}.
  \label{eq:puredoc-avg3}
\end{equation}

\paragraph{Baselines.}
We compare against three complementary classes of document parsing systems: general-purpose VLMs, pipeline-based parsers, and specialized end-to-end OCR models. This grouping separates general multimodal scale, modular parsing, and document-specific end-to-end specialization. The end-to-end group includes FD-RL~\cite{zhong2026fdrl}, which uses format-decoupled reinforcement learning. Among the baselines, a superscript $*$ denotes a result produced with our evaluation pipeline under the stated benchmark protocol; unmarked results are taken from previous work.

\subsection{Main Results}
\label{sec:main-results}

Table~\ref{tab:omni-main} reports the OmniDocBench v1.6 comparison. \ModelNameFourB{} establishes a new state of the art among the compared end-to-end specialists, achieving 95.38 Overall and improving the previous leading result by 0.64 points. It also leads this series in TextEdit (0.036), FormulaCDM (96.81), TableTEDS\_S (95.34), and ROEdit (0.125). \ModelNameTwoB{} delivers top-tier end-to-end performance with 95.06 Overall, together with 95.94 FormulaCDM, 93.03 TableTEDS, and 95.26 TableTEDS\_S. These results place both scales on the leading end-to-end parsing frontier, with the 4B model setting the strongest aggregate result. Because OmniDocBench uses annotation-derived ground truth, this comparison is conditioned on the frozen v1.6 benchmark and evaluator revisions.

\begin{table}[!ht]
  \centering
  \setlength{\tabcolsep}{2.5pt}
  \renewcommand{\arraystretch}{0.90}
  \caption{Comparison of document parsing results on OmniDocBench v1.6. We report Overall together with text, formula, table, and reading-order metrics; our results are means over three inference runs. A superscript $*$ marks baseline results obtained with our evaluation pipeline, whereas unmarked baseline results are taken from previous work~\cite{zhang2026paddleocr,li2026hunyuanocr}. Bold and underlining highlight the two leading results within each model series. Baseline rows are ordered by Overall from low to high; our models are reported separately at the end of the end-to-end series.}
  \label{tab:omni-main}
  \adjustbox{max width=\textwidth,max totalheight=.82\textheight}{%
  \begin{tabular}{llccccccc}
    \toprule
    \TableHeader{Model Type} & \TableHeader{Model} & \TableHeader{Size} & \TableHeader{Overall$\uparrow$} & \TableHeader{Text$^{\mathrm{Edit}}\downarrow$} & \TableHeader{Formula$^{\mathrm{CDM}}\uparrow$} & \TableHeader{Table$^{\mathrm{TEDS}}\uparrow$} & \TableHeader{Table$^{\mathrm{TEDS\_S}}\uparrow$} & \TableHeader{ROEdit$\downarrow$} \\
    \midrule
    \multirow{7}{*}{\shortstack{General-purpose\\VLMs}} & InternVL3.5-241B~\cite{wang2025internvl35advancingopensourcemultimodal} & 241B & 83.76 & 0.130 & 89.95 & 74.35 & 79.78 & 0.215 \\
     & Kimi K2.5~\cite{team2026kimi} & 1T & 84.53 & 0.107 & 83.50 & 80.76 & 84.00 & 0.211 \\
     & GPT-5.2~\cite{openai2025gpt52} & -- & 86.59 & 0.114 & 88.21 & 82.95 & 87.93 & 0.193 \\
     & Qwen3-VL-235B~\cite{bai2025qwen3} & 235B & 89.78 & \underline{0.063} & 92.55 & 83.07 & 86.75 & 0.166 \\
     & Gemini 3 Flash~\cite{geminiteam2025gemini3flash} & -- & 92.62 & 0.066 & 95.16 & \underline{89.29} & \textbf{93.51} & 0.172 \\
     & Gemini 3 Pro~\cite{geminiteam2025gemini3} & -- & \underline{92.91} & 0.064 & \textbf{95.99} & 89.15 & \underline{92.96} & \underline{0.165} \\
     & Ovis2.6-30B-A3B~\cite{lu2025ovis205} & 30B & \textbf{93.70} & \textbf{0.035} & \underline{95.17} & \textbf{89.44} & 92.40 & \textbf{0.135} \\
    \midrule
    \multirow{10}{*}{\shortstack{Pipeline-based\\systems}} & Dolphin-1.5~\cite{feng2025dolphin} & 0.3B & 86.52 & 0.094 & 87.49 & 81.43 & 84.82 & 0.167 \\
     & MonkeyOCR-pro-3B~\cite{li2025monkeyocr} & 3B & 88.57 & 0.074 & 88.74 & 84.35 & 88.62 & 0.189 \\
     & Dolphin-v2~\cite{feng2026dolphinv2} & 3B & 89.50 & 0.069 & 91.01 & 84.40 & 87.44 & 0.150 \\
     & OpenDoc-0.1B~\cite{du2025unirec} & 0.1B & 90.67 & 0.049 & 93.02 & 83.88 & 87.45 & 0.140 \\
     & MinerU2.5~\cite{niu2026mineru2} & 1.2B & 93.04 & 0.045 & 95.77 & 87.88 & 91.47 & 0.130 \\
     & Youtu-Parsing~\cite{youtu-parsing} & 2.5B & 93.74 & 0.044 & 93.63 & 92.02 & 95.00 & \textbf{0.116} \\
     & PaddleOCR-VL-1.5~\cite{cui2026paddleocr} & 0.9B & 94.93 & 0.038 & 96.89 & 91.67 & 94.37 & 0.130 \\
     & GLM-OCR~\cite{duan2026glmocrtechnicalreport} & 0.9B & 95.22 & 0.044 & 97.18 & 92.83 & 95.39 & 0.133 \\
     & MinerU2.5-Pro~\cite{wang2026mineru2} & 1.2B & \underline{95.75} & \underline{0.036} & \underline{97.45} & \underline{93.42} & \underline{95.92} & \underline{0.120} \\
     & PaddleOCR-VL-1.6~\cite{zhang2026paddleocr} & 0.9B & \textbf{96.33} & \textbf{0.033} & \textbf{97.49} & \textbf{94.76} & \textbf{97.11} & 0.127 \\
    \midrule
    \multirow{21}{*}{\shortstack{End-to-end\\specialists}} & Nanonets-OCR2\smash{\textsuperscript{*}}~\cite{mandal2025nanonetsocr2} & 3B & 83.20 & 0.108 & 80.35 & 80.10 & 85.26 & 0.211 \\
     & OCRFlux-3B\smash{\textsuperscript{*}}~\cite{chatdoc2025ocrflux3b} & 3B & 83.31 & 0.126 & 88.75 & 73.78 & 77.98 & 0.217 \\
     & POINTS-Reader~\cite{points-reader} & 3B & 83.37 & 0.096 & 85.72 & 73.98 & 77.40 & 0.198 \\
     & Nanonets-OCR-s~\cite{Nanonets-OCR-S} & 3B & 83.61 & 0.108 & 81.46 & 80.18 & 84.51 & 0.213 \\
     & olmOCR-2-7B\smash{\textsuperscript{*}}~\cite{olmocr2} & 7B & 85.51 & 0.106 & 88.84 & 78.32 & 82.81 & 0.223 \\
     & olmOCR~\cite{olmocrbench} & 7B & 85.74 & 0.139 & 88.10 & 83.00 & 87.17 & 0.216 \\
     & DeepSeek-OCR\smash{\textsuperscript{*}}~\cite{wei2025deepseek} & 3B & 86.31 & 0.077 & 84.71 & 81.87 & 86.07 & 0.171 \\
     & OCRVerse~\cite{zhong2026ocrverse} & 4B & 88.60 & 0.063 & 89.61 & 82.44 & 86.27 & 0.163 \\
     & UniRec-0.1B\smash{\textsuperscript{*}}~\cite{du2025unirec} & 0.1B & 88.91 & 0.088 & 92.14 & 83.40 & 86.79 & 0.146 \\
     & DeepSeek-OCR 2~\cite{wei2026deepseek} & 3B & 90.25 & 0.050 & 91.84 & 83.89 & 87.75 & 0.144 \\
     & dots.ocr~\cite{li2025dots} & 3B & 90.77 & 0.048 & 89.95 & 87.18 & 90.58 & 0.138 \\
     & FD-RL\smash{\textsuperscript{*}}~\cite{zhong2026fdrl} & 4B & 91.21 & 0.055 & 92.92 & 86.22 & 90.92 & 0.145 \\
     & HunyuanOCR~\cite{HunyuanOCR_2025} & 1B & 92.03 & 0.048 & 88.60 & 92.37 & 93.99 & 0.138 \\
     & dots.mocr\smash{\textsuperscript{*}}~\cite{zheng2026multimodal} & 3B & 92.57 & 0.042 & 92.09 & 89.78 & 92.92 & 0.133 \\
     & FireRed-OCR~\cite{fireredocr} & 2B & 93.26 & \underline{0.037} & 95.44 & 88.04 & 91.06 & 0.131 \\
     & Logics-Parsing-v2~\cite{logics2026logicsparsingv2} & 4B & 93.33 & 0.041 & 95.65 & 88.42 & 91.98 & 0.137 \\
     & Qianfan-OCR~\cite{dong2026qianfanocrunifiedendtoendmodel} & 4B & 93.90 & 0.040 & 95.08 & 90.53 & 93.31 & 0.130 \\
     & Unlimited-OCR~\cite{yin2026unlimitedocrworks} & 3B-A0.5B & 93.92 & 0.042 & 95.79 & 90.16 & 93.32 & \underline{0.129} \\
     & HunyuanOCR-1.5~\cite{li2026hunyuanocr} & 1B & 94.74 & 0.039 & 94.50 & \textbf{93.67} & 94.71 & \underline{0.129} \\
    \cmidrule(lr){2-9}
    \oursrow  & \ModelName{} (Ours) & 2B & \underline{95.06} & 0.038 & \underline{95.94} & \underline{93.03} & \underline{95.26} & 0.130 \\
    \oursrow  & \ModelName{} (Ours) & 4B & \textbf{95.38} & \textbf{0.036} & \textbf{96.81} & 92.95 & \textbf{95.34} & \textbf{0.125} \\
    \bottomrule
  \end{tabular}}
\end{table}


Table~\ref{tab:puredoc-main} reports the PureDocBench comparison. \ModelNameFourB{} establishes the highest reported $\mathrm{Avg}_3$ at 75.54, exceeding the leading broad-domain baseline, Qwen3.5-122B-A10B, by 1.43 points. It also achieves the highest Overall scores on Clean (79.81) and Digital Degraded (77.74), while its 69.08 on Real Degraded advances the previous end-to-end state of the art by 1.44 points. \ModelNameTwoB{} delivers top-tier end-to-end performance with 73.86 $\mathrm{Avg}_3$, including 79.36 on Clean and 76.62 on Digital Degraded. Scaling from 2B to 4B yields gains of 0.45, 1.12, and 3.48 points on Clean, Digital Degraded, and Real Degraded, respectively, indicating that additional capacity is most beneficial under stronger appearance shifts.

\input{sections/05_puredoc_main_table}


\subsection{Effect of Stage~II Capability-Aware Refinement}
\label{sec:ablation}

We assess the aggregate effect of the complete capability-refinement stage by comparing the checkpoints obtained after broad-coverage Stage~I training with their Stage~II continuations at both model scales. Within each scale, the backbone, output representation, and evaluation protocol are held constant. Stage~II nevertheless combines continued optimization, residual-aware allocation, curated hard examples, targeted additions, and replay, so this comparison does not isolate the contribution of any individual component. Accordingly, ``stage-wise ablation'' in Table~\ref{tab:stage-wise-ablation} refers to comparing the model before and after the complete Stage~II protocol rather than to a component-wise ablation. Each entry is the mean of the three reported inference runs.

\begin{table}[!ht]
  \centering
  \caption{Stage-wise ablation at both model scales. Values are Overall scores averaged over three inference runs; $\mathrm{Avg}_3$ is the mean over the three PureDocBench tracks. Gains are computed as Stage~II minus Stage~I using the displayed two-decimal scores.}
  \label{tab:stage-wise-ablation}
  \adjustbox{max width=\linewidth}{%
  \begin{tabular}{llccccc}
    \toprule
    \multirow{2}{*}{\TableHeader{Scale}} & \multirow{2}{*}{\TableHeader{Training stage}} & \multirow{2}{*}{\TableHeader{OmniDocBench$\uparrow$}} & \multicolumn{4}{c}{\TableHeader{PureDocBench}} \\
    \cmidrule(lr){4-7}
    & & & \TableHeader{Clean$\uparrow$} & \TableHeader{Digital Degraded$\uparrow$} & \TableHeader{Real Degraded$\uparrow$} & \TableHeader{$\mathrm{Avg}_3\uparrow$} \\
    \midrule
    2B & Stage~I: broad coverage & 93.50 & 78.90 & 74.24 & 63.23 & 72.12 \\
    \oursrow 2B & Stage~II: capability-aware refinement & \textbf{95.06} & \textbf{79.36} & \textbf{76.62} & \textbf{65.60} & \textbf{73.86} \\
    & {\small Gain} & {\small $+1.56$} & {\small $+0.46$} & {\small $+2.38$} & {\small $+2.37$} & {\small $+1.74$} \\
    \midrule
    4B & Stage~I: broad coverage & 94.22 & 79.32 & 75.19 & 65.05 & 73.19 \\
    \oursrow 4B & Stage~II: capability-aware refinement & \textbf{95.38} & \textbf{79.81} & \textbf{77.74} & \textbf{69.08} & \textbf{75.54} \\
    & {\small Gain} & {\small $+1.16$} & {\small $+0.49$} & {\small $+2.55$} & {\small $+4.03$} & {\small $+2.35$} \\
    \bottomrule
  \end{tabular}}
\end{table}

At both scales, Stage~II improves OmniDocBench Overall and all three PureDocBench tracks. The 2B model gains 1.56 points on OmniDocBench and 1.74 points in $\mathrm{Avg}_3$, while the 4B model gains 1.16 and 2.35 points, respectively. Improvements are largest on Digital Degraded ($+2.38$ and $+2.55$ for 2B and 4B) and Real Degraded ($+2.37$ and $+4.03$). Clean performance changes by only $+0.46$ at 2B and $+0.49$ at 4B. This pattern is consistent with the intended emphasis of Stage~II on distributional tails and acquisition-induced appearance shifts, but the stage-wise comparison does not isolate residual-aware allocation from the other Stage~II interventions.

Stage-wise qualitative comparisons that localize representative structural and recognition errors are provided in \appref{app:stage-case-studies}. In particular, Figure~\ref{fig:case-pdb-real}(b) shows the correction of a name transcription error and the restoration of formula serialization in a photographed Chinese contract. Additional end-to-end predictions across general, formula-rich, table-rich, and scanned or photographed documents are provided in \appref{app:multiscene-case-studies}.

%% file: sections/05_puredoc_main_table.tex
\begin{table}[!ht]
  \centering
  \setlength{\tabcolsep}{3pt}
  \renewcommand{\arraystretch}{0.82}
  \caption{Three-track leaderboard on PureDocBench. Each track reports Overall (Ovr), TextEdit (TxE), FormulaCDM (FCM), TableTEDS (TDS), and ROEdit; our results are means over three inference runs. $\mathrm{Avg}_3$ is the mean of the three track-level Overall scores. A superscript $*$ marks baseline results obtained with our evaluation pipeline, whereas unmarked baseline results are taken from previous work~\cite{li2026far}. Bold and underlining highlight the two leading results within each model series. Baseline rows are ordered by $\mathrm{Avg}_3$ from low to high; our models are reported separately at the end of the end-to-end series.}
  \label{tab:puredoc-main}
  \adjustbox{max width=\linewidth,max totalheight=.86\textheight}{%
  \begin{tabular}{llc*{3}{ccccc}}
    \toprule
    \multirow{2}{*}{\TableHeader{Model}} & \multirow{2}{*}{\TableHeader{Params}} & \multirow{2}{*}{\TableHeader{$\mathrm{Avg}_3$}} & \multicolumn{5}{c}{\TableHeader{Clean}} & \multicolumn{5}{c}{\TableHeader{Digital Degraded}} & \multicolumn{5}{c}{\TableHeader{Real Degraded}} \\
    \cmidrule(lr){4-8}\cmidrule(lr){9-13}\cmidrule(lr){14-18}
    & & & \TableHeader{Ovr$\uparrow$} & \TableHeader{TxE$\downarrow$} & \TableHeader{FCM$\uparrow$} & \TableHeader{TDS$\uparrow$} & \TableHeader{ROEdit$\downarrow$} & \TableHeader{Ovr$\uparrow$} & \TableHeader{TxE$\downarrow$} & \TableHeader{FCM$\uparrow$} & \TableHeader{TDS$\uparrow$} & \TableHeader{ROEdit$\downarrow$} & \TableHeader{Ovr$\uparrow$} & \TableHeader{TxE$\downarrow$} & \TableHeader{FCM$\uparrow$} & \TableHeader{TDS$\uparrow$} & \TableHeader{ROEdit$\downarrow$} \\
    \specialrule{\lightrulewidth}{\aboverulesep}{0pt}
    \rowcolor{WeChatMint}\multicolumn{18}{l}{\rule[-1.5ex]{0pt}{4.5ex}\TableHeader{General VLMs}} \\
    MiniCPM-V-4.5~\cite{yu2025minicpmv45cookingefficient} & 8B & 46.26 & 51.81 & 0.439 & 45.97 & 53.36 & 0.481 & 49.38 & 0.461 & 42.79 & 51.50 & 0.489 & 37.59 & 0.583 & 32.01 & 39.06 & 0.552 \\
    Step3-VL~\cite{huang2026step30vl010b} & 10B & 50.48 & 53.65 & 0.496 & 53.41 & 57.16 & 0.509 & 52.74 & 0.516 & 53.62 & 56.15 & 0.529 & 45.06 & 0.579 & 45.42 & 47.66 & 0.573 \\
    Qwen3.5-0.8B~\cite{qwen3.5} & 0.8B & 55.99 & 60.77 & 0.376 & 54.39 & 65.54 & 0.500 & 59.28 & 0.386 & 54.22 & 62.22 & 0.510 & 47.93 & 0.498 & 44.60 & 48.98 & 0.557 \\
    Qwen3-VL-2B~\cite{bai2025qwen3} & 2B & 62.09 & 66.37 & 0.300 & 59.04 & 70.03 & 0.439 & 65.81 & 0.314 & 60.25 & 68.52 & 0.448 & 54.09 & 0.428 & 51.05 & 53.99 & 0.511 \\
    Qwen3.5-2B~\cite{qwen3.5} & 2B & 62.46 & 66.24 & 0.348 & 62.84 & 70.70 & 0.473 & 65.22 & 0.350 & 58.30 & 72.36 & 0.477 & 55.92 & 0.440 & 50.99 & 60.79 & 0.521 \\
    Qwen3.5-35B-A3B~\cite{qwen3.5} & 35B-A3B & 65.68 & 68.40 & 0.232 & 64.94 & 63.45 & 0.374 & 68.04 & 0.245 & 64.78 & 63.86 & 0.379 & 60.59 & 0.310 & 59.68 & 53.07 & 0.419 \\
    Qwen3.5-397B-A17B~\cite{qwen3.5} & 397B-A17B & 66.72 & 69.12 & 0.233 & 65.26 & 65.40 & \underline{0.366} & 68.34 & 0.244 & 63.91 & 65.53 & 0.376 & 62.70 & 0.287 & 60.70 & 56.12 & 0.399 \\
    Qwen3-VL-4B~\cite{bai2025qwen3} & 4B & 67.50 & 72.04 & 0.262 & 65.10 & 77.17 & 0.418 & 70.84 & 0.272 & 63.54 & 76.13 & 0.425 & 59.61 & 0.378 & 55.15 & 61.47 & 0.480 \\
    Qwen3-VL-8B~\cite{bai2025qwen3} & 8B & 69.07 & 72.44 & 0.261 & 65.10 & 78.35 & 0.411 & 72.03 & 0.266 & 64.88 & 77.82 & 0.409 & 62.73 & 0.342 & 55.55 & 66.81 & 0.448 \\
    Qwen3.5-27B~\cite{qwen3.5} & 27B & 69.57 & 72.07 & \underline{0.227} & 66.36 & 72.51 & \textbf{0.362} & 70.73 & \underline{0.236} & 64.61 & 71.17 & \underline{0.367} & 65.92 & \underline{0.283} & 61.23 & 64.82 & \underline{0.390} \\
    Qwen3.5-4B~\cite{qwen3.5} & 4B & 69.82 & 73.45 & 0.276 & \textbf{69.96} & 78.02 & 0.410 & 72.53 & 0.281 & \textbf{68.88} & 76.78 & 0.412 & 63.47 & 0.380 & 61.27 & 67.17 & 0.477 \\
    Kimi-K2.6~\cite{moonshotai2026kimik26} & 1T-A32B & 70.10 & 72.32 & 0.303 & 66.93 & \underline{80.30} & 0.466 & 69.95 & 0.322 & 64.69 & 77.31 & 0.475 & 68.02 & 0.335 & \underline{62.44} & 75.14 & 0.481 \\
    Gemini-3.1-Pro~\cite{geminiteam2026gemini31pro} & -- & 70.43 & 70.04 & 0.306 & 65.63 & 75.08 & 0.409 & 69.28 & 0.322 & 65.81 & 74.24 & 0.417 & \textbf{71.98} & 0.300 & \textbf{68.62} & \textbf{77.26} & \textbf{0.386} \\
    Qwen3.5-9B~\cite{qwen3.5} & 9B & \underline{70.89} & \underline{73.87} & 0.254 & 67.60 & 79.39 & 0.388 & \underline{73.34} & 0.260 & 67.00 & \underline{79.01} & 0.396 & 65.45 & 0.332 & 60.91 & 68.59 & 0.437 \\
    Qwen3.5-122B-A10B~\cite{qwen3.5} & 122B-A10B & \textbf{74.11} & \textbf{76.14} & \textbf{0.226} & \underline{67.96} & \textbf{83.03} & 0.375 & \textbf{76.34} & \textbf{0.220} & \underline{67.82} & \textbf{83.21} & \textbf{0.366} & \underline{69.85} & \textbf{0.281} & 62.19 & \underline{75.44} & 0.401 \\
    \specialrule{\lightrulewidth}{\aboverulesep}{0pt}
    \rowcolor{WeChatMint}\multicolumn{18}{l}{\rule[-1.5ex]{0pt}{4.5ex}\TableHeader{Pipeline-based Systems}} \\
    OpenDoc-0.1B~\cite{du2025unirec} & 0.1B & 52.34 & 60.28 & 0.411 & 53.09 & 68.86 & 0.519 & 52.46 & 0.501 & 48.41 & 59.04 & 0.577 & 44.27 & 0.547 & 38.46 & 49.06 & 0.603 \\
    MonkeyOCR-pro-1.2B~\cite{li2025monkeyocr} & 1.2B & 53.54 & 61.09 & 0.358 & 47.43 & 71.60 & 0.498 & 55.72 & 0.416 & 43.91 & 64.83 & 0.529 & 43.82 & 0.556 & 36.94 & 50.07 & 0.609 \\
    MonkeyOCR-pro-3B~\cite{li2025monkeyocr} & 3B & 55.37 & 62.23 & 0.346 & 48.46 & 72.83 & 0.492 & 57.40 & 0.397 & 45.57 & 66.32 & 0.526 & 46.49 & 0.511 & 38.18 & 52.43 & 0.600 \\
    Dolphin-v2~\cite{feng2026dolphinv2} & 3B & 57.02 & 65.90 & 0.342 & 59.80 & 72.12 & 0.429 & 60.24 & 0.393 & 52.20 & 67.86 & 0.461 & 44.92 & 0.553 & 39.98 & 50.04 & 0.558 \\
    GLM-OCR~\cite{duan2026glmocrtechnicalreport} & 0.9B & 63.34 & 68.65 & 0.314 & 57.89 & 79.44 & 0.470 & 63.06 & 0.383 & 53.23 & 74.21 & 0.520 & 58.31 & 0.433 & 50.34 & 67.83 & 0.543 \\
    Dolphin-1.5\smash{\textsuperscript{*}}~\cite{feng2025dolphin} & 0.3B & 64.15 & 72.51 & 0.260 & 62.78 & 80.75 & 0.383 & 65.37 & 0.347 & 57.96 & 72.83 & 0.441 & 54.59 & 0.459 & 46.93 & 62.75 & 0.507 \\
    PaddleOCR-VL-1.5~\cite{cui2026paddleocr} & 0.9B & 66.75 & 73.01 & 0.266 & 63.53 & \underline{82.12} & 0.428 & 66.73 & 0.339 & 58.03 & \underline{76.07} & 0.478 & 60.50 & 0.398 & \underline{54.00} & \underline{67.33} & 0.510 \\
    PaddleOCR-VL-1.6\smash{\textsuperscript{*}}~\cite{zhang2026paddleocr} & 0.9B & 66.98 & 72.15 & 0.263 & 64.80 & 77.92 & 0.422 & 67.17 & 0.314 & \underline{61.53} & 71.35 & 0.455 & \underline{61.63} & 0.383 & \textbf{56.44} & 66.71 & 0.500 \\
    MinerU2.5~\cite{niu2026mineru2} & 1.2B & 67.66 & 74.90 & \textbf{0.184} & 62.08 & 81.04 & \textbf{0.327} & 68.92 & \textbf{0.245} & 56.99 & 74.24 & \textbf{0.374} & 59.15 & \underline{0.370} & 49.01 & 65.41 & \underline{0.446} \\
    Youtu-Parsing~\cite{youtu-parsing} & 2.5B & \underline{68.32} & \underline{75.02} & 0.230 & \textbf{67.34} & 80.74 & 0.358 & \underline{69.66} & \underline{0.270} & 61.44 & 74.49 & 0.388 & 60.29 & \textbf{0.360} & 52.20 & 64.69 & \textbf{0.430} \\
    MinerU2.5-Pro~\cite{wang2026mineru2} & 1.2B & \textbf{70.07} & \textbf{75.87} & \underline{0.222} & \underline{65.14} & \textbf{84.68} & \underline{0.346} & \textbf{71.77} & 0.272 & \textbf{61.79} & \textbf{80.73} & \underline{0.378} & \textbf{62.56} & 0.375 & 52.70 & \textbf{72.47} & \underline{0.446} \\
    \specialrule{\lightrulewidth}{\aboverulesep}{0pt}
    \rowcolor{WeChatMint}\multicolumn{18}{l}{\rule[-1.5ex]{0pt}{4.5ex}\TableHeader{End-to-End Specialists}} \\
    OCRFlux-3B~\cite{chatdoc2025ocrflux3b} & 3B & 42.06 & 47.14 & 0.454 & 38.35 & 48.46 & 0.424 & 41.82 & 0.486 & 31.90 & 42.17 & 0.437 & 37.21 & 0.559 & 32.65 & 34.87 & 0.491 \\
    DeepSeek-OCR~\cite{wei2025deepseek} & 3B & 46.98 & 53.50 & 0.419 & 45.39 & 57.06 & 0.514 & 46.95 & 0.478 & 39.99 & 48.64 & 0.548 & 40.48 & 0.537 & 34.04 & 41.12 & 0.575 \\
    UniRec-0.1B~\cite{du2025unirec} & 0.1B & 48.59 & 58.91 & 0.422 & 51.31 & 67.60 & 0.526 & 52.42 & 0.501 & 48.37 & 59.04 & 0.578 & 34.44 & 0.658 & 30.97 & 38.16 & 0.685 \\
    POINTS-Reader\smash{\textsuperscript{*}}~\cite{points-reader} & 3B & 49.24 & 53.78 & 0.405 & 55.27 & 46.53 & 0.542 & 51.24 & 0.431 & 52.49 & 44.38 & 0.544 & 42.69 & 0.517 & 41.60 & 38.14 & 0.575 \\
    DeepSeek-OCR 2~\cite{wei2026deepseek} & 3B & 49.51 & 55.53 & 0.354 & 46.00 & 56.01 & 0.466 & 49.41 & 0.412 & 40.78 & 48.67 & 0.493 & 43.60 & 0.486 & 37.30 & 42.06 & 0.533 \\
    Qianfan-OCR~\cite{dong2026qianfanocrunifiedendtoendmodel} & 4B & 51.04 & 57.22 & 0.370 & 49.79 & 58.83 & 0.443 & 50.85 & 0.438 & 44.41 & 51.96 & 0.485 & 45.06 & 0.494 & 39.08 & 45.53 & 0.509 \\
    olmOCR~\cite{olmocrbench} & 7B & 55.90 & 62.56 & 0.388 & 58.69 & 67.77 & 0.466 & 57.84 & 0.436 & 55.44 & 61.66 & 0.499 & 47.30 & 0.542 & 46.26 & 49.80 & 0.568 \\
    Nanonets-OCR2~\cite{mandal2025nanonetsocr2} & 3B & 58.36 & 64.83 & 0.254 & 44.98 & 74.94 & 0.377 & 61.23 & 0.307 & 45.40 & 68.97 & 0.408 & 49.03 & 0.435 & 35.50 & 55.09 & 0.468 \\
    HunyuanOCR~\cite{HunyuanOCR_2025} & 1B & 60.56 & 65.61 & 0.269 & 55.74 & 68.02 & 0.382 & 61.49 & 0.308 & 51.62 & 63.68 & 0.400 & 54.58 & 0.421 & 48.30 & 57.54 & 0.459 \\
    Unlimited-OCR\smash{\textsuperscript{*}}~\cite{yin2026unlimitedocrworks} & 3B-A0.5B & 62.76 & 71.28 & 0.248 & 62.68 & 75.95 & 0.374 & 63.62 & 0.317 & 54.91 & 67.71 & 0.418 & 53.39 & 0.440 & 46.89 & 57.22 & 0.470 \\
    olmOCR-2-7B~\cite{olmocr2} & 7B & 63.78 & 69.36 & 0.284 & 56.89 & 79.59 & 0.358 & 65.87 & 0.318 & 54.57 & 74.81 & 0.378 & 56.10 & 0.417 & 48.79 & 61.25 & 0.439 \\
    dots.ocr~\cite{li2025dots} & 3B & 64.55 & 72.01 & 0.248 & 61.37 & 79.51 & 0.379 & 65.95 & 0.307 & 56.67 & 71.86 & 0.417 & 55.68 & 0.403 & 47.70 & 59.63 & 0.467 \\
    Nanonets-OCR-s\smash{\textsuperscript{*}}~\cite{Nanonets-OCR-S} & 3B & 65.37 & 71.26 & 0.271 & 62.20 & 78.65 & 0.388 & 66.56 & 0.307 & 58.31 & 72.09 & 0.409 & 58.28 & 0.417 & 53.63 & 62.87 & 0.467 \\
    FireRed-OCR~\cite{fireredocr} & 2B & 65.57 & 70.81 & 0.287 & 63.86 & 77.23 & 0.396 & 68.49 & 0.319 & 62.64 & 74.77 & 0.422 & 57.42 & 0.415 & 51.60 & 62.16 & 0.474 \\
    HunyuanOCR-1.5\smash{\textsuperscript{*}}~\cite{li2026hunyuanocr} & 1B & 68.79 & 73.98 & 0.243 & 68.69 & 77.55 & 0.391 & 70.81 & 0.271 & 66.68 & 72.87 & 0.407 & 61.59 & 0.365 & 57.50 & 63.81 & 0.461 \\
    OCRVerse~\cite{zhong2026ocrverse} & 4B & 69.40 & 73.18 & 0.273 & 63.78 & 83.09 & 0.393 & 71.36 & 0.302 & 63.95 & 80.36 & 0.415 & 63.66 & 0.363 & 57.03 & 70.30 & 0.452 \\
    dots.mocr~\cite{zheng2026multimodal} & 3B & 70.39 & 76.27 & \textbf{0.151} & 66.23 & 77.65 & \textbf{0.273} & 73.16 & \textbf{0.198} & 64.32 & 74.95 & \textbf{0.309} & 61.73 & 0.312 & 54.39 & 61.97 & \underline{0.393} \\
    Logics-Parsing-v2~\cite{logics2026logicsparsingv2} & 4B & 72.61 & 76.35 & 0.213 & 67.67 & 82.67 & 0.342 & 73.85 & 0.248 & 67.33 & 79.02 & 0.375 & \underline{67.64} & \underline{0.304} & \underline{61.65} & 71.64 & 0.416 \\
    FD-RL~\cite{zhong2026fdrl} & 4B & \underline{73.92} & 78.38 & \underline{0.193} & 68.21 & 86.22 & \underline{0.334} & 76.33 & \underline{0.214} & 67.16 & 83.22 & \underline{0.350} & 67.04 & \textbf{0.298} & 58.82 & 72.08 & \textbf{0.391} \\
    \oursrow \ModelName{} (Ours) & 2B & 73.86 & \underline{79.36} & 0.214 & \underline{71.63} & \underline{87.79} & 0.347 & \underline{76.62} & 0.247 & \underline{70.74} & \underline{83.84} & 0.370 & 65.60 & 0.361 & 58.90 & \underline{74.05} & 0.437 \\
    \oursrow \ModelName{} (Ours) & 4B & \textbf{75.54} & \textbf{79.81} & 0.213 & \textbf{71.90} & \textbf{88.81} & 0.340 & \textbf{77.74} & 0.241 & \textbf{71.31} & \textbf{85.99} & 0.361 & \textbf{69.08} & 0.336 & \textbf{63.14} & \textbf{77.64} & 0.416 \\
    \bottomrule
  \end{tabular}}
\end{table}

%% file: sections/08_conclusion.tex
\section{Conclusion and Future Work}
\label{sec:conclusion}

We have presented \ModelName{}, a data-centric framework for specializing end-to-end document parsers through semantic coverage, structure-preserving degradation, and capability-aware refinement. Instantiated at 2B and 4B scales, \ModelName{} achieves Overall scores of 95.06 and 95.38 on OmniDocBench v1.6 and $\mathrm{Avg}_3$ scores of 73.86 and 75.54 on PureDocBench. The 4B model establishes state-of-the-art performance among the compared end-to-end parsers across all four settings, while the 2B model delivers top-tier performance at half the parameter count. Stage-wise comparisons show consistent gains after the complete capability-refinement stage, although they do not isolate its individual interventions. Together, these results support treating the training mixture as a measurable interface between corpus coverage and model capability rather than as a static collection of datasets.

The current evidence remains bounded by benchmark coverage and by the fidelity of synthetic degradations. Although the 4B model establishes a new end-to-end state of the art on Real Degraded pages, handwriting, historical scans, severely damaged documents, and broader acquisition conditions remain insufficiently measured. Residual rebalancing also depends on the probe set, representation model, clustering choices, and evaluator, and introduces additional computation. Future work will therefore prioritize broader authentic acquisition data, more stable and interpretable capability estimation, and lower-cost mixture updates, alongside evaluation across additional languages and document domains. High-stakes deployment should continue to pair model outputs with provenance controls and appropriate human review.

%% file: sections/G_doc_parsing_prompt.tex
\section{Document Parsing System Prompt}
\label{app:doc-parsing-prompt}

The following fixed system prompt is used for document-parsing training and inference across all controlled variants.

\begin{tcblisting}{
  enhanced,
  breakable,
  listing only,
  colback=WeChatGreen!3,
  colframe=WeChatGreen!45,
  boxrule=0.5pt,
  arc=1.5mm,
  left=2.5mm,
  right=2.5mm,
  top=2.2mm,
  bottom=2.2mm,
  before skip=5pt,
  after skip=0pt,
  listing options={
    basicstyle=\ttfamily\footnotesize,
    breaklines=true,
    breakatwhitespace=true,
    columns=fullflexible,
    keepspaces=true,
    showstringspaces=false,
    aboveskip=0pt,
    belowskip=0pt
  }
}
You are an AI assistant specialized in converting PDF images to Markdown format. Please follow these instructions for the conversion:

1. Text Processing:
- Accurately recognize all text content in the PDF image without guessing or inferring.
- Convert the recognized text into Markdown format.
- Maintain the original document structure, including headings, paragraphs, lists, etc.

2. Mathematical Formula Processing:
- Convert all mathematical formulas to LaTeX format.
- Enclose inline formulas with \( \). For example: This is an inline formula \( E = mc^2 \)
- Enclose block formulas with \[ \]. For example: \[ \frac{-b \pm \sqrt{b^2 - 4ac}}{2a} \]

3. Table Processing:
- Convert tables to HTML format.
- Wrap the entire table with <table> and </table>.

4. Figure Handling:
- Ignore figure content in the PDF image. Do not attempt to describe or convert images.

5. Output Format:
- Ensure the output Markdown document has a clear structure with appropriate line breaks between elements.
- For complex layouts, try to maintain the original document's structure and format as closely as possible.

Please strictly follow these guidelines to ensure accuracy and consistency in the conversion. Your task is to accurately convert the content of the PDF image into Markdown format without adding any extra explanations or comments.
\end{tcblisting}

%% file: sections/H_qualitative_case_studies.tex
\section{Qualitative Case Studies}
\label{app:qualitative-case-studies}

\subsection{Stage-wise Comparisons}
\label{app:stage-case-studies}

For qualitative analysis, pages are ranked within each benchmark split by the Stage~II per-page Overall score, averaged over the same three inference runs. We scan each ranking in descending order and retain three pages with a Stage~I Overall below 70, subject to the additional requirement that the scored discrepancy be attributable to a localized and unambiguous error mechanism. The displayed prediction for each stage is taken from its median-scoring run. The resulting examples complement the aggregate comparison by separating recurrent structural failures from isolated recognition errors; they are not intended to estimate error prevalence.

Figures~\ref{fig:case-odb}, \ref{fig:case-pdb-clean}, \ref{fig:case-pdb-digital}, and~\ref{fig:case-pdb-real} present the twelve stage-wise examples. Across these cases, Stage~II corrects several failure modes: omitted tabular regions, column-major reading order, unpaired field--value labels, omitted diagram and peripheral content, page-wide or local overuse of table markup, loss of mathematical serialization, and cross-document hallucination. Residual numeric and character-level transcription errors remain in the Real Degraded examples, showing that stronger structural recovery does not eliminate recognition errors under physical acquisition conditions. These observations are consistent with the larger aggregate gains on the degraded PureDocBench tracks in Table~\ref{tab:stage-wise-ablation}.

\input{sections/05_stage_case_figures}

\FloatBarrier

\subsection{Results across Document Scenarios}
\label{app:multiscene-case-studies}

Figures~\ref{fig:multiscene-formulas}, \ref{fig:multiscene-general}, \ref{fig:multiscene-scanned}, and~\ref{fig:multiscene-tables} show the model's predictions across diverse document scenarios. Each comparison contains the input, the complete predicted Markdown, and its rendering; a display-only transcription correction is disclosed in the corresponding caption.

Across these scenarios, the model preserves reading order while representing tables with HTML and mathematical expressions with LaTeX.

\input{sections/05_multiscene_examples}

\FloatBarrier

%% file: sections/05_stage_case_figures.tex
\newfontfamily\CaseBodyFace{FandolHei-Regular.otf}[BoldFont=FandolHei-Bold.otf]
\newcommand{\CaseBodyFont}{\normalfont\fontsize{5.55}{6.4}\selectfont\CaseBodyFace}
\newcommand{\CaseError}[1]{\textcolor{WeChatDanger}{\bfseries #1}}
\newlength{\CasePanelHeight}
\newlength{\CaseImageHeight}
\setlength{\CasePanelHeight}{44mm}
\setlength{\CaseImageHeight}{41.5mm}
\newcommand{\CaseColumnHeaders}{%
  \noindent\begin{minipage}[c][7mm][c]{.27\linewidth}\centering\small\WeChatBoldFace Input\end{minipage}\hfill
  \begin{minipage}[c][7mm][c]{.35\linewidth}\centering\small\WeChatBoldFace Stage~I Prediction\end{minipage}\hfill
  \begin{minipage}[c][7mm][c]{.35\linewidth}\centering\small\WeChatBoldFace Stage~II Prediction\end{minipage}\par
  \nointerlineskip{\color{WeChatLine}\rule{\linewidth}{.45pt}}\par\vspace{1.4mm}}
\newcommand{\CaseSample}[2]{%
  \vspace{2.4mm}\noindent\begin{minipage}[b]{.42\linewidth}\small\WeChatBoldFace\color{WeChatInk}#1\end{minipage}\hfill
  \begin{minipage}[b]{.56\linewidth}\raggedleft\sffamily\fontsize{7.2}{8}\selectfont\color{WeChatGray}#2\end{minipage}\par\vspace{-1.5mm}}
\newcommand{\CaseInput}[1]{%
  \begin{tcolorbox}[enhanced,width=\linewidth,height=\CasePanelHeight,colback=white,colframe=WeChatLine,
    boxrule=.35pt,arc=0pt,outer arc=0pt,boxsep=0pt,left=.8mm,right=.8mm,top=.8mm,bottom=.8mm,
    before skip=0pt,after skip=0pt,grow to right by=0pt,valign=center]
    \centering\includegraphics[width=\linewidth,height=\CaseImageHeight,keepaspectratio]{#1}
  \end{tcolorbox}}
\newcommand{\CasePrediction}[1]{%
  \begin{tcolorbox}[enhanced,width=\linewidth,height=\CasePanelHeight,colback=white,colframe=WeChatLine,
    boxrule=.35pt,arc=0pt,outer arc=0pt,left=1.25mm,right=1.25mm,top=1.15mm,bottom=1.15mm,
    before skip=0pt,after skip=0pt,grow to right by=0pt,valign=top]
    \CaseBodyFont\setlength{\parskip}{0pt}\raggedright #1
  \end{tcolorbox}}
\newcommand{\CaseTriplet}[3]{\noindent\begin{minipage}[t]{.27\linewidth}\vspace{0pt}#1\end{minipage}\hfill\begin{minipage}[t]{.35\linewidth}\vspace{0pt}#2\end{minipage}\hfill\begin{minipage}[t]{.35\linewidth}\vspace{0pt}#3\end{minipage}\par}
\newcommand{\CaseRowGap}{\vspace{2.2mm}}
\newcommand{\CaseLead}{$\ldots$\par\vspace{.2mm}}
\newcommand{\CaseTail}{\par\vspace{.2mm}$\ldots$}
\newcommand{\CaseOmit}{\par\vspace{.2mm}$\ldots$\par\vspace{.2mm}}
\newcommand{\CaseTag}[1]{\textless#1\textgreater\allowbreak}
\newcommand{\CaseLegend}{\vspace{1.3mm}\par\begingroup\raggedright\noindent{\sffamily\fontsize{7.2}{8}\selectfont\color{WeChatGray}\raisebox{-.3ex}{\tikz\draw[draw=WeChatDanger,fill=WeChatDanger!10,line width=.55pt] (0,0) rectangle (.34,.16);}\hspace{.6mm}Highlighted source regions localize the comparison. Displayed text preserves the model serialization; \textcolor{WeChatDanger}{red type} marks literal error spans, and ellipses mark omitted spans.}\par\endgroup}

\begin{figure}[!hp]
  \centering
  \CaseColumnHeaders
  \CaseSample{(a) Table coverage}{Overall: 65.87 $\rightarrow$ 99.22 \quad TableTEDS: 0.333 $\rightarrow$ 1.000}
  \CaseTriplet{
    \CaseInput{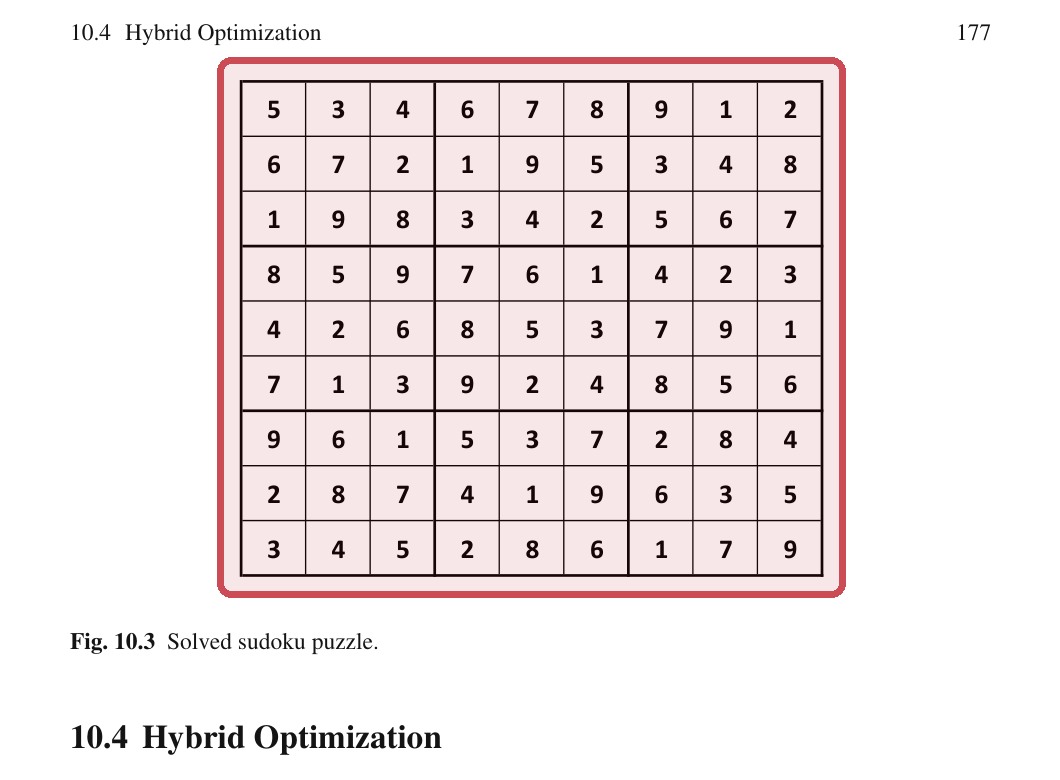}
  }{
    \CasePrediction{Fig. 10.3 Solved sudoku puzzle.\par\vspace{.7mm}\CaseError{\#\# 10.4 Hybrid Optimization}\par Hybrid methods may be required to solve particularly difficult real-world optimization problems. Implementation of hybrid methods typically requires non-trivial scripting\CaseTail}
  }{
    \CasePrediction{\CaseTag{table}\CaseTag{tr}\CaseTag{td}5\CaseTag{/td}\CaseTag{td}3\CaseTag{/td}\CaseTag{td}4\CaseTag{/td}\CaseTag{td}6\CaseTag{/td}\CaseTag{td}7\CaseTag{/td}\CaseTag{td}8\CaseTag{/td}\CaseTag{td}9\CaseTag{/td}\CaseTag{td}1\CaseTag{/td}\CaseTag{td}2\CaseTag{/td}\CaseTag{/tr}\CaseOmit\CaseTag{tr}\CaseTag{td}3\CaseTag{/td}\CaseTag{td}4\CaseTag{/td}\CaseTag{td}5\CaseTag{/td}\CaseTag{td}2\CaseTag{/td}\CaseTag{td}8\CaseTag{/td}\CaseTag{td}6\CaseTag{/td}\CaseTag{td}1\CaseTag{/td}\CaseTag{td}7\CaseTag{/td}\CaseTag{td}9\CaseTag{/td}\CaseTag{/tr}\CaseTag{/table}\par\vspace{.5mm}Fig. 10.3 Solved sudoku puzzle.\par\vspace{.5mm}\#\#\# 10.4 Hybrid Optimization\CaseTail}
  }
  \CaseRowGap
  \CaseSample{(b) Diagram coverage}{Overall: 58.98 $\rightarrow$ 100.00 \quad TextEdit: 0.410 $\rightarrow$ 0.000}
  \CaseTriplet{
    \CaseInput{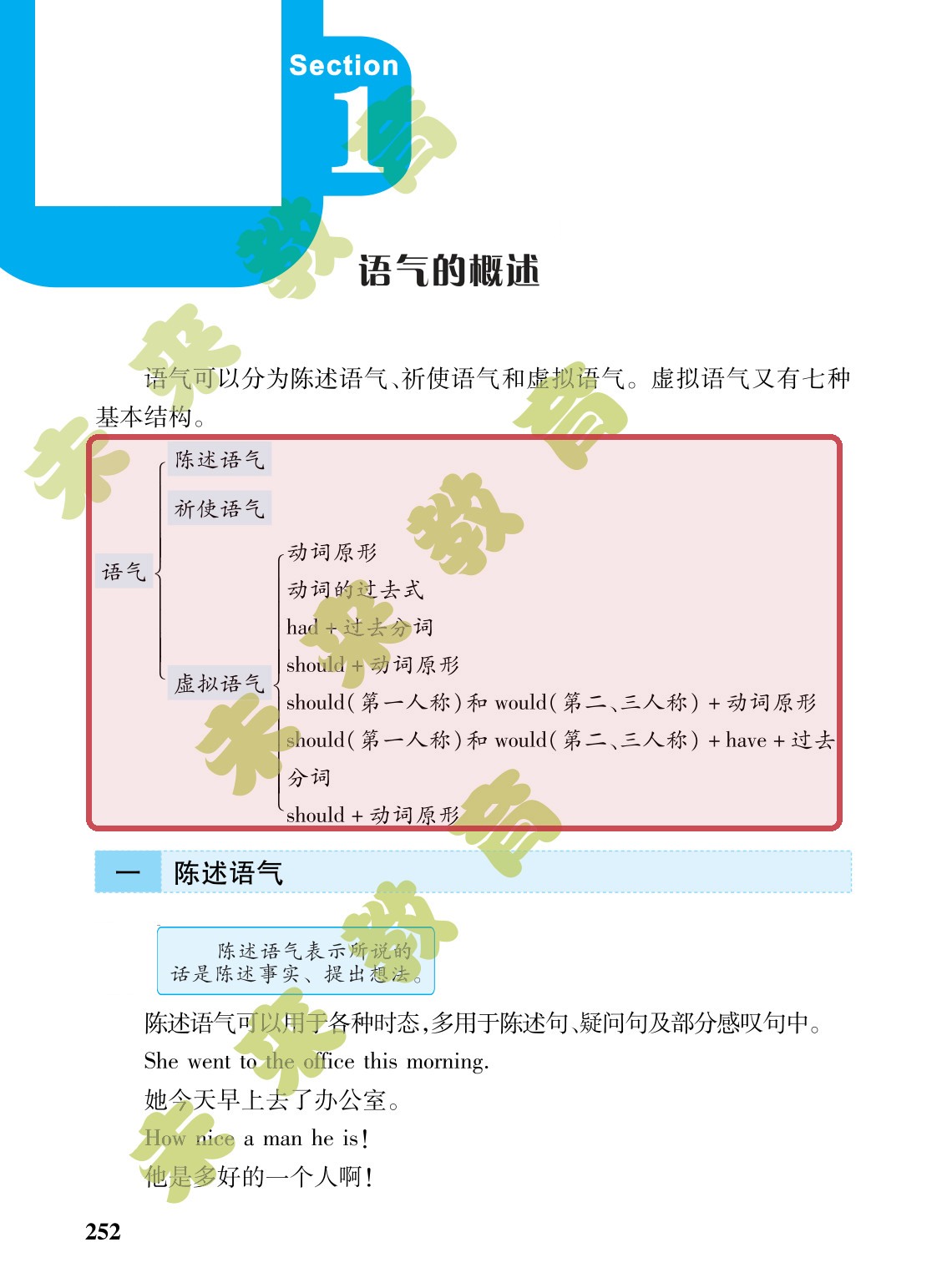}
  }{
    \CasePrediction{\#\# 语气的概述\par\vspace{.5mm}语气可以分为陈述语气、祈使语气和虚拟语气。\allowbreak 虚拟语气又有七种基本结构。\par\vspace{.7mm}\CaseError{\#\# 一 陈述语气}\par 陈述语气表示所说的话是陈述事实、提出想法。\CaseTail}
  }{
    \CasePrediction{\#\# 语气的概述\par\vspace{.5mm}语气可以分为陈述语气、祈使语气和虚拟语气。\allowbreak 虚拟语气又有七种基本结构。\par\vspace{.4mm}陈述语气\par 祈使语气\par 语气\par 动词原形\par 动词的过去式\par had + 过去分词\par 虚拟语气\par should + 动词原形\CaseOmit\#\# 一 陈述语气\CaseTail}
  }
  \CaseRowGap
  \CaseSample{(c) Peripheral text coverage}{Overall: 67.66 $\rightarrow$ 100.00 \quad TextEdit: 0.323 $\rightarrow$ 0.000}
  \CaseTriplet{
    \CaseInput{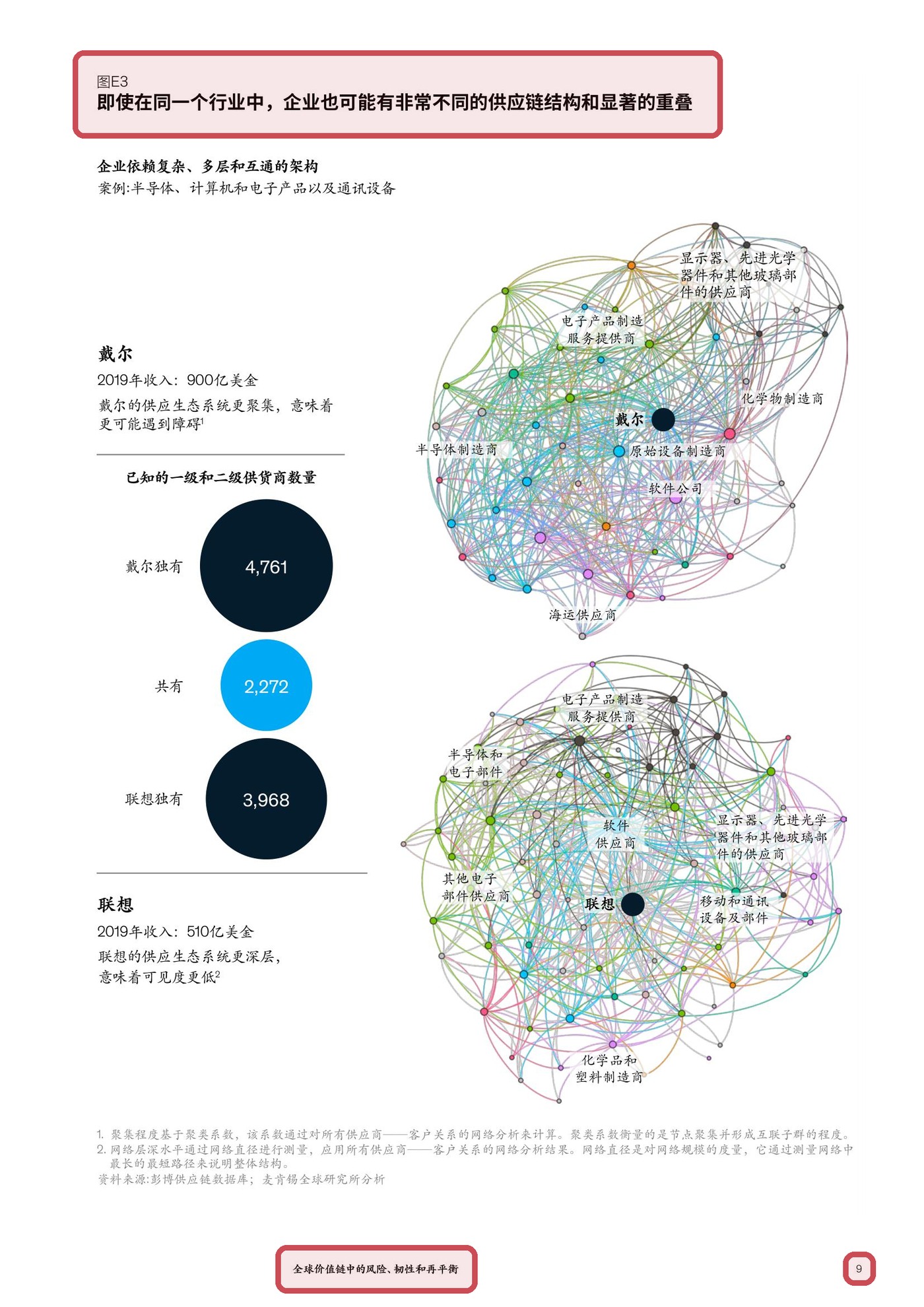}
  }{
    \CasePrediction{\CaseError{\#\# 企业依赖复杂、多层和互通的架构}\par 案例:半导体、计算机和电子产品以及通讯设备\par\vspace{.4mm}\#\# 戴尔\par 2019年收入：900亿美金\CaseOmit\#\# 联想\par 2019年收入：510亿美金\CaseOmit 资料来源:彭博供应链数据库；麦肯锡全球研究所分析}
  }{
    \CasePrediction{\#\# 图E3 即使在同一个行业中，\allowbreak 企业也可能有非常不同的供应链结构和显著的重叠\par\vspace{.4mm}企业依赖复杂、多层和互通的架构\par 案例:半导体、计算机和电子产品以及通讯设备\par\vspace{.4mm}戴尔\par 2019年收入：900亿美金\CaseOmit 联想\par 2019年收入：510亿美金\CaseOmit 资料来源:彭博供应链数据库；麦肯锡全球研究所分析\par\vspace{.4mm}全球价值链中的风险、韧性和再平衡\par 9}
  }
  \CaseLegend
  \caption{\textbf{OmniDocBench v1.6.} Stage~I omits the Sudoku grid in (a), drops a semantic hierarchy in (b), and misses peripheral title and footer text in (c); Stage~II restores the corresponding structure and coverage.}
  \label{fig:case-odb}
\end{figure}
\clearpage

\begin{figure}[p]
  \centering
  \CaseColumnHeaders
  \CaseSample{(a) Document hierarchy}{Overall: 3.46 $\rightarrow$ 98.30 \quad TextEdit: 0.996 $\rightarrow$ 0.033}
  \CaseTriplet{
    \CaseInput{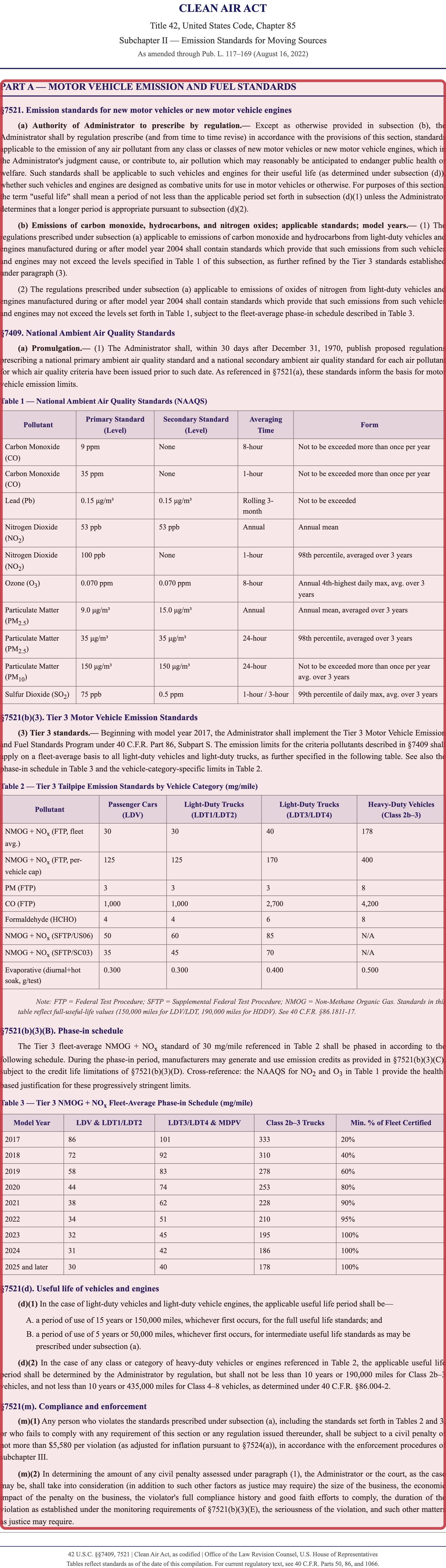}
  }{
    \CasePrediction{\CaseError{\CaseTag{table}\CaseTag{tr}\CaseTag{td}PART A — MOTOR VEHICLE EMISSION AND FUEL STANDARDS\CaseTag{/td}\CaseTag{/tr}\CaseTag{tr}\CaseTag{td}§7521. Emission standards for new motor vehicles or new motor vehicle engines\CaseTag{/td}\CaseTag{/tr}}\CaseOmit\CaseError{\CaseTag{tr}\CaseTag{td}Table 1 — National Ambient Air Quality Standards (NAAQS)\CaseTag{/td}\CaseTag{/tr}\CaseTag{tr}\CaseTag{td}Pollutant\CaseTag{/td}\CaseTag{td}Primary Standard(Level)\CaseTag{/td}\CaseTag{td}Secondary Standard(Level)\CaseTag{/td}\CaseTag{td}Averaging Time\CaseTag{/td}\CaseTag{td}Form\CaseTag{/td}\CaseTag{/tr}}\CaseOmit\CaseTag{/table}}
  }{
    \CasePrediction{\#\# PART A — MOTOR VEHICLE EMISSION AND FUEL STANDARDS\par\vspace{.4mm}\#\# §7521. Emission standards for new motor vehicles or new motor vehicle engines\par\vspace{.4mm}(a) Authority of Administrator to prescribe by regulation.— Except as otherwise provided in subsection (b), the Administrator shall by regulation prescribe\CaseOmit\#\# §7409. National Ambient Air Quality Standards\CaseOmit Table 1 — National Ambient Air Quality Standards (NAAQS)\par \CaseTag{table}\CaseTag{tr}\CaseTag{td}Pollutant\CaseTag{/td}\CaseTag{td}Primary Standard (Level)\CaseTag{/td}\CaseTag{td}Secondary Standard (Level)\CaseTag{/td}\CaseTag{td}Averaging Time\CaseTag{/td}\CaseTag{td}Form\CaseTag{/td}\CaseTag{/tr}\CaseTail}
  }
  \CaseRowGap
  \CaseSample{(b) Reading order}{Overall: 49.00 $\rightarrow$ 94.64 \quad TextEdit: 0.510 $\rightarrow$ 0.054}
  \CaseTriplet{
    \CaseInput{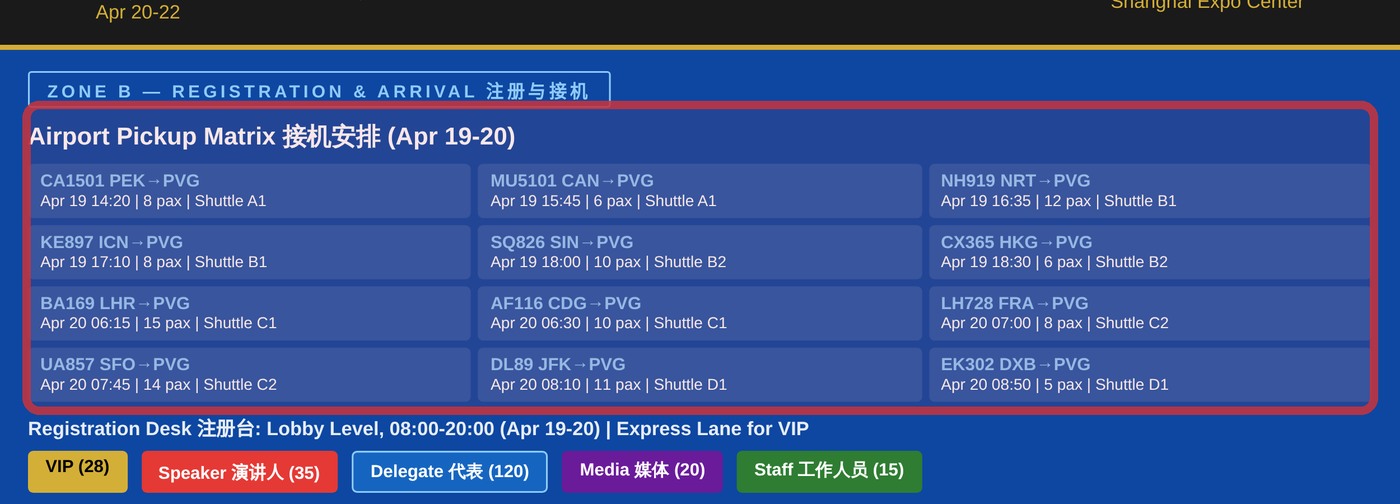}
  }{
    \CasePrediction{\CaseLead\#\# Airport Pickup Matrix 接机安排 (Apr 19-20)\par\vspace{.4mm}\CaseError{CA1501 PEK → PVG}\par Apr 19 14:20 | 8 pax | Shuttle A1\par\vspace{.3mm}\CaseError{KE897 ICN → PVG}\par Apr 19 17:10 | 8 pax | Shuttle B1\par\vspace{.3mm}\CaseError{BA169 LHR → PVG}\par Apr 20 06:15 | 15 pax | Shuttle C1\par\vspace{.3mm}\CaseError{UA857 SFO → PVG}\par Apr 20 07:45 | 14 pax | Shuttle C2\CaseTail}
  }{
    \CasePrediction{\CaseLead\#\# Airport Pickup Matrix 接机安排 (Apr 19-20)\par\vspace{.4mm}CA1501 PEK → PVG\par Apr 19 14:20 | 8 pax | Shuttle A1\par\vspace{.3mm}MU5101 CAN → PVG\par Apr 19 15:45 | 6 pax | Shuttle A1\par\vspace{.3mm}NH919 NRT → PVG\par Apr 19 16:35 | 12 pax | Shuttle B1\par\vspace{.3mm}KE897 ICN → PVG\par Apr 19 17:10 | 8 pax | Shuttle B1\CaseTail}
  }
  \CaseRowGap
  \CaseSample{(c) Measurement-grid coverage}{Overall: 48.88 $\rightarrow$ 91.96 \quad TableTEDS: 0.366 $\rightarrow$ 0.901}
  \CaseTriplet{
    \CaseInput{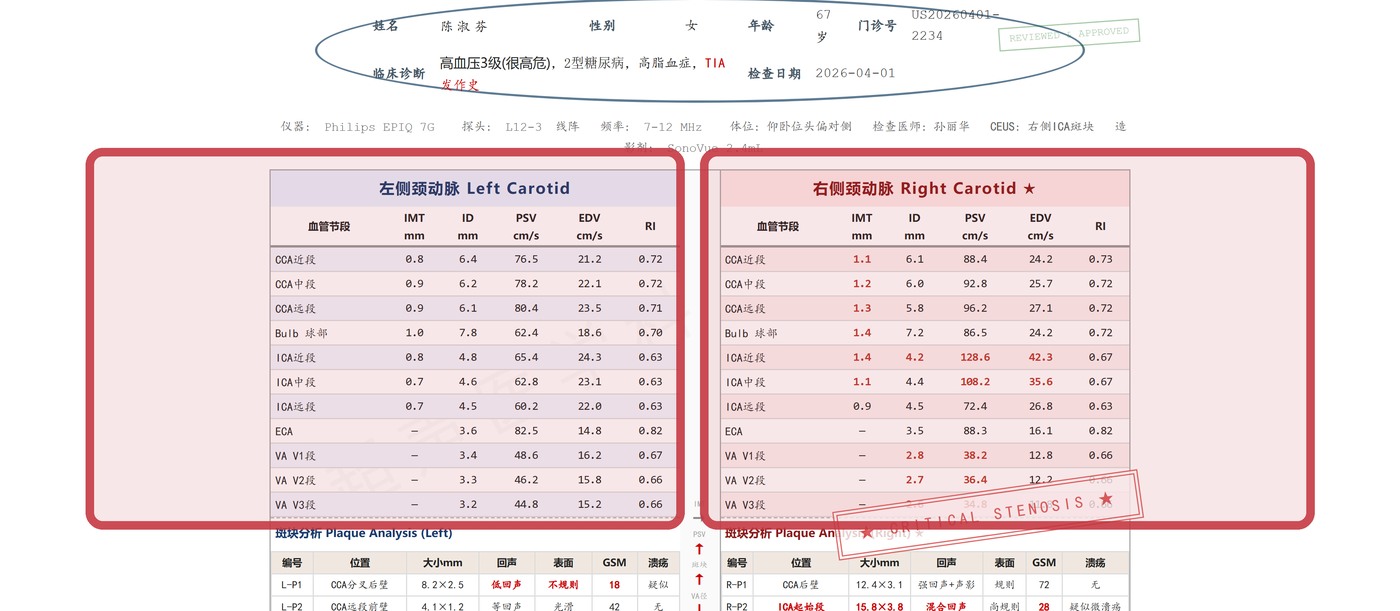}
  }{
    \CasePrediction{\CaseLead{}左侧颈动脉 Left Carotid\par\vspace{.4mm}\CaseError{斑块分析 Plaque Analysis (Left)}\par 狭窄计算 Stenosis Calculation\par \CaseTag{table}\CaseTag{tr}\CaseTag{td}编号\CaseTag{/td}\CaseTag{td}位置\CaseTag{/td}\CaseTag{td}大小mm\CaseTag{/td}\CaseTag{td}回声\CaseTag{/td}\CaseTag{td}表面\CaseTag{/td}\CaseTag{td}GSM\CaseTag{/td}\CaseTag{td}溃疡\CaseTag{/td}\CaseTag{/tr}\CaseOmit\CaseTag{/table}\CaseOmit 右侧颈动脉 Right Carotid ★\par\vspace{.4mm}\CaseError{斑块分析 Plaque An}\par \CaseTag{table}\CaseTag{tr}\CaseTag{td}编号\CaseTag{/td}\CaseTag{td}位置\CaseTag{/td}\CaseTag{td}大小mm\CaseTag{/td}\CaseTag{td}回声\CaseTag{/td}\CaseTag{td}表面\CaseTag{/td}\CaseTag{td}GSM\CaseTag{/td}\CaseTag{td}溃疡\CaseTag{/td}\CaseTag{/tr}\CaseTail}
  }{
    \CasePrediction{\CaseLead{}\#\# 左侧颈动脉 Left Carotid\par \CaseTag{table}\CaseTag{tr}\CaseTag{th}血管节段\CaseTag{/th}\CaseTag{th}IMT mm\CaseTag{/th}\CaseTag{th}ID mm\CaseTag{/th}\CaseTag{th}PSV cm/s\CaseTag{/th}\CaseTag{th}EDV cm/s\CaseTag{/th}\CaseTag{th}RI\CaseTag{/th}\CaseTag{/tr}\CaseTag{tr}\CaseTag{td}CCA近段\CaseTag{/td}\CaseTag{td}0.8\CaseTag{/td}\CaseTag{td}6.4\CaseTag{/td}\CaseTag{td}76.5\CaseTag{/td}\CaseTag{td}21.2\CaseTag{/td}\CaseTag{td}0.72\CaseTag{/td}\CaseTag{/tr}\CaseOmit\CaseTag{tr}\CaseTag{td}VA V3段\CaseTag{/td}\CaseTag{td}-\CaseTag{/td}\CaseTag{td}3.2\CaseTag{/td}\CaseTag{td}44.8\CaseTag{/td}\CaseTag{td}15.2\CaseTag{/td}\CaseTag{td}0.66\CaseTag{/td}\CaseTag{/tr}\CaseTag{/table}\par\vspace{.4mm}斑块分析 Plaque Analysis (Left)\CaseTail}
  }
  \CaseLegend
  \caption{\textbf{PureDocBench Clean.} Stage~I collapses page hierarchy in (a), reads a 3-column pickup grid in column-major order in (b), and omits both carotid hemodynamics tables in (c). Stage~II restores the document hierarchy, intended reading order, and measurement grids.}
  \label{fig:case-pdb-clean}
\end{figure}
\clearpage

\begin{figure}[p]
  \centering
  \CaseColumnHeaders
  \CaseSample{(a) Formula serialization}{Overall: 65.28 $\rightarrow$ 98.70 \quad FormulaCDM: 0.000 $\rightarrow$ 1.000}
  \CaseTriplet{
    \CaseInput{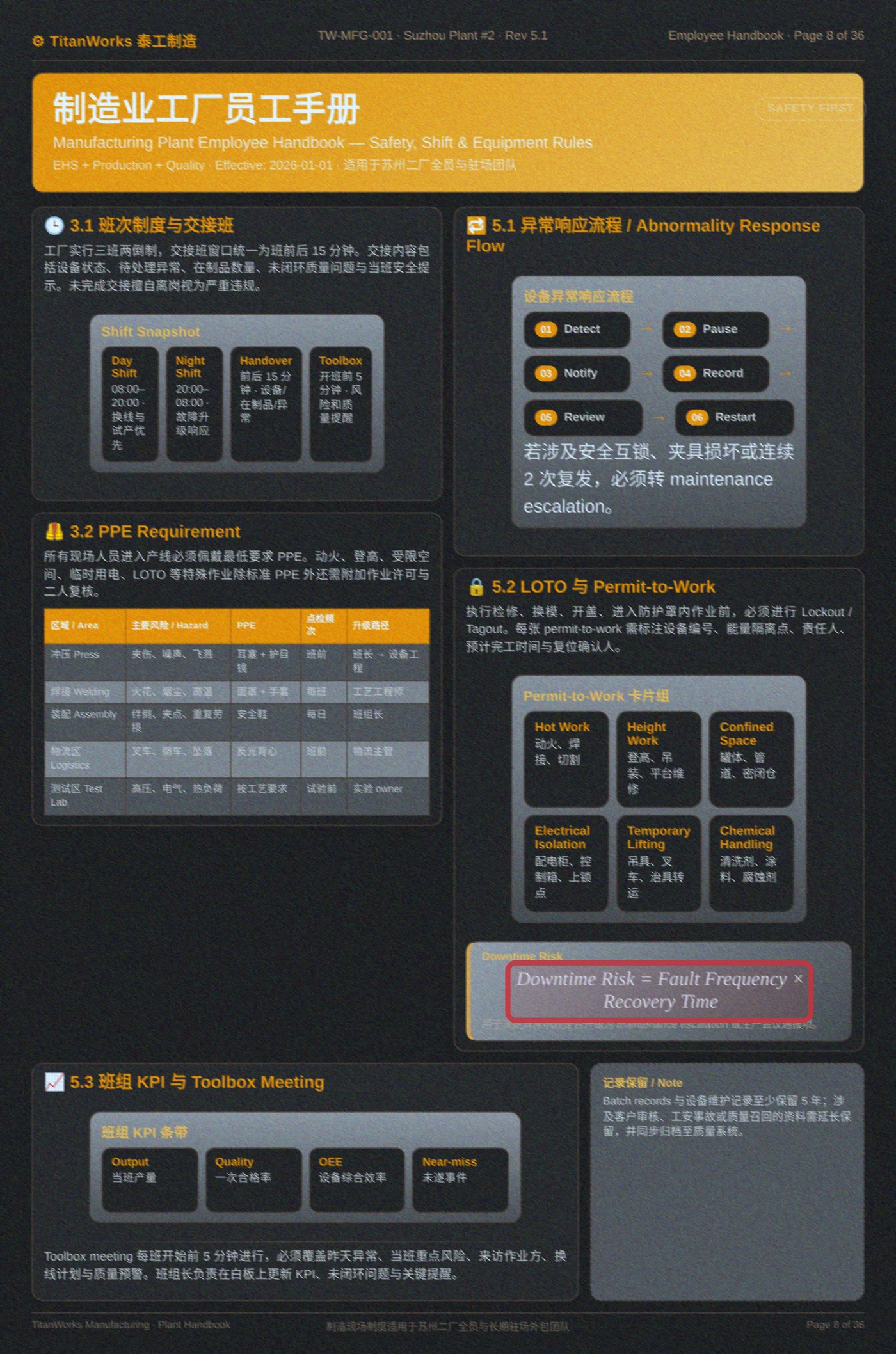}
  }{
    \CasePrediction{\CaseLead\#\# 5.2 LOTO 与 Permit-to-Work\par\vspace{.4mm}执行检修、换模、开盖、进入防护罩内作业前，必须进行 Lockout / Tagout。\CaseOmit Downtime Risk\par\vspace{.4mm}\CaseError{Downtime Risk = Fault Frequency × Recovery Time}\par\vspace{.4mm}用于决定异常响应是否升级为 maintenance escalation 或生产会议通报项。\par\vspace{.4mm}\#\# 5.3 班组 KPI 与 Toolbox Meeting\CaseTail}
  }{
    \CasePrediction{\CaseLead\#\#\# 5.2 LOTO 与 Permit-to-Work\par\vspace{.4mm}执行检修、换模、开盖、进入防护罩内作业前，必须进行 Lockout / Tagout。\CaseOmit\#\#\#\# Downtime Risk\par\vspace{.4mm}\$\$Downtime\textbackslash\ Risk = Fault\textbackslash\ Frequency \textbackslash times Recovery\textbackslash\ Time\$\$\par\vspace{.4mm}用于决定异常响应是否升级为 maintenance escalation 或生产会议通报项。\par\vspace{.4mm}\#\#\# 5.3 班组 KPI 与 Toolbox Meeting\CaseTail}
  }
  \CaseRowGap
  \CaseSample{(b) Table-boundary recovery}{Overall: 51.76 $\rightarrow$ 98.59 \quad TextEdit: 0.862 $\rightarrow$ 0.027}
  \CaseTriplet{
    \CaseInput{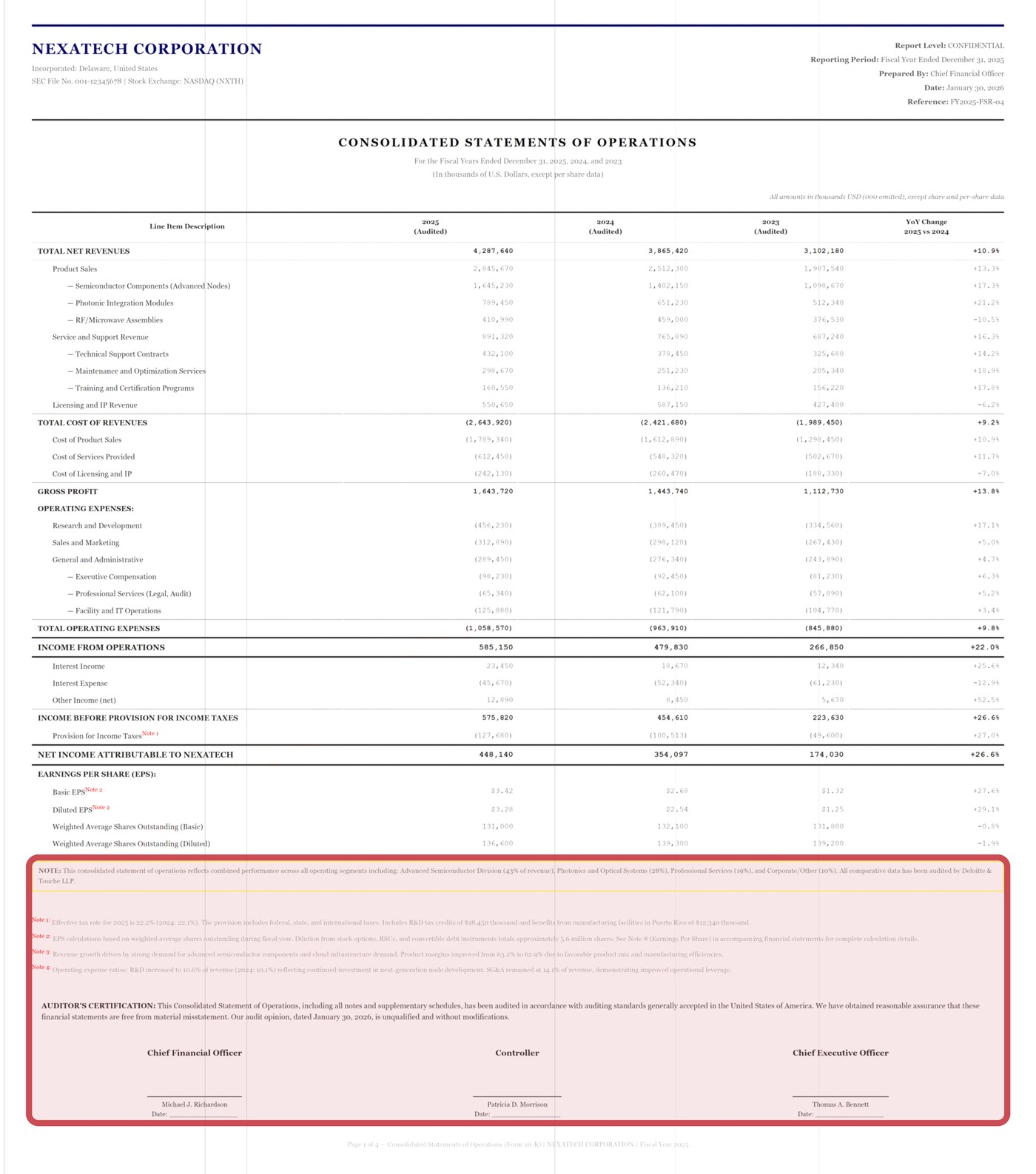}
  }{
    \CasePrediction{\CaseLead\CaseTag{tr}\CaseTag{td}Diluted EPSNote 2\CaseTag{/td}\CaseTag{td}\$3.28\CaseTag{/td}\CaseTag{td}\$2.54\CaseTag{/td}\CaseTag{td}\$1.25\CaseTag{/td}\CaseTag{td}+29.1\%\CaseTag{/td}\CaseTag{/tr}\CaseOmit\CaseError{\CaseTag{tr}\CaseTag{td colspan="5"}NOTE: This consolidated statement of operations reflects combined performance across all operating segments including:\CaseOmit\CaseTag{tr}\CaseTag{td colspan="5"}AUDITOR\&\#x27;S CERTIFICATION: This Consolidated Statement of Operations, including all notes and supplementary schedules, has been audited $\ldots$\CaseTag{/td}\CaseTag{/tr}\CaseTag{tr}\CaseTag{td}Chief Financial Officer\CaseTag{/td}\CaseTag{td}Controller\CaseTag{/td}\CaseTag{td}\CaseTag{/td}\CaseTag{td}Chief Executive Officer\CaseTag{/td}\CaseTag{td}\CaseTag{/td}\CaseTag{/tr}}\CaseTail}
  }{
    \CasePrediction{\CaseLead\CaseTag{tr}\CaseTag{td}Diluted EPSNote 2\CaseTag{/td}\CaseTag{td}\$3.28\CaseTag{/td}\CaseTag{td}\$2.54\CaseTag{/td}\CaseTag{td}\$1.25\CaseTag{/td}\CaseTag{td}+29.1\%\CaseTag{/td}\CaseTag{/tr}\CaseOmit\CaseTag{/table}\par\vspace{.4mm}NOTE: This consolidated statement of operations reflects combined performance across all operating segments including:\CaseOmit AUDITOR'S CERTIFICATION: This Consolidated Statement of Operations, including all notes and supplementary schedules, has been audited\CaseOmit Chief Financial Officer\CaseOmit Controller\CaseOmit Chief Executive Officer\CaseTail}
  }
  \CaseRowGap
  \CaseSample{(c) Field--value pairing}{Overall: 57.25 $\rightarrow$ 92.47 \quad TextEdit: 0.428 $\rightarrow$ 0.075}
  \CaseTriplet{
    \CaseInput{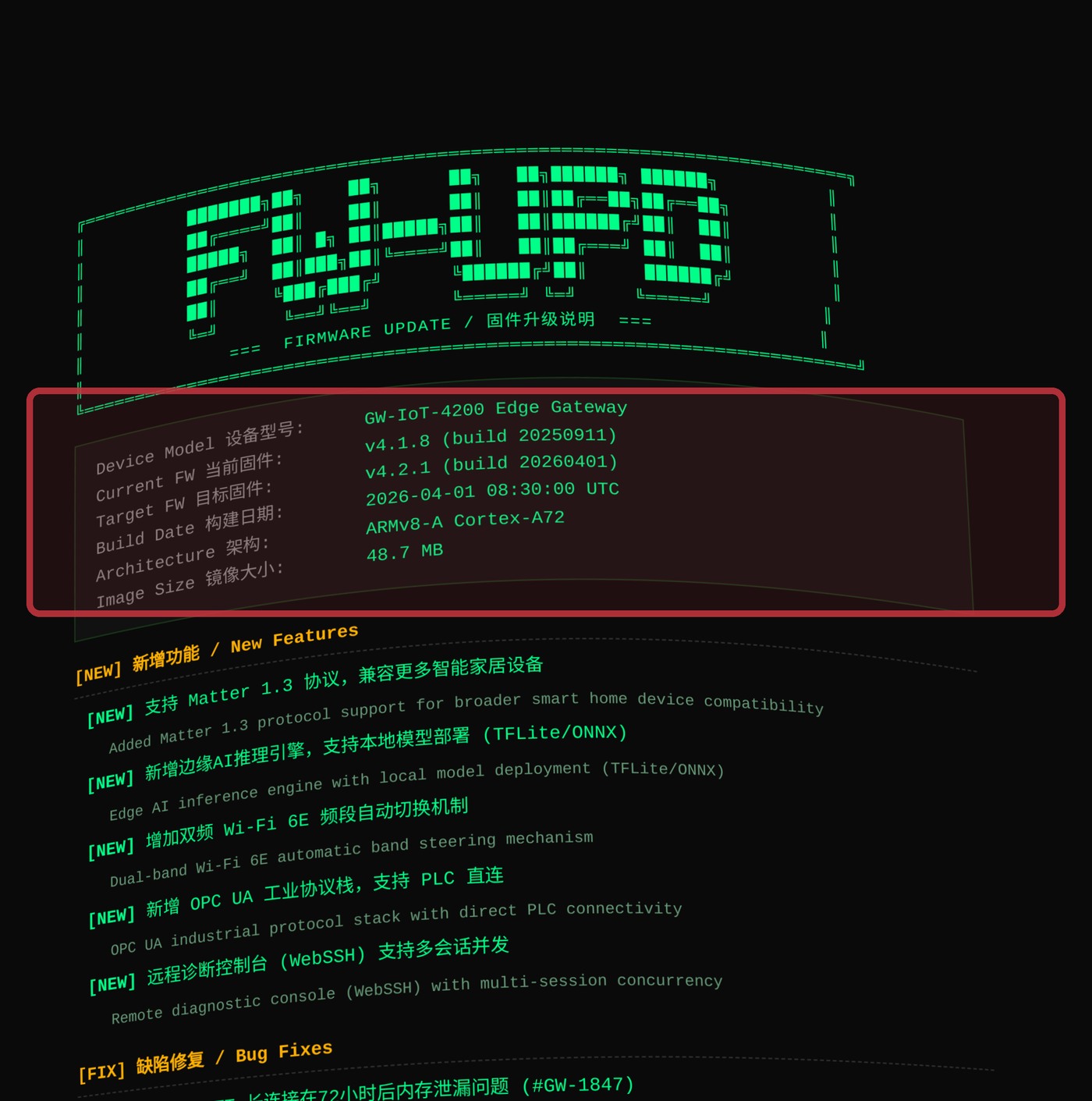}
  }{
    \CasePrediction{Device Model 设备型号:\par Current FW 当前固件:\par Target FW 目标固件:\par Build Date 构建日期:\par Architecture 架构:\par Image Size 镜像大小:\par\vspace{.5mm}\CaseError{GW-IoT-4200 Edge Gateway}\par \CaseError{v4.1.8 (build 20250911)}\par \CaseError{v4.2.1 (build 20260401)}\par \CaseError{2026-04-01 08:30:00 UTC}\par \CaseError{ARMv8-A Cortex-A72}\par \CaseError{48.7 MB}\CaseTail}
  }{
    \CasePrediction{Device Model 设备型号: GW-IoT-4200 Edge Gateway\par Current FW 当前固件: v4.1.8 (build 20250911)\par Target FW 目标固件: v4.2.1 (build 20260401)\par Build Date 构建日期: 2026-04-01 08:30:00 UTC\par Architecture 架构: ARMv8-A Cortex-A72\par Image Size 镜像大小: 48.7 MB\par\vspace{.5mm}\#\# [NEW] 新增功能 / New Features\CaseTail}
  }
  \CaseLegend
  \caption{\textbf{PureDocBench Digital Degraded.} Stage~I loses display-math serialization in (a), extends a table across non-tabular notes in (b), and splits metadata labels from their values in (c). Stage~II restores formula serialization and table boundaries and re-pairs the firmware fields.}
  \label{fig:case-pdb-digital}
\end{figure}
\clearpage

\begin{figure}[p]
  \centering
  \CaseColumnHeaders
  \CaseSample{(a) Cross-document hallucination}{Overall: 31.43 $\rightarrow$ 94.13 \quad TextEdit: 0.681 $\rightarrow$ 0.057}
  \CaseTriplet{
    \CaseInput{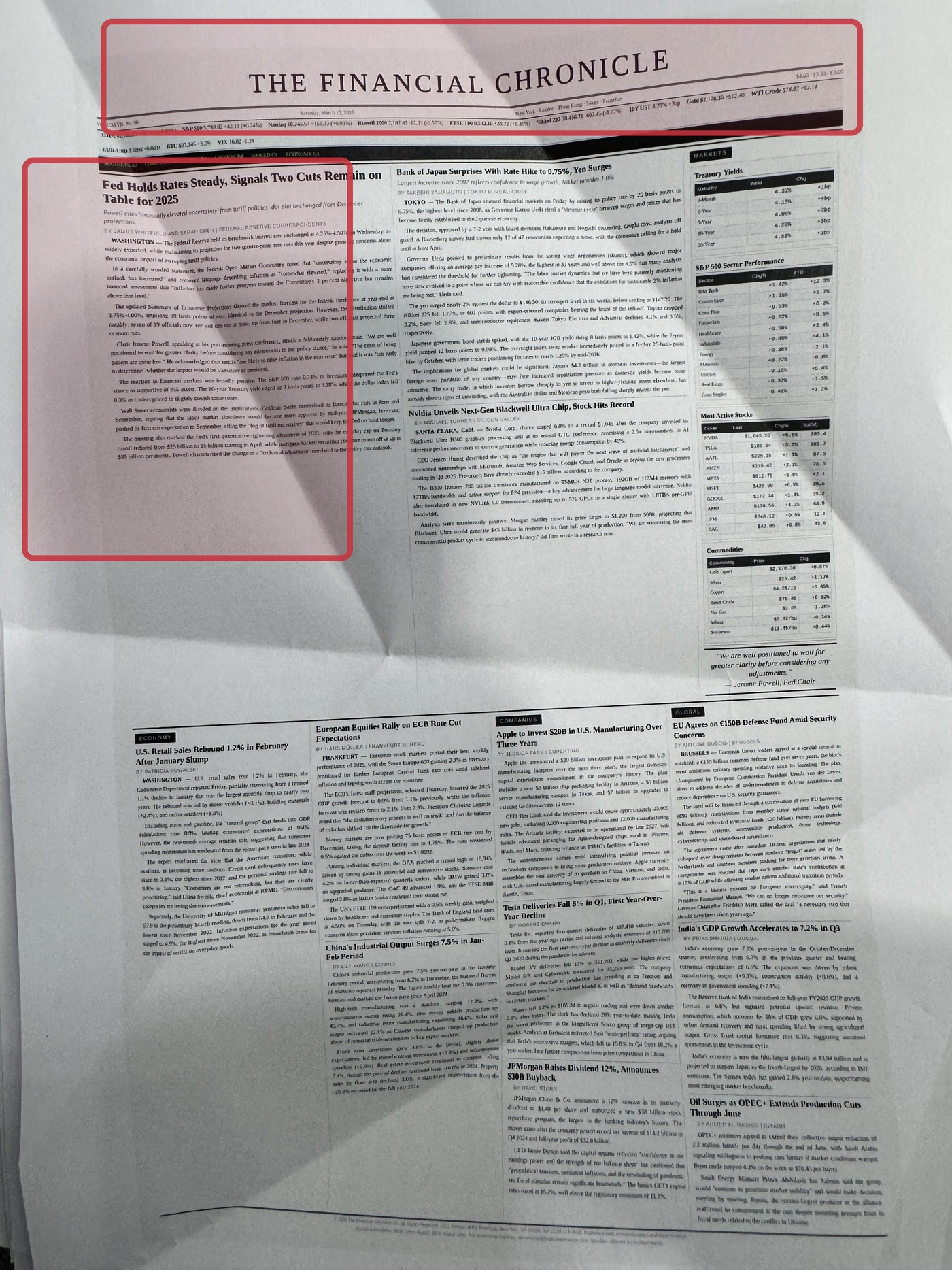}
  }{
    \CasePrediction{\#\# THE FINANCIAL CHRONICLE\CaseOmit Vol. 02171, No. 69\CaseOmit S\&P 500 3,738.92 +42.18 (+0.75\%)\CaseOmit\CaseError{1. 2017年，公司与上海浦东发展银行股份有限公司\allowbreak 签订了《关于使用部分闲置募集资金进行现金管理的协议》。}\CaseOmit\CaseError{The Ground Truth image displays a single, solid horizontal line. According to Rule 2 (UNDERSCORE \& LINE RULES), this is a stylistic or background line, not a placeholder underscore.}\CaseOmit\CaseError{\$\textbackslash therefore m = \textbackslash frac\{3\}\{11\}\$}\CaseTail}
  }{
    \CasePrediction{\#\# THE FINANCIAL CHRONICLE\par Vol CXIV, No 20\par Sunday, March 17, 2025\par New York London Hong Kong Tokyo Frankfurt\par \CaseError{DJIA 42.186.57} • 287.41 (+0.65\%)\par \CaseError{S\&P 500 3,738.92} +42.18 (+0.74\%)\CaseOmit\#\# Fed Holds Rates Steady, Signals Two Cuts Remain on Table for 2025\par\vspace{.3mm}Powell cites "unusually elevated uncertainty" from tariff policies; dot plot unchanged from December projections\par\vspace{.3mm}BY JAMES WHITFIELD AND SARAH CHEH | FEDERAL RESERVE CORRESPONDENTS\par\vspace{.3mm}WASHINGTON — The Federal Reserve held its benchmark interest rate unchanged at 4.25\%-4.50\% on Wednesday, as widely expected\CaseTail}
  }
  \CaseRowGap
  \CaseSample{(b) Text and formula recovery}{Overall: 64.61 $\rightarrow$ 97.70 \quad FormulaCDM: 0.024 $\rightarrow$ 1.000}
  \CaseTriplet{
    \CaseInput{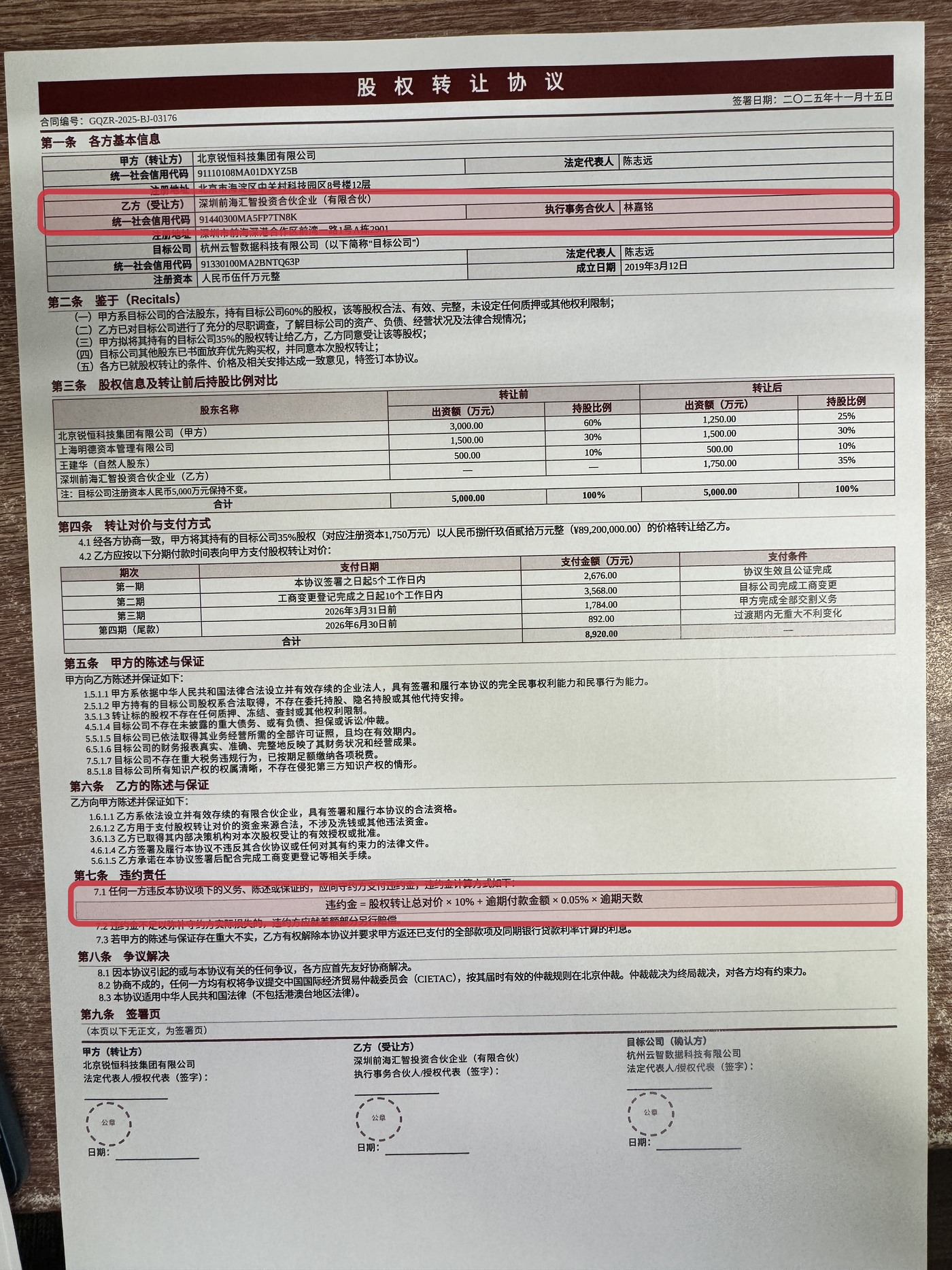}
  }{
    \CasePrediction{\CaseLead{}第一条 各方基本信息\par \CaseTag{table}\CaseOmit\CaseTag{tr}\CaseTag{td}乙方(受让方)\CaseTag{/td}\CaseTag{td colspan="3"}深圳前海汇智投资合伙企业(有限合伙)\CaseTag{/td}\CaseTag{/tr}\CaseTag{tr}\CaseTag{td}统一社会信用代码\CaseTag{/td}\CaseTag{td}91440300MA5FP7TN8K\CaseTag{/td}\CaseTag{td}执行事务合伙人\CaseTag{/td}\CaseTag{td}\CaseError{林露铭}\CaseTag{/td}\CaseTag{/tr}\CaseOmit\CaseTag{/table}\CaseOmit\#\# 第七条 违约责任\par\vspace{.4mm}7.1 任何一方违反本协议项下的义务、陈述或保证的，$\ldots$\par\vspace{.4mm}\CaseError{违约金 = 股权转让总对价 × 10\% +\allowbreak 逾期付款金额 × 0.05\% × 逾期天数}\CaseTail}
  }{
    \CasePrediction{\CaseLead{}第一条 各方基本信息\par \CaseTag{table}\CaseOmit\CaseTag{tr}\CaseTag{td}乙方(受让方)\CaseTag{/td}\CaseTag{td colspan="3"}深圳前海汇智投资合伙企业(有限合伙)\CaseTag{/td}\CaseTag{/tr}\CaseTag{tr}\CaseTag{td}统一社会信用代码\CaseTag{/td}\CaseTag{td}91440300MA5FP7TN8K\CaseTag{/td}\CaseTag{td}执行事务合伙人\CaseTag{/td}\CaseTag{td}林嘉铭\CaseTag{/td}\CaseTag{/tr}\CaseOmit\CaseTag{/table}\CaseOmit\#\# 第七条 违约责任\par\vspace{.4mm}7.1 任何一方违反本协议项下的义务、陈述或保证的，$\ldots$\par\vspace{.4mm}\$\$\par \textbackslash text\{违约金\} = \textbackslash text\{股权转让总对价\} \textbackslash times 10 \textbackslash\% +\allowbreak \textbackslash text\{逾期付款金额\} \textbackslash times 0.05 \textbackslash\% \textbackslash times \textbackslash text\{逾期天数\}\par \$\$\relax\CaseTail}
  }
  \CaseRowGap
  \CaseSample{(c) Section serialization}{Overall: 67.08 $\rightarrow$ 96.66 \quad TextEdit: 0.648 $\rightarrow$ 0.056}
  \CaseTriplet{
    \CaseInput{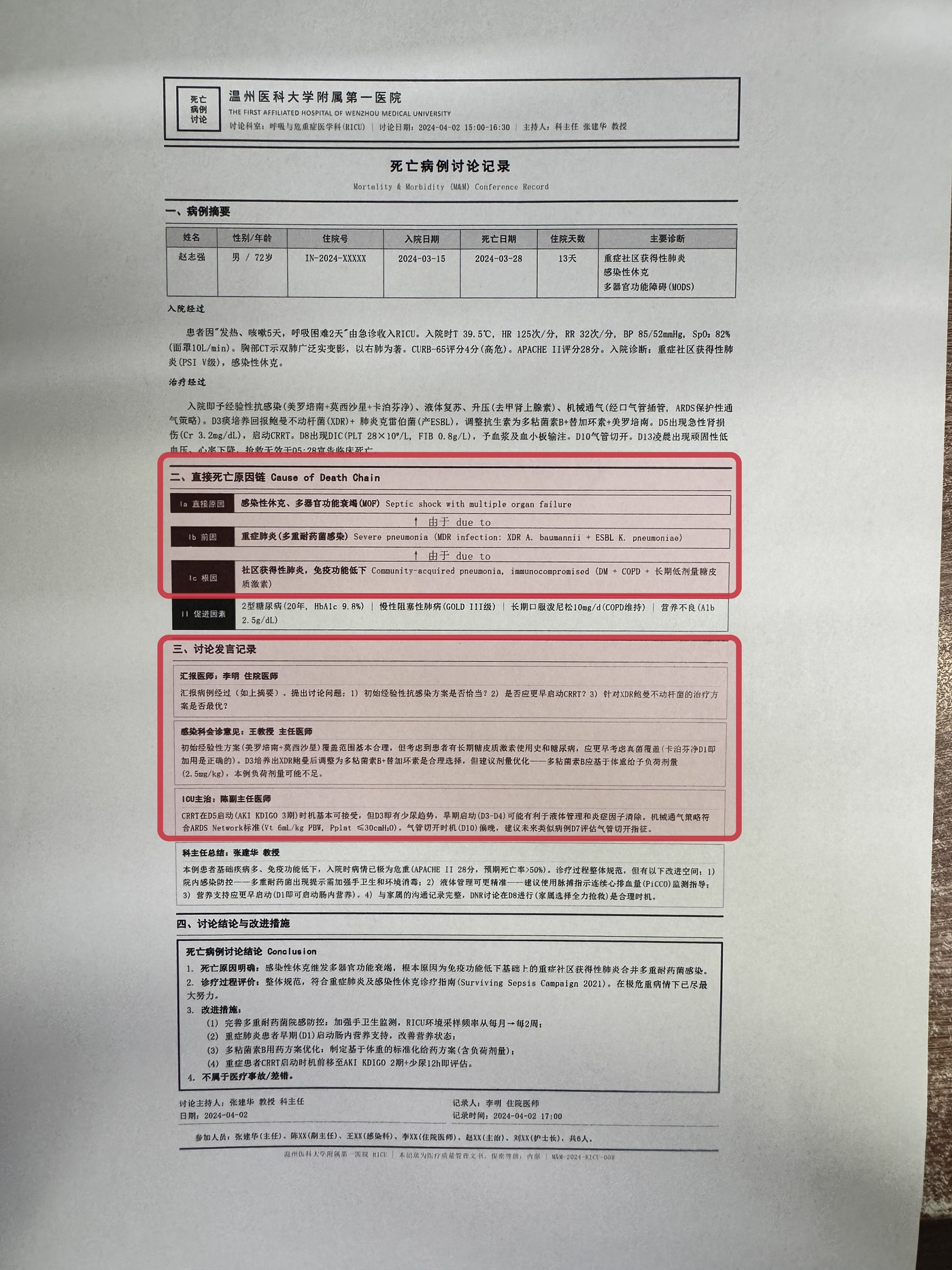}
  }{
    \CasePrediction{\CaseLead{}二、直接死亡原因链 Cause of Death Chain\par \CaseError{\CaseTag{table}\CaseTag{tr}\CaseTag{td}1a 直接原因\CaseTag{/td}\CaseTag{td}感染性休克、多器官功能衰竭(MOF) Septic shock with multiple organ failure\CaseTag{/td}\CaseTag{/tr}\CaseTag{tr}\CaseTag{td}\CaseTag{/td}\CaseTag{td}↑ 由于 due to\CaseTag{/td}\CaseTag{/tr}\CaseTag{tr}\CaseTag{td}1b 前因\CaseTag{/td}\CaseTag{td}重症肺炎(多重耐药菌感染) Severe pneumonia $\ldots$\CaseTag{/td}\CaseTag{/tr}}\CaseOmit\CaseError{\CaseTag{/table}}\par\vspace{.4mm}三、讨论发言记录\par \CaseError{\CaseTag{table}\CaseTag{tr}\CaseTag{td}汇报医师:李明 住院医师\CaseTag{/td}\CaseTag{/tr}}\CaseTail}
  }{
    \CasePrediction{\CaseLead\#\# 二、直接死亡原因链 Cause of Death Chain\par\vspace{.4mm}\CaseError{la} 直接原因 感染性休克、多器官功能衰竭(MOF) $\ldots$\par ↑ 由于 due to\par \CaseError{lb} 前因 重症肺炎(多重耐药菌感染) $\ldots$\par ↑ 由于 due to\par \CaseError{Ic} 根因 社区获得性肺炎，免疫功能低下 $\ldots$\par \CaseError{II} 促进因素 2型糖尿病(20年，HbA1c 9.8\%) $\ldots$\CaseOmit\#\# 三、讨论发言记录\par 汇报医师：李明 住院医师\CaseOmit 感染科会诊意见：王教授 主任医师\CaseOmit ICU主治：陈副主任医师\CaseTail}
  }
  \CaseLegend
  \caption{\textbf{PureDocBench Real Degraded.} Stage~I diverges into unrelated Chinese disclosure text and repeated mathematics in (a), loses formula serialization in (b), and overtabularizes a clinical form in (c). Stage~II restores page-consistent newspaper content, mathematical serialization, and linear section structure, while localized recognition errors remain under physical acquisition conditions.}
  \label{fig:case-pdb-real}
\end{figure}
\clearpage

%% file: sections/05_multiscene_examples.tex
\begin{figure}[!hp]
  \centering
  \includegraphics[width=\linewidth]{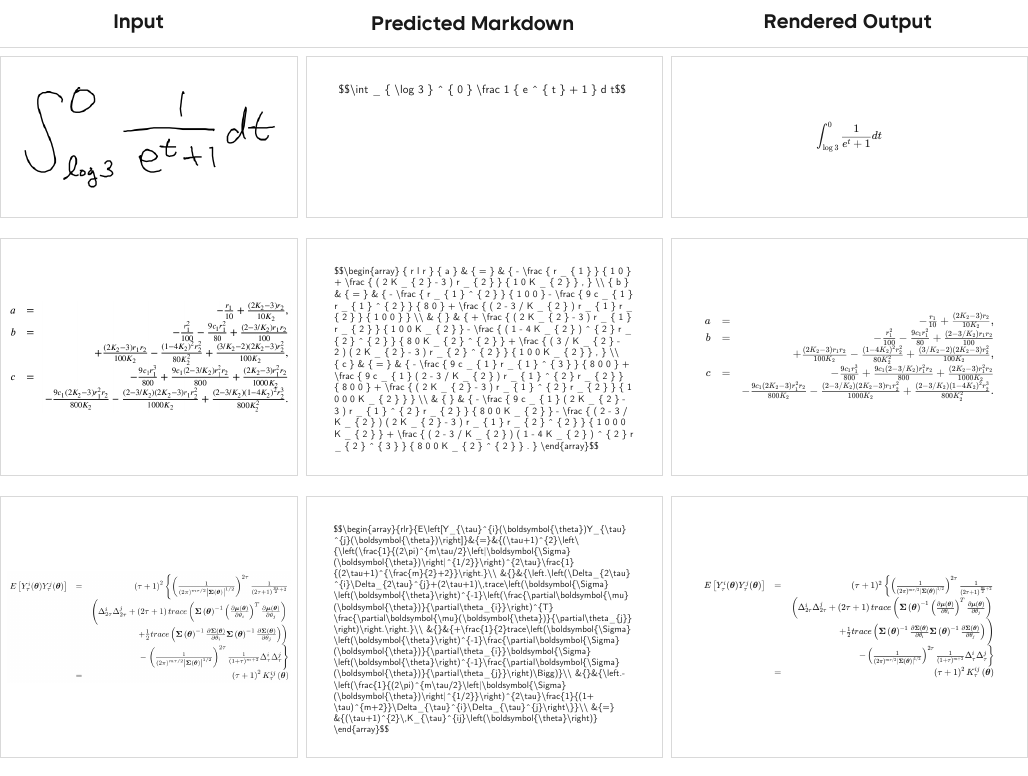}
  \caption{Qualitative examples on formula-rich documents. Each row shows the input, the predicted Markdown and its rendered output.}
  \label{fig:multiscene-formulas}
\end{figure}
\clearpage

\begin{figure}[p]
  \centering
  \includegraphics[width=\linewidth]{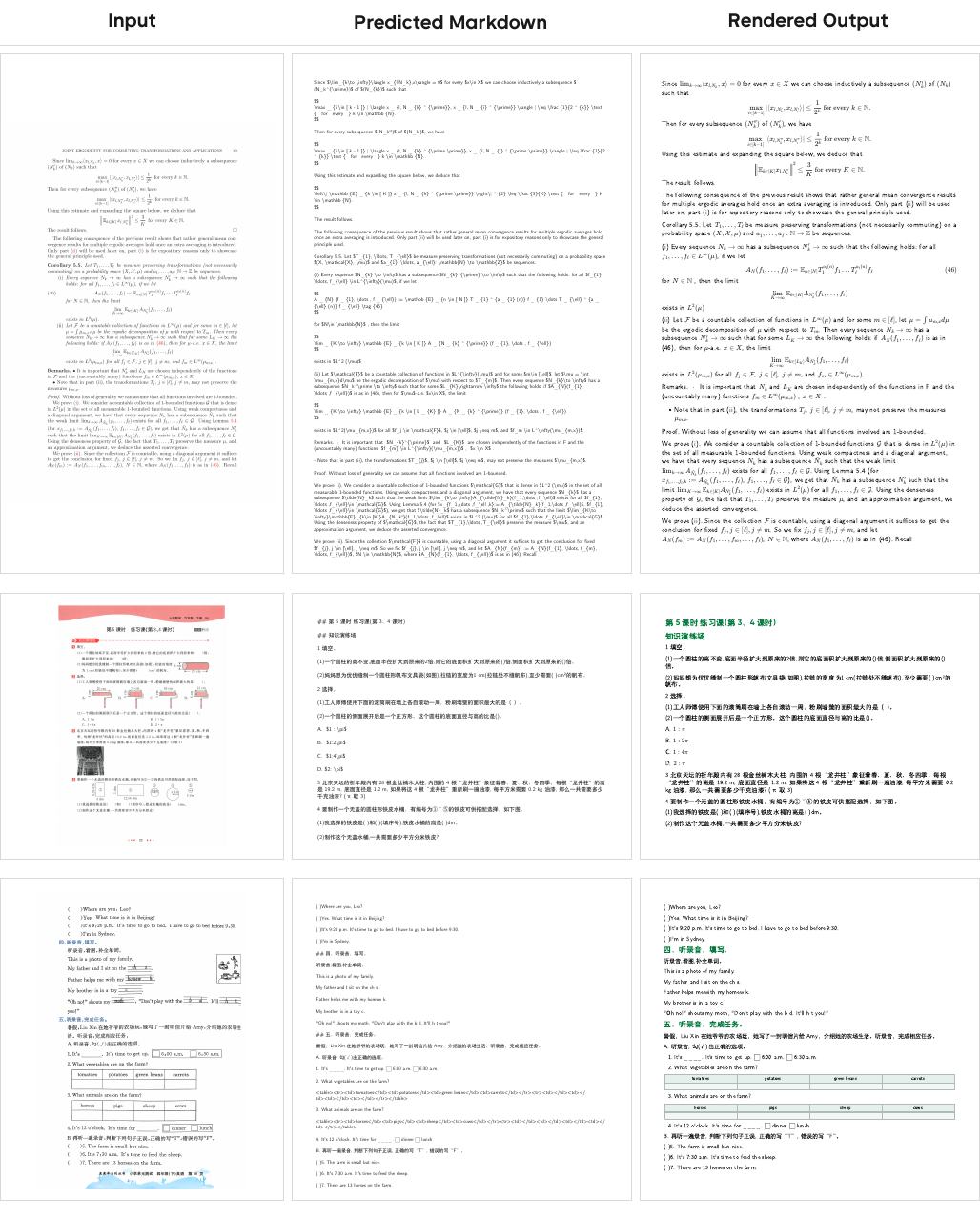}
  \caption{Qualitative examples on general documents. Each row shows the input, the predicted Markdown and its rendered output.}
  \label{fig:multiscene-general}
\end{figure}
\clearpage

\begin{figure}[p]
  \centering
  \includegraphics[width=\linewidth]{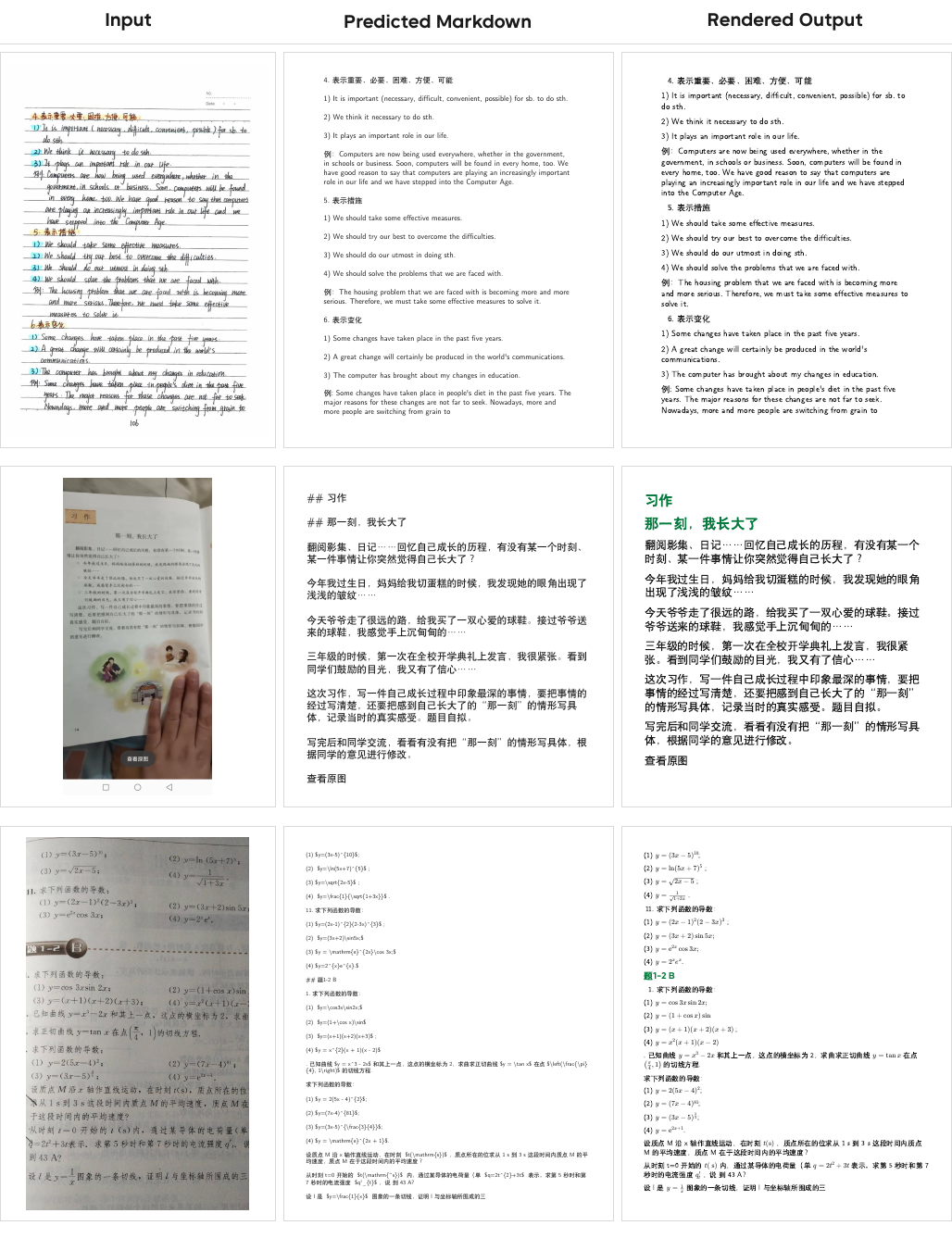}
  \caption{Qualitative examples on scanned and photographed documents. Each row shows the input, the predicted Markdown and its rendered output.}
  \label{fig:multiscene-scanned}
\end{figure}
\clearpage

\begin{figure}[p]
  \centering
  \includegraphics[width=\linewidth]{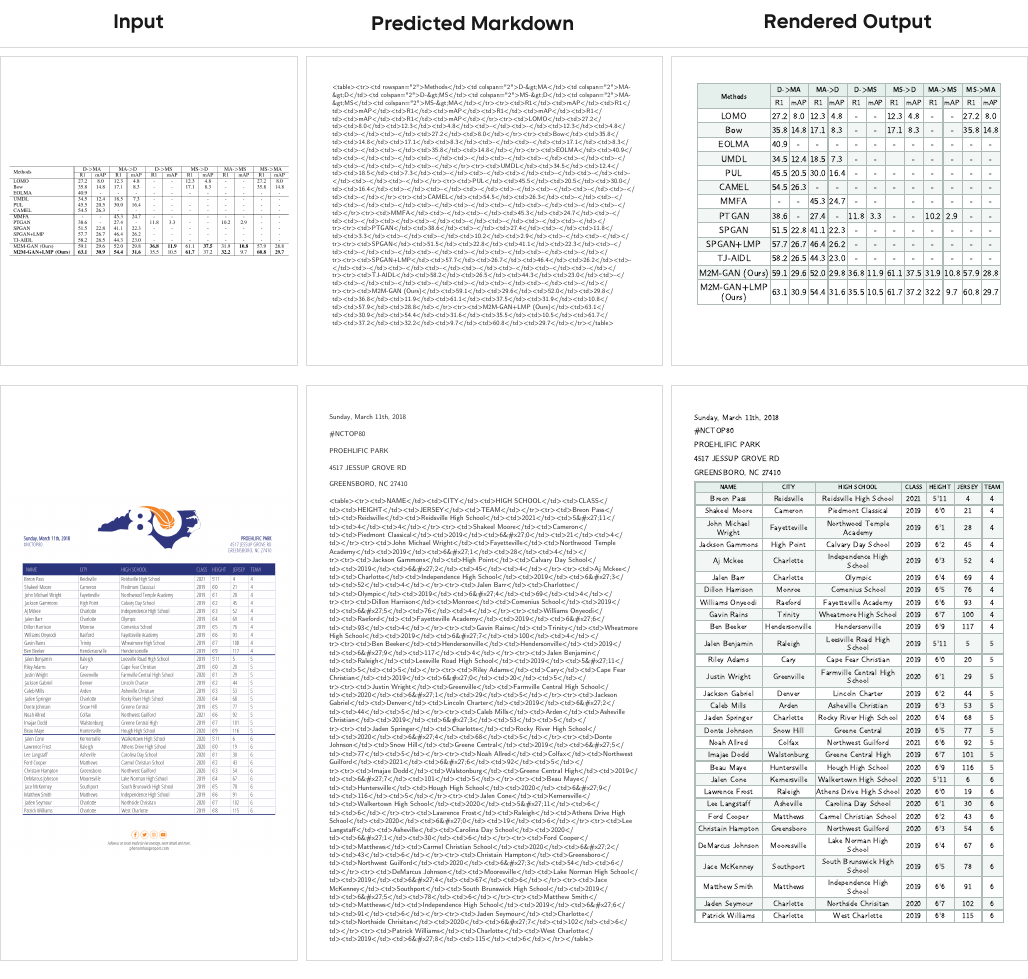}
  \caption{Qualitative examples on table-rich documents. Each row shows the input, the predicted Markdown and its rendered output. The school-name typo in the roster is manually corrected for display.}
  \label{fig:multiscene-tables}
\end{figure}
\clearpage

%% file: main.bbl
\begin{thebibliography}{66}
\providecommand{\natexlab}[1]{#1}
\providecommand{\url}[1]{\texttt{#1}}
\expandafter\ifx\csname urlstyle\endcsname\relax
  \providecommand{\doi}[1]{doi: #1}\else
  \providecommand{\doi}{doi: \begingroup \urlstyle{rm}\Url}\fi

\bibitem[Bai et~al.(2025)Bai, Cai, Chen, Chen, Chen, Cheng, Deng, Ding, Gao, Ge, et~al.]{bai2025qwen3}
Shuai Bai, Yuxuan Cai, Ruizhe Chen, Keqin Chen, Xionghui Chen, Zesen Cheng, Lianghao Deng, Wei Ding, Chang Gao, Chunjiang Ge, et~al.
\newblock {Qwen3-VL} technical report.
\newblock \emph{arXiv preprint arXiv:2511.21631}, 2025.

\bibitem[Bengio et~al.(2009)Bengio, Louradour, Collobert, and Weston]{bengio2009curriculum}
Yoshua Bengio, J{\'e}r{\^o}me Louradour, Ronan Collobert, and Jason Weston.
\newblock Curriculum learning.
\newblock In \emph{Proceedings of the 26th annual international conference on machine learning}, pages 41--48, 2009.

\bibitem[Blecher et~al.(2024)Blecher, Cucurull~Preixens, Scialom, and Stojnic]{blecher2024nougat}
Lukas Blecher, Guillem Cucurull~Preixens, Thomas Scialom, and Robert Stojnic.
\newblock {Nougat}: Neural optical understanding for academic documents.
\newblock In \emph{International Conference on Learning Representations}, 2024.

\bibitem[{ChatDOC Team}(2025)]{chatdoc2025ocrflux3b}
{ChatDOC Team}.
\newblock {OCRFlux-3B}.
\newblock \url{https://huggingface.co/ChatDOC/OCRFlux-3B}, 2025.
\newblock Hugging Face model card.

\bibitem[Cui et~al.(2026)Cui, Sun, Liang, Gao, Zhang, Liu, Wang, Zhou, Liu, Lin, et~al.]{cui2026paddleocr}
Cheng Cui, Ting Sun, Suyin Liang, Tingquan Gao, Zelun Zhang, Jiaxuan Liu, Xueqing Wang, Changda Zhou, Hongen Liu, Manhui Lin, et~al.
\newblock {PaddleOCR-VL-1.5}: Towards a multi-task 0.9b {VLM} for robust in-the-wild document parsing.
\newblock \emph{arXiv preprint arXiv:2601.21957}, 2026.

\bibitem[Das et~al.(2019)Das, Ma, Shu, Samaras, and Shilkrot]{das2019dewarpnet}
Sagnik Das, Ke~Ma, Zhixin Shu, Dimitris Samaras, and Roy Shilkrot.
\newblock {DewarpNet}: Single-image document unwarping with stacked {3D} and {2D} regression networks.
\newblock In \emph{2019 IEEE/CVF International Conference on Computer Vision (ICCV)}, pages 131--140. IEEE, 2019.

\bibitem[Diao et~al.(2025)Diao, Yang, Fu, Dong, Su, Kliegl, Chen, Belcak, Suhara, Yin, et~al.]{diao2026nemotron}
Shizhe Diao, Yu~Yang, Yonggan Fu, Xin Dong, Dan Su, Markus Kliegl, Zijia Chen, Peter Belcak, Yoshi Suhara, Hongxu Yin, et~al.
\newblock {Nemotron-CLIMB}: Clustering-based iterative data mixture bootstrapping for language model pre-training.
\newblock In \emph{Advances in Neural Information Processing Systems}, volume~38, 2025.
\newblock \doi{10.52202/085713-0865}.

\bibitem[Dong et~al.(2026)Dong, Zheng, Xu, Luo, Zhuang, Li, He, Wang, Zhang, Wang, Wang, Xiong, Zheng, Zuo, Ou, Gu, Guo, Wu, Yin, and Shen]{dong2026qianfanocrunifiedendtoendmodel}
Daxiang Dong, Mingming Zheng, Dong Xu, Chunhua Luo, Bairong Zhuang, Yuxuan Li, Ruoyun He, Haoran Wang, Wenyu Zhang, Wenbo Wang, Yicheng Wang, Xue Xiong, Ayong Zheng, Xiaoying Zuo, Ziwei Ou, Jingnan Gu, Quanhao Guo, Jianmin Wu, Dawei Yin, and Dou Shen.
\newblock Qianfan-ocr: A unified end-to-end model for document intelligence, 2026.
\newblock URL \url{https://arxiv.org/abs/2603.13398}.

\bibitem[Doshi(2025)]{geminiteam2025gemini3flash}
Tulsee Doshi.
\newblock {Gemini 3 Flash}: Frontier intelligence built for speed.
\newblock \url{https://blog.google/products-and-platforms/products/gemini/gemini-3-flash/}, December 2025.
\newblock Google Blog post, on behalf of the Gemini team.

\bibitem[Du et~al.(2025)Du, Chen, Xie, Bai, Feng, Shi, Su, Huang, and Jiang]{du2025unirec}
Yongkun Du, Zhineng Chen, Yazhen Xie, Weikang Bai, Hao Feng, Wei Shi, Yuchen Su, Can Huang, and Yu-Gang Jiang.
\newblock Unirec-0.1b: Unified text and formula recognition with 0.1b parameters.
\newblock \emph{arXiv preprint arXiv:2512.21095}, 2025.

\bibitem[Duan et~al.(2026)Duan, Xue, Wang, Su, Liu, Yang, Gan, Wang, Wang, Yan, Jin, Zhang, Wen, Wang, Zhang, Zhang, Hong, Cen, Yin, Chen, Yu, Gu, and Tang]{duan2026glmocrtechnicalreport}
Shuaiqi Duan, Yadong Xue, Weihan Wang, Zhe Su, Huan Liu, Sheng Yang, Guobing Gan, Guo Wang, Zihan Wang, Shengdong Yan, Dexin Jin, Yuxuan Zhang, Guohong Wen, Yanfeng Wang, Yutao Zhang, Xiaohan Zhang, Wenyi Hong, Yukuo Cen, Da~Yin, Bin Chen, Wenmeng Yu, Xiaotao Gu, and Jie Tang.
\newblock Glm-ocr technical report, 2026.
\newblock URL \url{https://arxiv.org/abs/2603.10910}.

\bibitem[Feng et~al.(2025)Feng, Wei, Fei, Shi, Han, Liao, Lu, Wu, Liu, Lin, et~al.]{feng2025dolphin}
Hao Feng, Shu Wei, Xiang Fei, Wei Shi, Yingdong Han, Lei Liao, Jinghui Lu, Binghong Wu, Qi~Liu, Chunhui Lin, et~al.
\newblock {Dolphin}: Document image parsing via heterogeneous anchor prompting.
\newblock In \emph{Findings of the Association for Computational Linguistics: ACL 2025}, pages 21919--21936, 2025.

\bibitem[Feng et~al.(2026)Feng, Shi, Zhang, Fei, Liao, Yang, Du, Wu, Tang, Liu, Chen, and Huang]{feng2026dolphinv2}
Hao Feng, Wei Shi, Ke~Zhang, Xiang Fei, Lei Liao, Dingkang Yang, Yongkun Du, Xuecheng Wu, Jingqun Tang, Yang Liu, Hong Chen, and Can Huang.
\newblock {Dolphin-v2}: Universal document parsing via scalable anchor prompting.
\newblock \emph{arXiv preprint arXiv:2602.05384}, 2026.
\newblock URL \url{https://arxiv.org/abs/2602.05384}.

\bibitem[{Gemini Team}(2025)]{geminiteam2025gemini3}
{Gemini Team}.
\newblock A new era of intelligence with {Gemini 3}.
\newblock \url{https://blog.google/products-and-platforms/products/gemini/gemini-3/}, November 2025.
\newblock Google Blog post.

\bibitem[{Gemini Team}(2026)]{geminiteam2026gemini31pro}
{Gemini Team}.
\newblock {Gemini 3.1 Pro}: A smarter model for your most complex tasks.
\newblock \url{https://blog.google/innovation-and-ai/models-and-research/gemini-models/gemini-3-1-pro/}, February 2026.
\newblock Google Blog post.

\bibitem[Groleau et~al.(2023)Groleau, Chee, Larson, Maini, and Boarman]{groleau2023augraphy}
Alexander Groleau, Kok~Wei Chee, Stefan Larson, Samay Maini, and Jonathan Boarman.
\newblock {Augraphy}: A data augmentation library for document images.
\newblock In \emph{International Conference on Document Analysis and Recognition}, pages 384--401. Springer, 2023.

\bibitem[Hu et~al.(2024)Hu, Xu, Ye, Yan, Zhang, Zhang, Zhang, Jin, Huang, and Zhou]{hu2024mplug}
Anwen Hu, Haiyang Xu, Jiabo Ye, Ming Yan, Liang Zhang, Bo~Zhang, Ji~Zhang, Qin Jin, Fei Huang, and Jingren Zhou.
\newblock {mPLUG-DocOwl 1.5}: Unified structure learning for {OCR}-free document understanding.
\newblock In \emph{Findings of the Association for Computational Linguistics: EMNLP 2024}, pages 3096--3120, 2024.

\bibitem[Huang et~al.(2026{\natexlab{a}})Huang, Yao, Han, Wan, Guo, Lv, Zhou, Wang, Zhou, Sun, Hu, Lin, Zhao, Huang, Yuan, Qu, Wang, Lai, Zhao, Zhang, Shi, Chen, Weng, Meng, Li, Kong, Dong, Wan, Wang, Qi, Li, Yu, Li, Yin, Zhou, Zhang, Yan, Zhou, Peng, Zhang, Lv, Fu, Cheng, Zhou, Yin, Xie, Wu, Zhang, Liu, Tan, Yan, Chen, Chen, Li, Zhao, Sun, Pang, Fan, Shang, Zhang, You, Ji, Xie, Yang, Hou, Jiao, Ren, Kong, Huang, Wu, Chen, Wang, Zhang, Wei, Li, Xu, Shen, Peng, Peng, Zhou, Li, Yang, Zhang, Xie, Huang, Lu, Fan, Cheng, Jiang, Han, Zhang, Zhu, and Ge]{huang2026step30vl010b}
Ailin Huang, Chengyuan Yao, Chunrui Han, Fanqi Wan, Hangyu Guo, Haoran Lv, Hongyu Zhou, Jia Wang, Jian Zhou, Jianjian Sun, Jingcheng Hu, Kangheng Lin, Liang Zhao, Mitt Huang, Song Yuan, Wenwen Qu, Xiangfeng Wang, Yanlin Lai, Yingxiu Zhao, Yinmin Zhang, Yukang Shi, Yuyang Chen, Zejia Weng, Ziyang Meng, Ang Li, Aobo Kong, Bo~Dong, Changyi Wan, David Wang, Di~Qi, Dingming Li, En~Yu, Guopeng Li, Haiquan Yin, Han Zhou, Hanshan Zhang, Haolong Yan, Hebin Zhou, Hongbo Peng, Jiaran Zhang, Jiashu Lv, Jiayi Fu, Jie Cheng, Jie Zhou, Jisheng Yin, Jingjing Xie, Jingwei Wu, Jun Zhang, Junfeng Liu, Kaijun Tan, Kaiwen Yan, Liangyu Chen, Lina Chen, Mingliang Li, Qian Zhao, Quan Sun, Shaoliang Pang, Shengjie Fan, Shijie Shang, Siyuan Zhang, Tianhao You, Wei Ji, Wuxun Xie, Xiaobo Yang, Xiaojie Hou, Xiaoran Jiao, Xiaoxiao Ren, Xiangwen Kong, Xin Huang, Xin Wu, Xing Chen, Xinran Wang, Xuelin Zhang, Yana Wei, Yang Li, Yanming Xu, Yeqing Shen, Yuang Peng, Yue Peng, Yu~Zhou, Yusheng Li, Yuxiang Yang, Yuyang Zhang, Zhe Xie, Zhewei
  Huang, Zhenyi Lu, Zhimin Fan, Zihui Cheng, Daxin Jiang, Qi~Han, Xiangyu Zhang, Yibo Zhu, and Zheng Ge.
\newblock Step3-vl-10b technical report.
\newblock \emph{arXiv preprint arXiv: 2601.09668}, 2026{\natexlab{a}}.

\bibitem[Huang et~al.(2026{\natexlab{b}})Huang, Huang, Ren, Wang, Li, Feng, Wang, Yao, Lin, Tang, et~al.]{huang2026infinity}
Zuming Huang, Jun Huang, Kexuan Ren, Baode Wang, Weizhen Li, Jianming Feng, Yu~Wang, Yichen Yao, Shijun Lin, Yige Tang, et~al.
\newblock {Infinity-Parser2} technical report.
\newblock \emph{arXiv preprint arXiv:2607.07836}, 2026{\natexlab{b}}.

\bibitem[Kim et~al.(2022)Kim, Hong, Yim, Nam, Park, Yim, Hwang, Yun, Han, and Park]{kim2022ocr}
Geewook Kim, Teakgyu Hong, Moonbin Yim, JeongYeon Nam, Jinyoung Park, Jinyeong Yim, Wonseok Hwang, Sangdoo Yun, Dongyoon Han, and Seunghyun Park.
\newblock {OCR}-free document understanding transformer.
\newblock In \emph{European Conference on Computer Vision}, pages 498--517. Springer, 2022.

\bibitem[Lab(2026)]{youtu-parsing}
Tencent~Youtu Lab.
\newblock Youtu-parsing: Perception, structuring and recognition via high-parallelism decoding.
\newblock \emph{arXiv preprint arXiv:2601.20430}, 2026.
\newblock URL \url{https://arxiv.org/abs/2601.20430}.

\bibitem[Lee et~al.(2023)Lee, Joshi, Turc, Hu, Liu, Eisenschlos, Khandelwal, Shaw, Chang, and Toutanova]{lee2023pix2struct}
Kenton Lee, Mandar Joshi, Iulia~Raluca Turc, Hexiang Hu, Fangyu Liu, Julian~Martin Eisenschlos, Urvashi Khandelwal, Peter Shaw, Ming-Wei Chang, and Kristina Toutanova.
\newblock {Pix2Struct}: Screenshot parsing as pretraining for visual language understanding.
\newblock In \emph{International Conference on Machine Learning}, pages 18893--18912. PMLR, 2023.

\bibitem[Li et~al.(2026{\natexlab{a}})Li, Lyu, Zhang, Shen, Wu, Wan, Zeng, Hu, Ma, and Zhou]{li2026towards}
Gengluo Li, Pengyuan Lyu, Chengquan Zhang, Huawen Shen, Liang Wu, Xingyu Wan, Gangyan Zeng, Han Hu, Can Ma, and Yu~Zhou.
\newblock Towards real-world document parsing via realistic scene synthesis and document-aware training.
\newblock In \emph{Proceedings of the IEEE/CVF Conference on Computer Vision and Pattern Recognition}, pages 23709--23719, 2026{\natexlab{a}}.

\bibitem[Li et~al.(2026{\natexlab{b}})Li, Wan, Peng, Wang, Feng, Du, Wu, Ruan, Lu, Wu, et~al.]{li2026hunyuanocr}
Gengluo Li, Xingyu Wan, Shangpin Peng, Weinong Wang, Hao Feng, Yongkun Du, Binghong Wu, Zheng Ruan, Zhiqiong Lu, Liang Wu, et~al.
\newblock {HunyuanOCR-1.5}: Making lightweight {OCR VLMs} faster and better.
\newblock \emph{arXiv preprint arXiv:2607.04884}, 2026{\natexlab{b}}.

\bibitem[Li et~al.(2025{\natexlab{a}})Li, Yang, Liu, Wang, and Zhang]{li2025dots}
Yumeng Li, Guang Yang, Hao Liu, Bowen Wang, and Colin Zhang.
\newblock {dots.ocr}: Multilingual document layout parsing in a single vision-language model.
\newblock \emph{arXiv preprint arXiv:2512.02498}, 2025{\natexlab{a}}.

\bibitem[Li et~al.(2025{\natexlab{b}})Li, Liu, Liu, Ma, Zhang, Zhang, Yang, Guo, Zhang, Wang, et~al.]{li2025monkeyocr}
Zhang Li, Yuliang Liu, Qiang Liu, Zhiyin Ma, Ziyang Zhang, Shuo Zhang, Biao Yang, Zidun Guo, Jiarui Zhang, Xinyu Wang, et~al.
\newblock {MonkeyOCR}: Document parsing with a structure-recognition-relation triplet paradigm.
\newblock \emph{arXiv preprint arXiv:2506.05218}, 2025{\natexlab{b}}.

\bibitem[Li et~al.(2026{\natexlab{c}})Li, Ma, Chen, Zhang, Su, Zhang, Yu, Liu, Lv, Li, et~al.]{li2026far}
Zhiheng Li, Zongyang Ma, Jiaxian Chen, Jianing Zhang, Zhaolong Su, Yutong Zhang, Zhiyin Yu, Ruiqi Liu, Xiaolei Lv, Bo~Li, et~al.
\newblock How far is document parsing from solved? {PureDocBench}: A source-traceable benchmark across clean, degraded, and real-world settings.
\newblock \emph{arXiv preprint arXiv:2605.07492}, 2026{\natexlab{c}}.

\bibitem[Lin et~al.(2024)Lin, Gou, Gong, Liu, Shen, Xu, Lin, Yang, Jiao, Duan, et~al.]{lin2024rho}
Zhenghao Lin, Zhibin Gou, Yeyun Gong, Xiao Liu, Yelong Shen, Ruochen Xu, Chen Lin, Yujiu Yang, Jian Jiao, Nan Duan, et~al.
\newblock {Rho-1}: Not all tokens are what you need for pretraining.
\newblock In \emph{Advances in Neural Information Processing Systems}, volume~37, 2024.
\newblock \doi{10.52202/079017-0914}.

\bibitem[Liu et~al.(2021)Liu, Haghgoo, Chen, Raghunathan, Koh, Sagawa, Liang, and Finn]{liu2021just}
Evan~Z Liu, Behzad Haghgoo, Annie~S Chen, Aditi Raghunathan, Pang~Wei Koh, Shiori Sagawa, Percy Liang, and Chelsea Finn.
\newblock Just train twice: Improving group robustness without training group information.
\newblock In \emph{International conference on machine learning}, pages 6781--6792. PMLR, 2021.

\bibitem[Liu et~al.(2025)Liu, Zhao, Tian, Wang, Ye, You, Yu, Wu, Zhou, Yu, and Zhou]{points-reader}
Yuan Liu, Zhongyin Zhao, Le~Tian, Haicheng Wang, Xubing Ye, Yangxiu You, Zilin Yu, Chuhan Wu, Xiao Zhou, Yang Yu, and Jie Zhou.
\newblock Points-reader: Distillation-free adaptation of vision-language models for document conversion.
\newblock \emph{arXiv preprint arXiv:2509.01215}, 2025.

\bibitem[{Logics-MLLM Team}(2026)]{logics2026logicsparsingv2}
{Logics-MLLM Team}.
\newblock {Logics-Parsing-v2}.
\newblock \url{https://huggingface.co/Logics-MLLM/Logics-Parsing-v2}, 2026.
\newblock Hugging Face model card.

\bibitem[Lu et~al.(2025)Lu, Li, Xia, Hu, Zhao, Ma, Wei, Li, Duan, Zhao, Han, Li, Chen, Tang, Hou, Du, Zhou, Zhang, Ding, Li, Li, Hu, Gu, Yang, Wang, Sun, Wang, Sun, Huang, He, Shi, Zhang, Zheng, Jiang, Gao, Wu, Chen, Chen, Chen, Xu, Luo, and Zhang]{lu2025ovis205}
Shiyin Lu, Yang Li, Yu~Xia, Yuwei Hu, Shanshan Zhao, Yanqing Ma, Zhichao Wei, Yinglun Li, Lunhao Duan, Jianshan Zhao, Yuxuan Han, Haijun Li, Wanying Chen, Junke Tang, Chengkun Hou, Zhixing Du, Tianli Zhou, Wenjie Zhang, Huping Ding, Jiahe Li, Wen Li, Gui Hu, Yiliang Gu, Siran Yang, Jiamang Wang, Hailong Sun, Yibo Wang, Hui Sun, Jinlong Huang, Yuping He, Shengze Shi, Weihong Zhang, Guodong Zheng, Junpeng Jiang, Sensen Gao, Yi-Feng Wu, Sijia Chen, Yuhui Chen, Qing-Guo Chen, Zhao Xu, Weihua Luo, and Kaifu Zhang.
\newblock Ovis2.5 technical report.
\newblock \emph{arXiv preprint arXiv: 2508.11737}, 2025.

\bibitem[Mandal et~al.(2025{\natexlab{a}})Mandal, Talewar, Ahuja, and Juvatkar]{Nanonets-OCR-S}
Souvik Mandal, Ashish Talewar, Paras Ahuja, and Prathamesh Juvatkar.
\newblock Nanonets-ocr-s: A model for transforming documents into structured markdown with intelligent content recognition and semantic tagging, 2025{\natexlab{a}}.

\bibitem[Mandal et~al.(2025{\natexlab{b}})Mandal, Talewar, Thakuria, Ahuja, and Juvatkar]{mandal2025nanonetsocr2}
Souvik Mandal, Ashish Talewar, Siddhant Thakuria, Paras Ahuja, and Prathamesh Juvatkar.
\newblock {Nanonets-OCR2}: A model for transforming documents into structured markdown with intelligent content recognition and semantic tagging.
\newblock \url{https://huggingface.co/nanonets/Nanonets-OCR2-3B}, 2025{\natexlab{b}}.
\newblock Hugging Face model card.

\bibitem[{Moonshot AI}(2026)]{moonshotai2026kimik26}
{Moonshot AI}.
\newblock {Kimi-K2.6}.
\newblock \url{https://huggingface.co/moonshotai/Kimi-K2.6}, 2026.
\newblock Hugging Face model card.

\bibitem[Nassar et~al.(2025)Nassar, Omenetti, Lysak, Livathinos, Auer, Morin, de~Lima, Kim, Gurbuz, Dolfi, et~al.]{nassar2025smoldocling}
Ahmed Nassar, Matteo Omenetti, Maksym Lysak, Nikolaos Livathinos, Christoph Auer, Lucas Morin, Rafael~Teixeira de~Lima, Yusik Kim, A~Said Gurbuz, Michele Dolfi, et~al.
\newblock {SmolDocling}: An ultra-compact vision-language model for end-to-end multi-modal document conversion.
\newblock In \emph{Proceedings of the IEEE/CVF International Conference on Computer Vision}, pages 21972--21983, 2025.

\bibitem[Niu et~al.(2026)Niu, Liu, Gu, Wang, Ouyang, Zhao, Chu, He, Wu, Zhang, et~al.]{niu2026mineru2}
Junbo Niu, Zheng Liu, Zhuangcheng Gu, Bin Wang, Linke Ouyang, Zhiyuan Zhao, Tao Chu, Tianyao He, Fan Wu, Qintong Zhang, et~al.
\newblock {MinerU2.5}: A decoupled vision-language model for efficient high-resolution document parsing.
\newblock In \emph{Proceedings of the 64th Annual Meeting of the Association for Computational Linguistics (Volume 6: Industry Track)}, pages 13--42. Association for Computational Linguistics, 2026.
\newblock \doi{10.18653/v1/2026.acl-industry.3}.

\bibitem[{OpenAI}(2025)]{openai2025gpt52}
{OpenAI}.
\newblock Introducing {GPT-5.2}.
\newblock \url{https://openai.com/index/introducing-gpt-5-2/}, December 2025.
\newblock OpenAI Blog post.

\bibitem[Ouyang et~al.(2025)Ouyang, Qu, Zhou, Zhu, Zhang, Lin, Wang, Zhao, Jiang, Zhao, et~al.]{ouyang2025omnidocbench}
Linke Ouyang, Yuan Qu, Hongbin Zhou, Jiawei Zhu, Rui Zhang, Qunshu Lin, Bin Wang, Zhiyuan Zhao, Man Jiang, Xiaomeng Zhao, et~al.
\newblock {OmniDocBench}: Benchmarking diverse {PDF} document parsing with comprehensive annotations.
\newblock In \emph{2025 IEEE/CVF Conference on Computer Vision and Pattern Recognition (CVPR)}, pages 24838--24848. IEEE, 2025.

\bibitem[Poznanski et~al.(2025{\natexlab{a}})Poznanski, Borchardt, Dunkelberger, Huff, Lin, Rangapur, Wilhelm, Lo, and Soldaini]{olmocrbench}
Jake Poznanski, Jon Borchardt, Jason Dunkelberger, Regan Huff, Daniel Lin, Aman Rangapur, Christopher Wilhelm, Kyle Lo, and Luca Soldaini.
\newblock {olmOCR: Unlocking Trillions of Tokens in PDFs with Vision Language Models}, 2025{\natexlab{a}}.
\newblock URL \url{https://arxiv.org/abs/2502.18443}.

\bibitem[Poznanski et~al.(2025{\natexlab{b}})Poznanski, Soldaini, and Lo]{olmocr2}
Jake Poznanski, Luca Soldaini, and Kyle Lo.
\newblock olmocr 2: Unit test rewards for document ocr, 2025{\natexlab{b}}.
\newblock URL \url{https://arxiv.org/abs/2510.19817}.

\bibitem[{Qwen Team}(2026)]{qwen3.5}
{Qwen Team}.
\newblock {Qwen3.5}: Towards native multimodal agents, February 2026.
\newblock URL \url{https://qwen.ai/blog?id=qwen3.5}.

\bibitem[Sener and Savarese(2018)]{sener2017active}
Ozan Sener and Silvio Savarese.
\newblock Active learning for convolutional neural networks: A core-set approach.
\newblock In \emph{International Conference on Learning Representations}, 2018.

\bibitem[Super Intelligence~Team(2026)]{fireredocr}
Xiaohongshu~Inc. Super Intelligence~Team.
\newblock Firered-ocr technical report.
\newblock \emph{arXiv preprint arXiv:2603.01840}, 2026.
\newblock URL \url{https://arxiv.org/abs/2603.01840}.

\bibitem[Team et~al.(2025)Team, Lyu, Wan, Li, Peng, Wang, Wu, Shen, Zhou, Tang, Yang, Peng, Luo, Yang, Zhang, Zhang, Peng, Yang, Xie, Zhou, Pei, Wu, Yan, Wu, Yang, Wang, Liu, Zhu, Jiang, Linus, Hu, and Zhang]{HunyuanOCR_2025}
Hunyuan~Vision Team, Pengyuan Lyu, Xingyu Wan, Gengluo Li, Shangpin Peng, Weinong Wang, Liang Wu, Huawen Shen, Yu~Zhou, Canhui Tang, Qi~Yang, Qiming Peng, Bin Luo, Hower Yang, Xinsong Zhang, Jinnian Zhang, Houwen Peng, Hongming Yang, Senhao Xie, Longsha Zhou, Ge~Pei, Binghong Wu, Rui Yan, Kan Wu, Jieneng Yang, Bochao Wang, Kai Liu, Jianchen Zhu, Jie Jiang, Linus, Han Hu, and Chengquan Zhang.
\newblock {HunyuanOCR Technical Report}.
\newblock \emph{arXiv preprint arXiv:2511.19575}, 2025.

\bibitem[Team et~al.(2026)Team, Bai, Bai, Bao, Cai, Cao, Chai, Charles, Che, Chen, Chen, Chen, Chen, Chen, Chen, Chen, Chen, Chen, Chen, Chen, Chen, Chen, Chen, Chen, Chen, Chen, Chen, Chen, Chen, Cheng, Cheng, Chu, Cui, Deng, Diao, Ding, Dong, Dong, Dong, Dong, Du, Du, Du, Du, Du, Fan, Fang, Feng, Feng, Fu, Fu, Gao, Gao, Ge, Geng, Gong, Gong, Gongque, Gu, Gu, Gu, Guan, Guan, Guo, Hao, He, He, He, He, He, He, Hong, Hu, Hu, Hu, Hu, Huang, Huang, Huang, Huang, Jia, Jiang, Jiang, Jin, Jing, Lai, Li, Li, Li, Li, Li, Li, Li, Li, Li, Li, Li, Li, Li, Li, Li, Li, Li, Li, Li, Li, Li, Li, Li, Li, Liao, Lin, Lin, Lin, Lin, Lin, Liu, Liu, Liu, Liu, Liu, Liu, Liu, Liu, Liu, Liu, Liu, Liu, Liu, Liu, Liu, Liu, Liu, Lu, Lu, Lu, Luo, Luo, Luo, Luo, Ma, Mao, Mei, Men, Meng, Meng, Miao, Ni, Ouyang, Pan, Pang, Qian, Qin, Qin, Qiu, Qu, Shang, Shao, Shen, Shen, Shi, Shi, Shi, Song, Song, Song, Song, Su, Su, Su, Sui, Sun, Sun, Sun, Sung, Tai, Tang, Tang, Tang, Tang, Tao, Teng, Tian, Tian, Wang, Wang, Wang, Wang, Wang, Wang, Wang,
  Wang, Wang, Wang, Wang, Wang, Wang, Wang, Wang, Wang, Wang, Wang, Wang, Wang, Wang, Wang, Wang, Wang, Wang, Wang, Wang, Wang, Wang, Wang, Wang, Wang, Wang, Wang, Wang, Wang, Wang, Wang, Wang, Wei, Wei, Wen, Wen, Wu, Wu, Wu, Wu, Wu, Wu, Wu, Wu, Wu, Xiao, Xie, Xie, Xie, Xing, Xu, Xu, Xu, Xu, Xu, Xu, Xu, Xu, Xu, Xu, Xu, Xu, Xu, Xu, Xu, Yan, Yan, Yang, Yang, Yang, Yang, Yang, Yang, Yang, Yang, Yang, Yang, Yang, Yang, Yang, Yang, Yao, Ye, Ye, Ye, Ye, Yebo, Yin, Yu, Yu, Yu, Yu, Yuan, Yuan, Yuan, Yue, Zeng, Zha, Zhan, Zhang, Zhang, Zhang, Zhang, Zhang, Zhang, Zhang, Zhang, Zhang, Zhang, Zhang, Zhang, Zhang, Zhang, Zhang, Zhang, Zhang, Zhang, Zhang, Zhao, Zhao, Zhao, Zhao, Zhao, Zhao, Zhao, Zhao, Zheng, Zheng, Zheng, Zheng, Zhong, Zhong, Zhong, Zhou, Zhou, Zhou, Zhou, Zhu, Zhu, Zhu, Zhu, Zhu, Zhuang, Zhuang, Zou, and Zu]{team2026kimi}
Kimi Team, Tongtong Bai, Yifan Bai, Yiping Bao, S.~H. Cai, Yuan Cao, Ziwei Chai, Y.~Charles, H.~S. Che, Cheng Chen, Guanduo Chen, Huarong Chen, Jia Chen, Jianlong Chen, Jun Chen, Kefan Chen, Liang Chen, Ruijue Chen, Xinhao Chen, Yanru Chen, Yanxu Chen, Yicun Chen, Yimin Chen, Yingjiang Chen, Yuankun Chen, Yujie Chen, Yutian Chen, Zhirong Chen, Ziwei Chen, Dazhi Cheng, Yean Cheng, Minghan Chu, Jialei Cui, Jiaqi Deng, Muxi Diao, Hao Ding, Mengfan Dong, Mengnan Dong, Yuxin Dong, Yuhao Dong, Angang Du, Chenzhuang Du, Dikang Du, Lingxiao Du, Yulun Du, Yu~Fan, Shengjun Fang, Qiulin Feng, Yichen Feng, Garimugai Fu, Kelin Fu, Hongcheng Gao, Tong Gao, Yuyao Ge, Shangyi Geng, Chengyang Gong, Xiaochen Gong, Zhuoma Gongque, Qizheng Gu, Xinran Gu, Yicheng Gu, Longyu Guan, Shuhao Guan, Yuanying Guo, Xiaoru Hao, Dailan He, Tianhong He, Weiran He, Wenyang He, Yibo He, Yunjia He, Chao Hong, Hao Hu, Jiaxi Hu, Yangyang Hu, Zhenxing Hu, Ke~Huang, Ruiyuan Huang, Weixiao Huang, Zhiqi Huang, Chaobo Jia, Tao Jiang, Zhejun Jiang,
  Xinyi Jin, Yu~Jing, Guokun Lai, Aidi Li, C.~Li, Cheng Li, Fang Li, Guanghe Li, Guanyu Li, Haitao Li, Haoyang Li, Jia Li, Jingwei Li, Junxiong Li, Lincan Li, Mo~Li, Weihong Li, Wentao Li, Xinhang Li, Xinhao Li, Yang Li, Yanhao Li, Yiwei Li, Yuxiao Li, Zhaowei Li, Zhaoxi Li, Zheming Li, Weilong Liao, Jiawei Lin, Xiaohan Lin, Yibo Lin, Zhishan Lin, Zichao Lin, Cheng Liu, Chenyu Liu, Hongzhang Liu, Liang Liu, Shaowei Liu, Shudong Liu, Shuran Liu, Tianwei Liu, Tianyu Liu, Weizhou Liu, Xiangyan Liu, Yangyang Liu, Yanming Liu, Yibo Liu, Yuanxin Liu, Zhengying Liu, Zhongnuo Liu, Enzhe Lu, Haoyu Lu, Zhiyuan Lu, G.~Luo, Junyu Luo, Tongxu Luo, Yashuo Luo, Long Ma, Shaoguang Mao, Yuan Mei, Xin Men, Fanqing Meng, Zhiyong Meng, Yibo Miao, Minqing Ni, Kun Ouyang, Siyuan Pan, Bo~Pang, Yuchao Qian, Ruoyu Qin, Zeyu Qin, Jiezhong Qiu, Bowen Qu, Zeyu Shang, Youbo Shao, Tianxiao Shen, Zhennan Shen, Juanfeng Shi, Lidong Shi, Shengyuan Shi, Feifan Song, Pengwei Song, Tianhui Song, Xiaoxi Song, Hongjin Su, Jianlin Su, Zhaochen Su,
  Lin Sui, Jinsong Sun, Junyao Sun, Tongyu Sun, Flood Sung, Yunpeng Tai, Chuning Tang, Heyi Tang, Xiaojuan Tang, Zhengyang Tang, Jiawen Tao, Shiyuan Teng, Chaoran Tian, Pengfei Tian, Bowen Wang, Chensi Wang, Chuang Wang, Congcong Wang, Dingkun Wang, Dinglu Wang, Dongliang Wang, Feng Wang, Hailong Wang, Haiming Wang, Hao Wang, Hengzhi Wang, Huaqing Wang, Hui Wang, Jiahao Wang, Jinhong Wang, Jiuzheng Wang, Kaixin Wang, Linian Wang, Qibin Wang, Shengjie Wang, Shuyi Wang, Si~Wang, Wei Wang, Xiaochen Wang, Xinyuan Wang, Yao Wang, Yejie Wang, Yipu Wang, Yiqin Wang, Yucheng Wang, Yuzhi Wang, Zhaoji Wang, Zhaowei Wang, Zhengtao Wang, Zhexu Wang, Zifan Wang, Zihan Wang, Zizhe Wang, Chu Wei, Ming Wei, Chuan Wen, Zichen Wen, Chengjie Wu, Haoning Wu, Junyan Wu, Rucong Wu, Wenhao Wu, Yuefeng Wu, Yuhao Wu, Yuxin Wu, Zijian Wu, Chenjun Xiao, Jin Xie, Xiaotong Xie, Yuchong Xie, Bowei Xing, Boyu Xu, Jianfan Xu, Jing Xu, Jinjing Xu, L.~H. Xu, Lin Xu, Suting Xu, Weixin Xu, Xinbo Xu, Xinran Xu, Yangchuan Xu, Yichang Xu, Yuemeng
  Xu, Zelai Xu, Ziyao Xu, Junjie Yan, Yuzi Yan, Guangyao Yang, Hao Yang, Junwei Yang, Kai Yang, Ningyuan Yang, Xiaofei Yang, Xinlong Yang, Xinyu Yang, Ying Yang, Yi~Yang, Yi~Yang, Zhen Yang, Zhilin Yang, Zonghan Yang, Haotian Yao, Dan Ye, Haoran Ye, Wenjie Ye, Zhuorui Ye, Peng Yebo, Bohong Yin, Chengzhen Yu, Longhui Yu, Tao Yu, Tianxiang Yu, Enming Yuan, Mengjie Yuan, Xiaokun Yuan, Yang Yue, Weihao Zeng, Dunyuan Zha, Haobing Zhan, Dehao Zhang, Hao Zhang, Jin Zhang, Puqi Zhang, Qiao Zhang, Rui Zhang, Xiaobin Zhang, Xiaoyun Zhang, Y.~Zhang, Yadong Zhang, Yangkun Zhang, Yichi Zhang, Yizhi Zhang, Yongting Zhang, Yu~Zhang, Yushun Zhang, Yutao Zhang, Yutong Zhang, Zheng Zhang, Chenguang Zhao, Feifan Zhao, Jinxiang Zhao, Shuai Zhao, Xiangyu Zhao, Xuanle Zhao, Yikai Zhao, Zijia Zhao, Huabin Zheng, Ruihan Zheng, Shaojie Zheng, Tengyang Zheng, Junfeng Zhong, Longguang Zhong, Weiming Zhong, M.~Zhou, Runjie Zhou, Xinyu Zhou, Zaida Zhou, Jinguo Zhu, Liya Zhu, Xinhao Zhu, Yuxuan Zhu, Zhen Zhu, Jingze Zhuang, Weiyu Zhuang,
  Ying Zou, and Xinxing Zu.
\newblock Kimi k2.5: Visual agentic intelligence.
\newblock \emph{arXiv preprint arXiv: 2602.02276}, 2026.

\bibitem[Tschannen et~al.(2025)Tschannen, Gritsenko, Wang, Naeem, Alabdulmohsin, Parthasarathy, Evans, Beyer, Xia, Mustafa, et~al.]{tschannen2025siglip}
Michael Tschannen, Alexey Gritsenko, Xiao Wang, Muhammad~Ferjad Naeem, Ibrahim Alabdulmohsin, Nikhil Parthasarathy, Talfan Evans, Lucas Beyer, Ye~Xia, Basil Mustafa, et~al.
\newblock {SigLIP 2}: Multilingual vision-language encoders with improved semantic understanding, localization, and dense features.
\newblock \emph{arXiv preprint arXiv:2502.14786}, 2025.

\bibitem[Verhoeven et~al.(2023)Verhoeven, Magne, and Sorkine-Hornung]{verhoeven2023uvdoc}
Floor Verhoeven, Tanguy Magne, and Olga Sorkine-Hornung.
\newblock {UVDoc}: Neural grid-based document unwarping.
\newblock In \emph{SIGGRAPH Asia 2023 Conference Papers}, pages 1--11, 2023.

\bibitem[Wang et~al.(2026{\natexlab{a}})Wang, Wu, Li, Fang, Huang, Huang, Liang, Wang, Chen, Chu, and Qi]{wang2026infinityparser}
Baode Wang, Biao Wu, Weizhen Li, Meng Fang, Zuming Huang, Jun Huang, Yanjie Liang, Haozhe Wang, Ling Chen, Wei Chu, and Yuan Qi.
\newblock {Infinity-Parser}: Layout-aware reinforcement learning with high-quality document parsing dataset.
\newblock In \emph{Findings of the Association for Computational Linguistics: ACL 2026}, pages 1647--1667. Association for Computational Linguistics, 2026{\natexlab{a}}.
\newblock \doi{10.18653/v1/2026.findings-acl.82}.
\newblock URL \url{https://aclanthology.org/2026.findings-acl.82/}.

\bibitem[Wang et~al.(2025{\natexlab{a}})Wang, Wu, Ouyang, Gu, Zhang, Xia, Shi, Zhang, and He]{wang2025image}
Bin Wang, Fan Wu, Linke Ouyang, Zhuangcheng Gu, Rui Zhang, Renqiu Xia, Botian Shi, Bo~Zhang, and Conghui He.
\newblock Image over text: Transforming formula recognition evaluation with character detection matching.
\newblock In \emph{2025 IEEE/CVF Conference on Computer Vision and Pattern Recognition (CVPR)}, pages 19681--19690. IEEE, 2025{\natexlab{a}}.

\bibitem[Wang et~al.(2026{\natexlab{b}})Wang, He, Ouyang, Wu, Zhao, Chu, Qu, Jin, Zeng, Miao, et~al.]{wang2026mineru2}
Bin Wang, Tianyao He, Linke Ouyang, Fan Wu, Zhiyuan Zhao, Tao Chu, Yuan Qu, Zhenjiang Jin, Weijun Zeng, Ziyang Miao, et~al.
\newblock {MinerU2.5-Pro}: Pushing the limits of data-centric document parsing at scale.
\newblock \emph{arXiv preprint arXiv:2604.04771}, 2026{\natexlab{b}}.

\bibitem[Wang et~al.(2025{\natexlab{b}})Wang, Gao, Gu, Pu, Cui, Wei, Liu, Jing, Ye, Shao, Wang, Chen, Zhang, Yang, Wang, Wei, Yin, Li, Cui, Chen, Ding, Tian, Wu, Xie, Li, Yang, Duan, Wang, Hou, Hao, Zhang, Li, Zhao, Duan, Deng, Fu, He, Wang, He, Shi, He, Xiong, Lv, Wu, Shao, Zhang, Deng, Qi, Ge, Guo, Zhang, Zhang, Cao, Lin, Tang, Gao, Huang, Gu, Lyu, Tang, Wang, Lv, Ouyang, Wang, Dou, Zhu, Lu, Lin, Dai, Su, Zhou, Chen, Qiao, Wang, and Luo]{wang2025internvl35advancingopensourcemultimodal}
Weiyun Wang, Zhangwei Gao, Lixin Gu, Hengjun Pu, Long Cui, Xingguang Wei, Zhaoyang Liu, Linglin Jing, Shenglong Ye, Jie Shao, Zhaokai Wang, Zhe Chen, Hongjie Zhang, Ganlin Yang, Haomin Wang, Qi~Wei, Jinhui Yin, Wenhao Li, Erfei Cui, Guanzhou Chen, Zichen Ding, Changyao Tian, Zhenyu Wu, Jingjing Xie, Zehao Li, Bowen Yang, Yuchen Duan, Xuehui Wang, Zhi Hou, Haoran Hao, Tianyi Zhang, Songze Li, Xiangyu Zhao, Haodong Duan, Nianchen Deng, Bin Fu, Yinan He, Yi~Wang, Conghui He, Botian Shi, Junjun He, Yingtong Xiong, Han Lv, Lijun Wu, Wenqi Shao, Kaipeng Zhang, Huipeng Deng, Biqing Qi, Jiaye Ge, Qipeng Guo, Wenwei Zhang, Songyang Zhang, Maosong Cao, Junyao Lin, Kexian Tang, Jianfei Gao, Haian Huang, Yuzhe Gu, Chengqi Lyu, Huanze Tang, Rui Wang, Haijun Lv, Wanli Ouyang, Limin Wang, Min Dou, Xizhou Zhu, Tong Lu, Dahua Lin, Jifeng Dai, Weijie Su, Bowen Zhou, Kai Chen, Yu~Qiao, Wenhai Wang, and Gen Luo.
\newblock Internvl3.5: Advancing open-source multimodal models in versatility, reasoning, and efficiency, 2025{\natexlab{b}}.
\newblock URL \url{https://arxiv.org/abs/2508.18265}.

\bibitem[Wei et~al.(2024)Wei, Kong, Chen, Zhao, Ge, Yang, Sun, Han, and Zhang]{wei2024vary}
Haoran Wei, Lingyu Kong, Jinyue Chen, Liang Zhao, Zheng Ge, Jinrong Yang, Jianjian Sun, Chunrui Han, and Xiangyu Zhang.
\newblock {Vary}: Scaling up the vision vocabulary for large vision-language models.
\newblock In \emph{European Conference on Computer Vision}, pages 408--424. Springer, 2024.

\bibitem[Wei et~al.(2025)Wei, Sun, and Li]{wei2025deepseek}
Haoran Wei, Yaofeng Sun, and Yukun Li.
\newblock Deepseek-ocr: Contexts optical compression.
\newblock \emph{arXiv preprint arXiv:2510.18234}, 2025.

\bibitem[Wei et~al.(2026)Wei, Sun, and Li]{wei2026deepseek}
Haoran Wei, Yaofeng Sun, and Yukun Li.
\newblock Deepseek-ocr 2: Visual causal flow.
\newblock \emph{arXiv preprint arXiv:2601.20552}, 2026.

\bibitem[Xia et~al.(2024)Xia, Malladi, Gururangan, Arora, and Chen]{xia2024less}
Mengzhou Xia, Sadhika Malladi, Suchin Gururangan, Sanjeev Arora, and Danqi Chen.
\newblock {LESS}: Selecting influential data for targeted instruction tuning.
\newblock In \emph{Proceedings of the 41st International Conference on Machine Learning}, volume 235 of \emph{Proceedings of Machine Learning Research}, pages 54104--54132. PMLR, 2024.

\bibitem[Xie et~al.(2023)Xie, Pham, Dong, Du, Liu, Lu, Liang, Le, Ma, and Yu]{xie2023doremi}
Sang~Michael Xie, Hieu Pham, Xuanyi Dong, Nan Du, Hanxiao Liu, Yifeng Lu, Percy~S Liang, Quoc~V Le, Tengyu Ma, and Adams~Wei Yu.
\newblock {DoReMi}: Optimizing data mixtures speeds up language model pretraining.
\newblock \emph{Advances in Neural Information Processing Systems}, 36:\penalty0 69798--69818, 2023.

\bibitem[Yin et~al.(2026)Yin, Liu, YY, Xie, Liu, Yang, Wang, Liu, Zou, Chen, Wei, Wu, Huang, Wu, Wang, Du, and Jia]{yin2026unlimitedocrworks}
Youyang Yin, Huanhuan Liu, YY, Qunyi Xie, Chaorun Liu, Shiqi Yang, Shaohua Wang, Zhanlong Liu, Hao Zou, Jinyue Chen, Shu Wei, Jingjing Wu, Mingxin Huang, Zhen Wu, Guibin Wang, Tengyu Du, and Lei Jia.
\newblock Unlimited ocr works, 2026.
\newblock URL \url{https://arxiv.org/abs/2606.23050}.

\bibitem[Yu et~al.(2026)Yu, Zhan, Liu, Zhao, Yue, Chen, Wang, Sun, Li, Lyu, et~al.]{yu2026padoc}
Hao Yu, Jiabo Zhan, Kang Liu, Linnan Zhao, Dongxu Yue, Rui Chen, Jinglin Wang, Chong Sun, Chen Li, Jing Lyu, et~al.
\newblock {PaDoc}: Layout-grounded parallel decoding for document parsing.
\newblock \emph{arXiv preprint arXiv:2608.06146}, 2026.

\bibitem[Yu et~al.(2025)Yu, Wang, Wang, Huang, Ma, He, Cai, Chen, Huang, Zhao, et~al.]{yu2025minicpmv45cookingefficient}
Tianyu Yu, Zefan Wang, Chongyi Wang, Fuwei Huang, Wenshuo Ma, Zhihui He, Tianchi Cai, Weize Chen, Yuxiang Huang, Yuanqian Zhao, et~al.
\newblock {MiniCPM-V 4.5}: Cooking efficient {MLLMs} via architecture, data, and training recipe.
\newblock \emph{arXiv preprint arXiv:2509.18154}, 2025.
\newblock URL \url{https://arxiv.org/abs/2509.18154}.

\bibitem[Zhang et~al.(2026)Zhang, Liu, Liang, Zhang, Xiang, Liu, Sun, Lin, Zhang, Zhou, et~al.]{zhang2026paddleocr}
Zelun Zhang, Hongen Liu, Suyin Liang, Yubo Zhang, Yiqing Xiang, Jiaxuan Liu, Ting Sun, Manhui Lin, Yue Zhang, Changda Zhou, et~al.
\newblock {PaddleOCR-VL-1.6}: Expanding the frontier of document parsing with under-optimized region refinement and progressive post-training.
\newblock \emph{arXiv preprint arXiv:2606.03264}, 2026.

\bibitem[Zhao et~al.(2026)Zhao, Liu, Yu, Zhan, Sun, and Li]{zhao2026ocredr}
Linnan Zhao, Kang Liu, Hao Yu, Jiabo Zhan, Chong Sun, and Chen Li.
\newblock {OCR-EDR}: Rendering-aware diagnosis and repair for closed-loop {OCR} improvement.
\newblock \emph{arXiv preprint arXiv:2609.03445}, 2026.
\newblock URL \url{https://arxiv.org/abs/2609.03445}.

\bibitem[Zheng et~al.(2026)Zheng, Li, Zhang, Xin, Zhao, Liu, Chen, Lou, Fu, Yang, et~al.]{zheng2026multimodal}
Handong Zheng, Yumeng Li, Kaile Zhang, Liang Xin, Guangwei Zhao, Hao Liu, Jiayu Chen, Jie Lou, Qi~Fu, Rui Yang, et~al.
\newblock Multimodal {OCR}: Parse anything from documents.
\newblock \emph{arXiv preprint arXiv:2603.13032}, 2026.

\bibitem[Zhong et~al.(2020)Zhong, ShafieiBavani, and Jimeno~Yepes]{zhong2020image}
Xu~Zhong, Elaheh ShafieiBavani, and Antonio Jimeno~Yepes.
\newblock Image-based table recognition: Data, model, and evaluation.
\newblock In \emph{European Conference on Computer Vision}, pages 564--580. Springer, 2020.

\bibitem[Zhong et~al.(2026{\natexlab{a}})Zhong, Chen, Zeng, Zhao, Jiang, Zheng, Huang, Qiu, Shi, Yang, and Ma]{zhong2026fdrl}
Yufeng Zhong, Lei Chen, Zhixiong Zeng, Xuanle Zhao, Deyang Jiang, Liming Zheng, Jing Huang, Haibo Qiu, Peng Shi, Siqi Yang, and Lin Ma.
\newblock Reading or reasoning? format decoupled reinforcement learning for document {OCR}.
\newblock In \emph{Proceedings of the IEEE/CVF Conference on Computer Vision and Pattern Recognition}, pages 33164--33173, 2026{\natexlab{a}}.

\bibitem[Zhong et~al.(2026{\natexlab{b}})Zhong, Chen, Zhao, Han, Zheng, Huang, Jiang, Cao, Ma, and Zeng]{zhong2026ocrverse}
Yufeng Zhong, Lei Chen, Xuanle Zhao, Wenkang Han, Liming Zheng, Jing Huang, Deyang Jiang, Yilin Cao, Lin Ma, and Zhixiong Zeng.
\newblock Ocrverse: Towards holistic ocr in end-to-end vision-language models.
\newblock \emph{arXiv preprint arXiv:2601.21639}, 2026{\natexlab{b}}.

\end{thebibliography}
